\documentclass{article}

\usepackage{arxiv}
\RequirePackage{amsthm,amsmath,amsfonts,amssymb,bm}
\RequirePackage[authoryear]{natbib}
\RequirePackage[colorlinks,citecolor=blue,urlcolor=blue, linkcolor=blue]{hyperref}
\usepackage{float}
\usepackage{comment}
\usepackage{algorithm}
\usepackage{subcaption}
\usepackage{tabularx}   % flexible column width (X)
\usepackage{array}       % better column types, spacing
\usepackage[utf8]{inputenc} % allow utf-8 input
\usepackage[T1]{fontenc}    % use 8-bit T1 fonts
\usepackage{url}            % simple URL typesetting
\usepackage{booktabs}       % professional-quality tables
\usepackage{nicefrac}       % compact symbols for 1/2, etc.
\usepackage{microtype}      % microtypography
\usepackage{graphicx}
\usepackage{xcolor}

\newtheorem{remark}{\it Remark}[section]
\newtheorem{property}{Property}

\title{Conformal bandits: bringing statistical validity and reward efficiency under weak arm separability}

\author{
    {\includegraphics[scale=0.06]{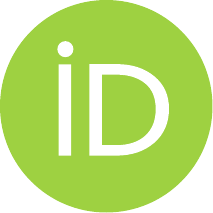}\hspace{1mm}Simone Cuonzo} \\
	MEMOTEF, Sapienza Università di Roma (IT)\\
	\texttt{simone.cuonzo@uniroma1.it}\\
    \And
    \href{https://orcid.org/0000-0003-2501-8795}
    {\includegraphics[scale=0.06]{orcid.pdf}\hspace{1mm}Nina Deliu}\\
	MEMOTEF, Sapienza Università di Roma (IT)\\
    MRC – Biostatistics Unit, University of Cambridge (UK) \\
    \texttt{nina.deliu@uniroma1.it} \\
}

\date{}

\begin{document}
\maketitle

\begin{abstract}
We introduce \textit{Conformal Bandits}, a novel framework integrating {\it Conformal Prediction} (CP) into bandit problems, a classic paradigm for sequential decision-making under uncertainty. Traditional regret-minimisation bandit strategies like Thompson Sampling and Upper Confidence Bound (UCB) typically rely on distributional assumptions or asymptotic guarantees; further, they remain largely focused on regret, neglecting their statistical properties. We address this gap. Through the adoption of CP, we bridge the regret-minimising potential of a decision-making bandit policy with statistical guarantees in the form of finite-sample prediction coverage. We demonstrate the potential of {\it Conformal Bandits} through simulation studies and an application to portfolio allocation, a typical scenario where differences in arm rewards are far too small (weak arm separability) for classical policies to be optimal in finite sample. We showcase our framework's practical advantage in terms of regret in this setting, as well as its added value in achieving nominal coverage guarantees where classical UCB policies may fail. Focusing on our application of interest, we further illustrate how integrating hidden Markov models to capture the regime-switching behaviour of financial markets, enhances the {\it exploration-exploitation} trade-off, and translates into higher risk-adjusted returns, while preserving coverage guarantees.
\end{abstract}

% keywords can be removed

\section{Introduction} \label{sec:intro}
% small gap emphasis
It is relatively straightforward to discern and make good decisions when differences in their expected outcomes or {\it rewards} are substantial. When one arm clearly dominates, even simple heuristics rapidly identify the best alternative, incurring minimal regret. However, despite practical relevance, realistic environments rarely exhibit such a clear separation. Many medical treatments show near-identical efficacy~\citep{villar2015multiarmed, thall2007practical}, yet even small improvements can affect patient outcomes, safety, and regulatory decisions~\citep{ousmen2018distribution, food2016non}. Digital and behavioural health interventions targeting, e.g., physical activity often yield negligible effects~\citep{lee2025personalizedexerciseassistantusing,szaszi2022no}; still, their impact may be notable within subpopulations affected by conditions such as diabetes~\citep{aguilera2024effectiveness}. Similarly, in domains like marketing and finance, competing advertisements and financial instruments seldom differ by large, persistent margins~\citep{demiguel2009optimal}. Specialists must therefore discern among options whose short-term engagement metrics or expected returns are nearly indistinguishable, with the understanding that even small margins can accumulate into significant differences over extended periods and large numbers of users. 

{\it Weak arm separability} reflects settings where the arms are difficult to distinguish in finite samples, making reliable inference and decision-making particularly difficult. Two related but conceptually distinct conditions can give rise to this scenario. The first is the {\it small-gap regime}: 
%{\it Small-gap regimes} represent settings where reliable inference and decision-making becomes particularly difficult. Formally, 
denoted by $\mu^*$ and $\mu$ the mean reward of the optimal arm and a suboptimal arm, respectively, this refers to a difference $\Delta = \mu^* - \mu$ that is very small, with the case $\Delta \asymp 1/\sqrt{n}$ being the statistically ``hardest'' scale. This regime is of interest in domains from reinforcement learning~\citep{simchowitz2019non}, classical bandits~\citep{lai1985asymptotically, kalvit2021closer} and best-arm identification~\citep{jamieson2014lil,mannor2004sample} to sequential hypothesis testing~\citep{le_cam_asymptotic_1986} and causal inference, where this gap is referred to as average treatment effect~\citep{imbens2015causal, hirano2009asymptotics}. It also covers the limit experiments framework~\citep{le_cam_asymptotic_1986, hirano2009asymptotics} and classical diffusion scaling~\citep{glynn1990diffusion}, which corresponds to instances that statistically constitute the ``worst-case'' for both hypothesis testing and regret minimisation. Connections to these and other domains are discussed in \cite{kato2022best} and \cite{kalvit2021closer}. The second is the {\it low signal-to-noise ratio} (SNR) regime, that is, a small ratio $\Delta/\sigma$, where $\sigma$ is the noise level of the reward distributions. In practice, and as we document empirically in our application (Section~\ref{sec: application}), the two conditions frequently co-occur: the small-gap regime is a sufficient condition for low SNR whenever rewards are noisy, and both jointly constitute the setting in which standard decision-making policies are least informative.

% Intro to bandits
The focus of this work is the classical {\it regret-minimising multi-armed bandit} (MAB) problem~\citep{lattimore2020bandit}, increasingly embraced by numerous applications, ranging from healthcare~\citep{villar2015multiarmed, deliu2024reinforcement} to finance~\citep{charpentier2023reinforcement}. MABs provide a fundamental framework for modelling sequential decision-making under uncertainty. In their standard formulation, a decision-maker repeatedly selects one {\it arm} from a pre-specified finite set and receives a {\it stochastic reward}. The objective is to minimise regret or, equivalently, to maximise cumulative rewards over time, while learning their distributions online as new data are observed. This tension between {\it exploration} of uncertain arms and {\it exploitation} of seemingly superior ones has made MAB algorithms central to modern theoretical and applied statistics and machine learning.

% focus su ucb (ottica confidence interval) e problema nello small-gap. Mostrare regret non-log

\begin{figure}
    \centering
    \includegraphics[width=1\linewidth]{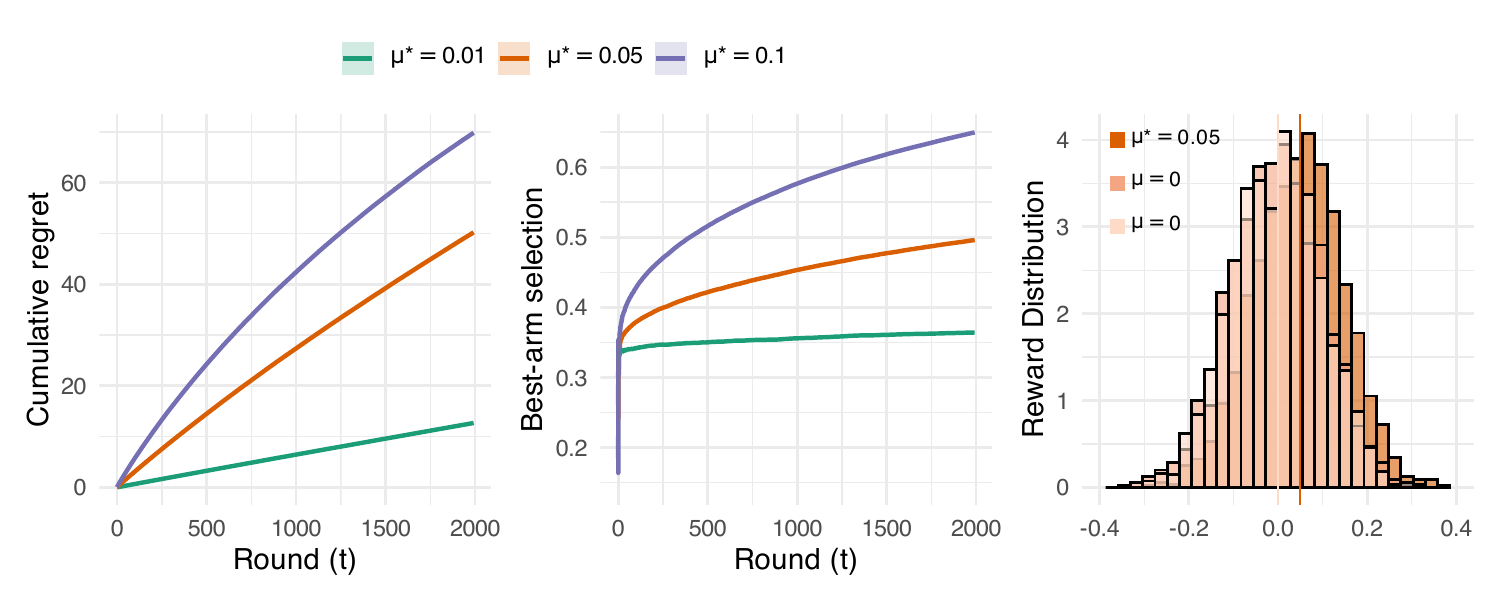}
    \caption{Cumulative regret (left) and best-arm selection (center) attained with UCB1~\citep{auer2002finite} in a three-armed bandit with gap $\Delta_k = \mu^* - \mu_k$, where $\mu_k = 0$ for suboptimal arms $k \neq k^*$, and optimal arm mean $\mu^* = \mu_{k^*} \in \{0.01, 0.05, 0.1\}$. Arm rewards are drawn from a normal distribution $\mathcal{N}(\mu_k, \sigma = 0.1)$; right plot provides a comparison for $\Delta = \mu^* = 0.05$, case study in Section~\ref{sec: Simulation Studies}.}
    \label{fig: Fig1}
\end{figure}

A wide range of strategies have been developed for the MAB problem. These include index-based approaches~\citep{gittins2011multi}, Bayesian bandits such as the Thompson sampling algorithm~\citep{thompson1933likelihood}, and the large family of {\it Upper Confidence Bounds}~\citep[UCB;][]{auer2002using}, among others. However, most of the existing literature addresses the large-gap regime, leaving the small-gap and the low-SNR challenges open. To illustrate, the celebrated UCB policy enjoys strong theoretical guarantees in terms of both logarithmic regret~\citep{lai1985asymptotically} and arm-sampling rates~\citep{audibert2009exploration}, in different parametric settings with $\Delta$ bounded away from $0$. The underlying principle of UCB is the so-called optimism in the face of uncertainty: for each arm, a confidence interval quantifying uncertainty around its mean estimate is constructed, and the arm with the highest upper bound is selected, thereby encouraging exploration of less certain options. Clearly, the performance of UCB critically relies on the quality of the derived confidence bounds. While for large gaps or high-SNR settings, UCB quickly identifies and converges to allocating the optimal arm, in small-gap or low-SNR regimes, the classical confidence intervals often become unstable or excessively conservative. This results in either slow learning and delayed identification of the best alternative (due to an excessive need of exploration) or in too-early (noise-driven) exploitation, with a systematic misallocation of arms. In practice, obtaining confidence intervals that are sufficiently informative to distinguish between competing arms demands large sample sizes per arm, i.e., approximately $1/\Delta^2$ for distributions with gap $\Delta$. Over finite horizons $T$, this leads to a cumulative regret that exhibits linear growth; Figure~\ref{fig: Fig1} provides an illustration for $T = 2000$ across a range of gap values. The degradation is most severe precisely at the small-gap scale $\Delta \asymp 1/\sqrt{T}$, which, for a fixed $\sigma$, directly governs the SNR, compounding the difficulty of distinguishing arms in finite samples for any policy relying on distribution-blind bounds. %Notably, since the minimax regret bound for UCB satisfies $R_T \lesssim \sqrt{nK\log n} + \sum_k\Delta_k$ \citep{lattimore2020bandit}, with the dominant first term being entirely {\it gap-independent}, a low SNR may be a fundamental issue in classical bandit policies. 
This challenge limits the practical effectiveness of UCB-type policies in many real-world problems dominated by noisy small-gap scenarios, such as financial markets, and underscores the importance of revisiting the classical confidence bounds used in UCB.

% cp (ottica prediction): possible solution for small-gap

{\it Conformal prediction}~\citep[CP;][]{vovk2005, fontana2023conformal, angelopoulos_learn_2025} offers an innovative yet flexible and robust non-parametric alternative to construct uncertainty bounds. It provides a framework for deriving statistically valid prediction regions that guarantee nominal finite-sample coverage under the sole assumption of exchangeability. Its model- and distribution-free nature makes CP particularly appealing for real-world problems where classical model assumptions may fail. In the context of bandit problems, CP offers a promising route to quantifying the uncertainty of future arm rewards, complementing the regret-efficiency of bandit algorithms with finite-sample statistical validity. 
This opens a natural path to replacing classical UCB's bounds based on the Hoeffding's inequality~\citep{hoeffding1963probability} with more flexible, adaptive, and potentially more informative solutions. Through guaranteed nominal coverage for future rewards, CP also facilitates risk monitoring and supports risk-informed decision-making for improved regret. We note that related ideas have appeared under distinct settings. \cite{krishnamurthy2023conformal} introduce Conformal Arm Sets, providing near-optimal minimax regret bounds. Our approach differs mechanically: rather than constructing sets of plausible optimal arms, we build arm-specific prediction intervals for future reward values. \cite{stanton2023bayesian} derive a conformal UCB acquisition function within Bayesian optimisation using a Gaussian process surrogate; their framework is model-based, batch, and does not target sequential coverage guarantees, making the settings sufficiently distinct. Furthermore, our proposal specifically targets the weak-separability regime, as well as risk-aware decision-making.

%The focus of CP is a prediction interval for a future outcome, thereby quantifying uncertainty in a prediction rather than a parameter or mean estimate. Being non-parametric, it provides a more robust and model-agnostic solution and particularly suited to heavy-tailed or heteroscedastic data. 
%By using the upper endpoint of the CP interval as the index for each arm, we obtain a CP-UCB hybrid that preserves the exploration–exploitation structure but benefits from distribution-free coverage, adaptive calibration, and robustness in small-gap environments. 

\subsection{Motivating problem: portfolio allocation}

Portfolio allocation is a central problem in finance, concerned with determining how to distribute capital across a set of investment assets so as to maximise returns while controlling risk~\citep{zhu_robust_2018}. Classic approaches, most notably Markowitz’s mean-variance framework ~\citep{markowitz1952modern, 10.1214/10-AOAS422}, provide a principled foundation but suffer from well-known limitations: parameter estimates are unstable, the resulting portfolios are sensitive to estimation error, and their performance can deteriorate substantially under changing market conditions. These challenges have motivated the development of sequential and adaptive portfolio selection methods capable of responding to evolving financial environments.

Within this dynamic stream of research, it appears natural to frame the sequential nature of investment decisions as a MAB: each potential portfolio composition represents an arm, and the realised portfolio performance or return corresponds to its reward. Contextual bandit variants can market conditions, economic indicators, or asset characteristics as context to inform allocation decisions. This formulation enables the use of exploration-exploitation mechanisms to adaptively identify profitable strategies, updating beliefs from ongoing performance and adjusting portfolio allocation over time. Several contributions have explored this idea. For instance, \cite{shen_wang_jiang_zha2015} constructed orthogonal portfolios from multiple assets and integrated them within a UCB-based online framework, yielding a strategy that blends passive and active investments through a risk-adjusted reward criterion. Later, \cite{shen_wang2016_portfolio_blending} applied Thompson sampling to blend two portfolios based on different investment principles, generating robust and high-quality strategies. This approach was later extended by \cite{fujishima_nakagawa2022}, who generalised the blending algorithm to allow investors to efficiently combine three or more portfolios.

Despite advances, portfolio allocation presents challenges that are substantially more complex than those encountered in classical bandits. First, asset returns often exhibit minimal differences in expected performance, making the arm reward separability small and difficult to exploit~\citep{demiguel2009optimal}. Second, return volatility presents heterogeneous and time-dependent patterns, yielding non-stationary reward distributions, with changing market conditions over time~\citep{fisher2020optimal}. Third, the objective in finance is rarely to maximise expected returns alone: practical allocation rules must balance reward and downside risk, as strategies that perform well in favourable regimes may experience severe losses during downturns~\citep{mcneil2015quantitative}. More technically, a policy must balance between regret and statistical guarantees, a tension that has been largely characterised by the literature at the intersection~\citep{ashutosh2021bandit, yao2021power, deliu2025finite}. Overall, these characteristics highlight the {\bf need for bandit policies that are not only regret efficient but also capable of reliably quantifying uncertainty while incorporating risk-awareness into the decision process, as well as adapting to time-varying market conditions}. %This motivates the development of the conformal bandit framework and their regime-aware variants introduced in this work.

\subsection{Contribution and structure of the work}
This paper makes four main contributions. First, we introduce \textit{Conformal UCB}, replacing classical Hoeffding-based confidence bounds with CP intervals to achieve finite-sample valid prediction regions. Second, we develop a general \textit{Conformal Bandits} framework that treats UCB as a special case but extends to risk-aware decision-making through %novel indices such as the \textit{Exploratory Skewness Index}, exploiting CP's ability to capture distributional asymmetry and distinguish upside potential from downside risk. 
a family of policies parametrised by a mixing coefficient
$\lambda \in [0,1]$, which combines the upper and lower conformal bounds via a convex combination. This formulation, grounded in the Hurwicz optimism--pessimism criterion~\citep{hurwicz1951optimality}, enables a principled and smooth interpolation between fully optimistic and fully conservative arm selection, while satisfying formal guarantees of shift-equivariance and
dominance-consistency. Third, through simulation studies, we demonstrate the advantages of {\it Conformal Bandits}, both in terms of nominal coverage guarantees (not attainable with classical UCB), and superior cumulative regret in the small-gap regime. Fourth, we showcase the applicability of {\it Conformal Bandits} to portfolio allocation, addressing the specific challenges of this setting: small gaps, non-stationarity, changing market conditions. For the latter, we integrate {\it hidden Markov models}~\citep[HMM;][]{zucchini_macdonald_langrock2016} to capture regime-switching market dynamics. By conditioning on inferred regimes, our CP-based bandit adapts exploration to changing conditions, achieving higher risk-adjusted returns while maintaining robust coverage guarantees across market environments.

The remainder of this work is structured as follows. Section~\ref{sec: notation} formalises the problem and Section~\ref{sec: MAB} outlines the MAB framework and the UCB type of policies. The main methodological contributions are presented in Section~\ref{sec: CP-bandits}, where CP is also introduced. Empirical evaluations are illustrated in Section~\ref{sec: Simulation Studies} and Section~\ref{sec: application}, for the simulation studies and the portfolio application, respectively. Section~\ref{sec: conclusion} concludes with a discussion on future work directions. 

\section{Problem Setting and Notation} \label{sec: notation}

We consider an online sequential decision-making problem, where, at each time-step or round $t = 1, 2, \ldots, T$, an agent must select an \textit{arm} $A_t$ from a finite set of alternatives $\mathcal{A}=\{1, \ldots, K\}$, with cardinality $|\mathcal{A}| = K$. For each selected arm, the agent receives a feedback or reward whose distribution is generally unknown and can differ across arms. Conforming to the potential outcomes framework of causal inference~\citep{imbens2015causal}, let $Y_{k,t}$ be the random \textit{reward} variable representing the outcome that would be observed if arm $A_t = k$ were chosen in round $t$. At this stage and for the general methodological framework that will be presented in Sections~\ref{sec: MAB} and~\ref{sec: CP-bandits}, we do not make any assumptions on the reward distribution. Denoted by $\mu_k = \mathbb{E}[Y_{k,t}]$ the (unknown but fixed) mean reward of arm $k$, the typical agent's objective is to design an optimal decision-making policy, say $\bm\pi^*$, so as to maximise the {\it expected cumulative reward}, that is,
\begin{align} \label{eq: maxrew}
\bm\pi^* \doteq\max_{\bm\pi} \; \mathbb{E}_{\bm\pi}\!\left[\sum_{t=1}^{T} Y_{A_t,t}\right],
\end{align}
where $\bm \pi = (\bm \pi_t)_{t=1,\dots,T}$ and $\bm \pi_t = (\pi_{1,t},\dots, \pi_{K,t})$. Denote now by $\mathcal{F}_{t-1} \doteq \{A_{\tau}, Y_{k_\tau, \tau}, \tau = 1,\dots,t-1 \}$ the filtration at time $t-1$, that is, the potential history of selected arms and associated rewards prior to round $t$. The policy at each $t \geq 1$ is conceived as a function of $\mathcal{F}_{t-1}$, that is, $\pi_{k,t} \doteq \mathbb{P}(A_{t}=k | \mathcal{F}_{t-1})$, for $k \in {1,\dots, K}$. Prior to the first decision point $t=1$, it is common to consider a warm-up or pure exploration phase where a minimum amount of information on each arm's reward is gained following a uniform policy $\pi_{0,k} = 1/K$~\citep{liu2025thompson, kaibel2021rethinking, pin2025informed}. In tandem with other regularisation tricks such as probability clipping~\citep{may2021evaluating, yao2021power}, this mitigates the risk of erroneous exploitation of inferior arms and improves statistical inference in terms of both estimation and hypothesis testing. In particular, clipping arms probabilities away form $0$ and $1$, e.g., setting $\pi_{k,t} \in [\pi_{\min},\pi_{\max}]$, with $ 0 < \pi_{\min} \leq \pi_{\max} < 1$, has shown important benefits in terms of statistical power to detect arm differences~\citep[see e.g.,][]{may2021evaluating, yao2021power}.
 %and randomized versions of the upper confidence bound (UCB)~\citep{vaswani20a}. 
This also resonates with the importance of randomisation in inference and causal inference%, both for mitigating sources of bias and enabling inference
~\citep[see e.g.,][]{rosenberger2019randomization}.

Clearly, despite statistical benefits, a high degree of exploration comes at the expense of a loss in potential reward due to the selection of sub-optimal arms. Conversely, always selecting the empirically best arm can lead to linear regret if the initially observed rewards are misleading. Therefore, the overall problem boils down to the long-studied {\it exploration-exploitation} dilemma, according to which: 
\begin{quote}
    ``one must face the conflict between taking actions which yield immediate reward and taking actions whose benefit will come only later.''--~\cite{whittle1980multi}
\end{quote}

\section{MABs and UCB Policies}\label{sec: MAB}

The MAB problem~\citep{lattimore2020bandit, sutton1998reinforcement} exemplifies the fundamental {\it exploration-exploitation} dilemma of sequential decision-making under uncertainty. In the canonical setting, only rewards associated with selected arms are observed, making MABs a prototypical framework for learning under partial information. 

In this work, we consider the stochastic bandit setting, where reward distributions are fixed over time. Following the seminal work of \cite{lai1985asymptotically}, we assume there exists a unique optimal arm $k^* \in \mathcal{A}$ with mean reward $\mu_{k^*} = \mu^* = \max_{k \in \mathcal{A}} \mu_k$. We define the \textit{suboptimality gap} of arm $k$ as $\Delta_k = \mu^* - \mu_k$, which remains constant across all rounds. 

The performance of a policy $\bm \pi$ is typically measured through the notion of \textit{expected regret}, defined as the expected difference between the reward of the optimal arm $\mu^*$ and that achieved by following the policy $\bm \pi$:
\[
R^{\bm{\pi}}_T = T \mu^* - \mathbb{E}_{\bm{\pi}}\!\left[\sum_{t=1}^{T} Y_{A_t,t}\right] = \mathbb{E}_{\bm{\pi}}\!\left[\sum_{t=1}^{T} \Delta_{A_t}\right].
\] 
Maximising the expected reward in Eq.~\eqref{eq: maxrew} is equivalent to minimising the expected regret across $T$ rounds. Under some regularity conditions, \cite{lai1985asymptotically} established fundamental lower bounds on the achievable regret, demonstrating that any optimal policy must incur logarithmic regret. %Subsequent work has refined these bounds and developed near-optimal algorithms~\citep{auer2002finite, cappe2013kullback}. 

A principled approach to resolving the {\it exploration-exploitation} trade-off, with regret bounds matching the fundamental lower bound~\citep{lai1985asymptotically}, is based on the \textit{optimism principle} characterising the UCB-type of policies~\citep{auer2002finite, cappe2013kullback}: maintain confidence bounds on the mean reward of each arm and select the arm with the highest upper confidence bound.

\subsection{The UCB1 policy and extensions} \label{sec: UCB1}
The family of UCB policies~\citep{auer2002finite} formalises the principle of \textit{optimism in the face of uncertainty}, by selecting, at each round $t$, the arm whose estimated reward appears the most promising in light of its mean and uncertainty, that is, under its upper confidence bound. The canonical UCB policy, namely UCB1, selects arms as
\begin{equation} \tag{{\it UCB1}}
a^*_{t+1} = \arg\max_{k \in \mathcal{A}} \Big\{ \widehat{\mu}_{k,t} + c\sqrt{\frac{\beta\log t}{2N_{k,t}}} \Big\},\quad t=1,\dots,T,
\label{eq:ucb1}
\end{equation}
with
\begin{align*}
    \widehat{\mu}_{k,t} = \frac{1}{N_{k,t}}\sum_{i=1}^{t-1} Y_{k,s}\mathbb{I}\{A_i = k\},\qquad N_{k,t} = \sum_{i=1}^{t-1} \mathbb{I}\{A_i = k\},
\end{align*}
representing the empirical mean reward of arm $k$ at round $t$, and the number of times arm $k$ has been selected up to round $t$, respectively. Here, $\widehat{\mu}_{k,t}$ captures the \textit{exploitation} component of the policy, whereas the second term represents the \textit{exploration bonus}, which quantifies the statistical uncertainty associated with the estimate $\widehat{\mu}_{k,t}$, which is framed as a decreasing function of $N_{k,t}$. The hyperparameter $c$ reflects the range of the arm rewards: for rewards in $[a,b]$ it is typically set to $b-a$. The hyperparameter $\beta > 0$ modulates the overall confidence level: larger $\beta$ values correspond to wider confidence bounds and therefore stronger exploration, while smaller values favour exploitation. The exploration bonus in UCB1 derives from Hoeffding's inequality~\citep{hoeffding1963probability}, which provides concentration bounds for bounded random variables. In its simplest form, with rewards in $[0,1]$, it guarantees that with probability at least $1 - \delta$, we have $|\widehat{\mu}_{k,t} - \mu_k| \leq \sqrt{\frac{\log(1/\delta)}{2N_{k,t}}}$ for the true mean. Setting $\delta = t^{-\beta}$ and applying a union bound over arms and rounds yields the logarithmic bound. By construction, this bound depends on the data only through $N_{k,t}$, neglecting important aspects of their distribution, therefore, with limited adaptability to the underlying reward process.

Several extensions of the UCB1 policy have been proposed in the literature. Notable variants include KL-UCB~\citep{cappe2013kullback}, which replaces Hoeffding-based bounds with tighter Kullback-Leibler divergence confidence intervals, Bayesian UCB~\citep{kaufmann2012bayesian}, which incorporates prior information, and randomised UCB~\citep{vaswani20a}, a randomised alternative to trade-off exploration and exploitation. A particularly relevant extension for our application augments UCB1 with risk sensitivity through a variance component~\citep{vakili16, hu2025multi}, directly translating the modern portfolio theory of~\cite{markowitz1952modern}. Building from pre-existing work on risk-averse bandits~\citep{sani12_riskaware,vakili16}, the idea is that of a {\it Mean-Variance UCB1} policy, which selects arms as
\begin{equation} \tag{\it MV-UCB1}
a^*_{t+1} = \arg\max_{k \in \mathcal{A}}
\Big\{\widehat{MV}^{\rho}_{k,t} + c\sqrt{\frac{\beta \log t}{2N_{k,t}}}
\Big\},
\quad t = 1, \ldots, T,
\label{eq:mvucb1}
\end{equation}
where
\begin{equation}
\widehat{MV}^{\rho}_{k,t}
=
\rho \, \widehat{\mu}_{k,t}
-
(1 - \rho)\, \widehat{\sigma}_{k,t},
\label{eq:mv_score}
\end{equation}
and $\widehat{\sigma}_{k,t}$ denotes the empirical standard deviation of the rewards associated with arm $k$ up to time $t$. The parameter $\rho \in [0,1]$ controls the trade-off between expected reward and volatility. For $\rho \to 1$, the policy converges to the standard UCB1, prioritising high mean rewards. Conversely, for $\rho \to 0$, the selection becomes increasingly conservative, favouring low-volatility arms. In this formulation, $\widehat{MV}^{\rho}_{k,t}$ acts as a risk-adjusted performance score, analogous in spirit to \cite{markowitz1952modern}'s theory, where investors seek to balance expected gain and risk. Given its relevance to out application of interest, this policy will be used as a comparator in Section \ref{sec: application}. 
    
\section{Revising the Bound(s): A Conformal Bandit Approach} \label{sec: CP-bandits}

This section illustrates the main innovation of the paper, which consists in the development of a new bandit framework grounded in a {\it predictive} perspective. Works of a similar flavour can be found in \cite{liu2023nonstationary} and \cite{duranmartingale}, both closely aligned with the alternative Bayesian bandit class of Thompson sampling~\citep{thompson1933likelihood}. These contributions show the potential of a predictive policy in moderate to large-gaps environments, with particular emphasis on non-stationarity~\citep{liu2023nonstationary} and algorithm scalability~\citep{duranmartingale}. However, they remain largely focused on reward and none directly investigates the statistical properties of their proposed predictive approach. Our technical innovation addresses this gap, through the adoption of the conformal prediction framework, which couples the regret-minimising potential of a bandit with statistical guarantees in the form of prediction coverage.

\subsection{Conformal Prediction}

Let ${(X_i, Y_i)}_{i=1}^{t}$ denote a sequence of observed data pairs, where $X_i \in \mathbb{R}^p$, with $p \ge 1$, is the covariate set (referred to as state or context in the bandit literature), and $Y_i \in \mathbb{R}$ the associated response (or reward), here assumed to be a univariate continuous variable.
Given a new state $X_{t+1}$, the goal of CP is to construct a \emph{prediction interval}, say $\mathcal{C}^{1-\alpha}_{t+1} \doteq \mathcal{C}^{1-\alpha}(X_{t+1}) \subseteq \mathbb{R}$, that contains a future unobserved response $Y_{t+1}$ with high probability. The CP framework is non-parametric, making its applicability possible to any underlying predictor $\hat{f}$, be it based on a simple linear model, or a deep learning system. For a comprehensive treatment, we refer to the foundational book of~\cite{vovk2005} and a recent review in \cite{fontana2023conformal}. 

The main statistical guarantee of CP is {\it finite-sample marginal coverage}; that is, given a user-specified miscoverage level $\alpha \in (0,1)$, and assuming that the paired data are {\it exchangeable}, the derived CP interval is ensured to satisfy:
\begin{equation}
\mathbb{P}\left( Y_{t+1} \in \mathcal{C}^{1-\alpha}_{t+1} \right) \ge 1 - \alpha,\quad \forall t.
\label{eq:validity}
\end{equation}
Here, the probability is taken over the joint distribution of all $t+1$ variables $Y_1, \dots, Y_{t+1}$, resembling the frequentist philosophy. In this work, we will interchangeably refer to this property as {\it statistical validity}. For a comparison with prediction intervals under a Bayesian philosophy, see~\cite{deliu2026interplay}.

There are two main approaches to implementing conformal prediction: full CP and split CP~\citep{vovk2005, fontana2023conformal}. A major limitation of former is its computational burden, making it impractical for online learning and bandit settings, where models must be updated frequently and computational efficiency is critical. Conversely, split CP proposes a computationally efficient procedure that decouples model training and calibration in a similar flavour to the validation-set approach~\citep[see e.g.;][]{james_witten_hastie_tibshirani2021}. Specifically, in split CP, available data are partitioned into two disjoint subsets, $\mathcal{D}_{\text{train}}$ (training set), on which a fitted model $\widehat{f}$ is obtained, and $\mathcal{D}_{\text{calib}}$ (calibration set), which is used to compute the so-called calibration or {\it conformity scores}. In a regression setting, these can be thought of as calibration residuals, which quantify the prediction errors of the previously fitted model $\widehat{f}$ on $\mathcal{D}_{\text{train}}$; that is,
\[
S_i = |\widehat{f}(X_i) - Y_i|, \qquad i \in \mathcal{D}_{\text{calib}}.
\]
Assuming that data are exchangeable, a split CP procedure based on a symmetric conformity function (e.g., residual scores), allows the construction of a {\it valid} $1-\alpha$ prediction interval $\mathcal{C}^{1-\alpha}_{t+1}$ for any new test point $X_{t+1}$ as:
\begin{equation}
\mathcal{C}^{1-\alpha}_{t+1} =
\big[
\widehat{f}(X_{t+1}) - Q_{1-\alpha}(S_{\mathcal{D}_{\text{calib}}}),\;
\widehat{f}(X_{t+1}) + Q_{1-\alpha}(S_{\mathcal{D}_{\text{calib}}})
\big],
\label{eq:split_cp}
\end{equation}
where, $Q_{1-\alpha}(S_{\mathcal{D}_{\text{calib}}})$ represents the empirical $(1-\alpha)(1 + 1/|\mathcal{D}_{\text{calib}}|)$-quantile of the calibration scores $S_{\mathcal{D}_{\text{calib}}} = \{S_i, i \in \mathcal{D}_{\text{calib}}\}$, with size given by the cardinality $|\mathcal{D}_{\text{calib}}|$ of the calibration set. The $(1 + 1/|\mathcal{D}_{\text{calib}}|)$ adjustment enables finite-sample guarantees.

\subsection{Conformalised Quantile Regression}\label{sec:CQR}
A main limitation of standard split CP intervals is their lack of adaptability to heteroscedasticity or asymmetry in the (conditional) distribution of the reward: it is clear from Eq.~\eqref{eq:split_cp} that these are symmetric around $\widehat{f}(X_{t+1})$. 
As a solution, \textit{Conformalised Quantile Regression}~\citep[CQR;][]{romano_patterson_candes2019} integrates within CP conformal scores in the form of upper and lower bound residuals from a conditional quantile regression.  

In quantile regression, given a context $X_{t+1}=x$, one can construct $1-\alpha$ prediction intervals for $Y_{t+1}$, by considering the conditional distribution $F(y \mid X_{t+1} = x)$, with $F$ denoting the cumulative density function, and modelling its \textit{conditional quantiles}~\citep{koenker2005quantile}. For a given level $\tau \in (0,1)$, the {\it conditional quantile function} at $X_{t+1}=x$ is defined as
\begin{equation}
q_\tau(x) = \inf\{ y \in \mathbb{R} : F(y \mid X_{t+1} = x) \ge \tau \}.
\label{eq:quantile_function}
\end{equation}
A quantile regression estimator, say $\widehat{q}_{\tau}(x)$, is obtained as the solution of the problem
\begin{equation}
\widehat{q}_\tau(x) = f_\tau(x; \widehat{\theta}_\tau),
\quad
\widehat{\theta}_\tau = \arg\min_{\theta}
\frac{1}{t}\sum_{i=1}^{t} \rho_\tau\!\left(Y_i - f_\tau(X_i; \theta)\right) + \mathcal{R}(\theta),
\label{eq:quantile_regression}
\end{equation}
where $f_\tau(x;\theta)$ is a parametric or nonparametric model (e.g., linear, tree-based, or neural net) posited for the quantile function, $\mathcal{R}(\theta)$ is an optional regularisation term, and $\rho_\tau(u)$ is the ``pinball'' loss function defined as
\[
\rho_\tau(u) =
\begin{cases}
\tau u, & u \ge 0,\\
(\tau - 1) u, & u < 0.
\end{cases}
\]
%Given the estimated conditional quantiles $\widehat{q}_{\alpha/2}(x)$ and $\widehat{q}_{1-\alpha/2}(x)$, one
On this basis, a $1-\alpha$ conditional prediction interval can be obtained as
\begin{equation}
\widehat{C}(x) = \big[\widehat{q}_{\alpha/2}(x),\ \widehat{q}_{1-\alpha/2}(x)\big],
\label{eq:conditional_interval}
\end{equation}
which, nonetheless, % aims to achieve conditional coverage $\mathbb{P}(Y_{t+1} \in \hat{C}(X_{t+1})\mid X_{t+1}=x) \approx 1-\alpha$. However, 
since the estimated quantiles $\widehat{q}_\tau(x)$ are subject to model misspecification and finite-sample errors, are not ensured to have nominal $1-\alpha$ coverage.

CP provides a correction mechanism to quantile regression to achieve valid coverage without assumptions on the distribution of $P_{XY}$, except exchangeability. Following the split CP rationale, first, a lower and an upper quantile regression model is fitted on $\mathcal{D}_{\text{train}}$ to get:
\[
\widehat{q}_{\alpha/2}(x) = f_{\alpha/2}(x; \widehat{\theta}_{\alpha/2}),
\quad
\widehat{q}_{1-\alpha/2}(x) = f_{1-\alpha/2}(x; \widehat{\theta}_{1-\alpha/2}),
\]
each minimising its respective pinball loss function $\rho_\tau(u)$ as in Eq.~\eqref{eq:quantile_regression}. Then, the calibration set is used to compute quantile conformity scores, defined as
\begin{equation}
S_i = \max\big(
\widehat{q}_{\alpha/2}(X_i) - Y_i,\;
Y_i - \widehat{q}_{1-\alpha/2}(X_i)
\big),\qquad i \in \mathcal{D}_{\text{calib}}.
\label{eq:cqr_scores}
\end{equation}

For a new context $X_{t+1}$, the conformalised quantile prediction interval is obtained as:
\begin{equation}
\mathcal{C}^{1-\alpha}_{t+1} =
\big[
\widehat{q}_{\alpha/2}(X_{t+1}) - Q_{1-\alpha}(S_{\mathcal{D}_{\text{calib}}}),\;
\widehat{q}_{1-\alpha/2}(X_{t+1}) + Q_{1-\alpha}(S_{\mathcal{D}_{\text{calib}}})
\big],
\label{eq:cqr_interval}
\end{equation}
with $Q_{1-\alpha}(S_{\mathcal{D}_{\text{calib}}})$ the empirical $(1-\alpha)(1 + 1/|\mathcal{D}_{\text{calib}}|)$-quantile of the calibration scores $S_{\mathcal{D}_{\text{calib}}}$, as per standard split CP.

Under the exchangeability of ${(X_i, Y_i)}_{i=1}^{t+1}$, such a prediction interval satisfies {\it validity} in Eq.~\eqref{eq:validity}, irrespective of the posited regression model, and addresses potential misspecification. Further, by construction, this interval adapts to heteroscedastic or heavy-tailed settings. As a result, CQR inherits the distribution-free, model-free and finite-sample coverage guarantee of CP, while retaining the \textit{local adaptivity} of quantile regression.

\begin{remark}
   The main assumption in conformal prediction is exchangeability, which is often violated in practice. For example, in time series, data may often exhibit dependency structures that are not symmetric over time. In bandits, this assumption is further challenged by the sequential dependency in arm selection introduced by the adaptive policy. Therefore, even in cases where the reward distribution is expected to have an exchangeable structure, the sequential dependency of arm choices may compromise exchangeability over the joint distribution of arm selection and reward. Extension of CP to addressing this problem, with a focus on times series, exists and have been recently surveyed and compared in \cite{stockerfontana2025time}. Among these, the {\it Adaptive Conformal Inference} (ACI) framework of \cite{gibbs2021adaptive} introduces a simple and intuitive, yet theoretically valid, solution. The idea is to adaptively adjust the $\alpha$ level to account for excessive or reduced coverage of the CP interval. Throughout Sections~\ref{sec: Simulation Studies} and \ref{sec: application}, we will adopt the ACI extension to CQR.
\end{remark}

\subsection{Conformal Bandits} The {\it Conformal Bandit} framework extends the standard bandit paradigm by integrating sequential decision-making under uncertainty with rigorous CP-based uncertainty quantification. Rather than relying on exploration bonuses defined through data-independent functional forms such as the classical $\sqrt{\beta \log t / N_{k,t}}$ term, {\it Conformal Bandits} construct empirically calibrated prediction intervals directly from historical reward observations. This data-driven approach offers a natural adaptation to the observed data, accounting for non-normal patterns, including heavy tails as well as asymmetries.

Consider a multi-armed setting with $k = 1,\dots,K$. At each round $t$ of a sequential decision making process, we have collected a set $\left(X_{i}, Y_{k,i}\right)_{i=1}^{N_{k,t}}$ of observed contexts / states and rewards associated with each arm $k$, explored for $N_{k,t}$ rounds. We assume states $X_1,\dots,X_t$ are not arm specific, omitting the index $k$. Given a new state $X_{t+1}$ and adopting the CQR approach illustrated in Section~\ref{sec:CQR}, we construct arm-specific prediction intervals for future rewards $Y_{k,t+1}$, $k = 1,\dots, K$ as:
\begin{align} \label{eq:cp_bandit_interval}
\mathcal{C}_{k,t+1}^{1-\alpha} &= \Big[L_{k,t+1},\, U_{k, t+1}\Big] = \Big[
\widehat{q}_{k,\frac{\alpha}{2}}(X_{t+1}) - Q_{1-\alpha}(S_k),\;
\widehat{q}_{k, 1-\frac{\alpha}{2}}(X_{t+1}) + Q_{k,1-\alpha}(S_k)
\Big],
\end{align}
where $\widehat{q}_{k, \tau}(x)$ is the $\tau$-level quantile predictor for the arm $k$ reward with state $x$, and $Q_{1-\alpha}(S_k)$ is the adjusted conformal quantile computed on arm-$k$ specific calibration scores. By virtue of CP properties, this interval satisfies:
\[ \mathbb{P}\!\left(Y_{k,t+1} \in \mathcal{C}^{1-\alpha}_{k,t+1}\right) \ge 1 - \alpha,\quad k=1,\dots K,\quad t=1,\dots,T.
\]

%\subsubsection{Conformal UCB Policy}

The upper bound $U_{k,t+1}$ in Eq.~\eqref{eq:cp_bandit_interval} can be directly used to reformulate the classical UCB1 policy under a conformal (predictive) paradigm, where, at each new round $t+1$,
\begin{equation} \tag{\it CP-UCB}
a^*_{t+1} = \arg\max_{k \in \mathcal{A}} U_{k,t+1},
\label{eq:cp_ucb}
\end{equation}
that is, the arm with the highest \textit{conformal upper bound} is selected. It preserves the optimism-in-the-face-of-uncertainty principle of standard UCB but replaces confidence intervals with statistically valid prediction intervals, guaranteeing finite-sample coverage and adaptive exploration. In light of its resemblance to UCB-type strategies, we term this first policy \textit{Conformal UCB} or \textit{CP-UCB}.

%\subsubsection{Conformal Risk-Aware (CRA) Policies}
While \textit{CP-UCB} is oriented toward reward maximisation, the CP framework is flexible enough to accommodate different risk profiles. %In particular, several variants can be defined depending on the decision-making context, especially relevant in financial applications, 
This is particularly important in contexts such as finance, where the balance between upside potential and downside protection is crucial. To this end, we introduce a more flexible decision mechanism that aggregates information from both ends of the CP interval through a \emph{convex combination} of the
upper and lower conformal bounds. Formally, for each new round $t+1$, the optimal arm is selected as:
\begin{equation} \tag{\it CP-Bandit}
a^*_{t+1} = \arg\max_{k \in \mathcal{A}} \Big\{
(1-\lambda) U_{k,t+1} + \lambda L_{k,t+1}\Big\},
\label{eq:cp_blend}
\end{equation}
where the mixing parameter $\lambda \in [0,1]$ regulates the trade-off between downside protection $L_{k,t+1}$ relative to upside potential $U_{k,t+1}$. The score is a proper weighted average of the two endpoints of the CP interval, and as such admits a natural interpretation in terms of the Hurwicz optimism--pessimism criterion of classical decision theory under uncertainty \citep[see e.g.][]{hurwicz1951optimality}. The convex-combination structure endows the selection rule with two desirable formal properties outlined below.

\begin{property}[Shift-equivariance]
\label{prop:shift}
The ranking induced by the CP-Bandit score is invariant to any additive shift of the
reward scale. That is, for any constant $d \in \mathbb{R}$, replacing
$(L_{k,t+1},\, U_{k,t+1})$ by $(L_{k,t+1}+d,\, U_{k,t+1}+d)$ for all $k$ leaves the
arm ordering unchanged.
\end{property}
\begin{proof}
Since the weights $(1-\lambda)$ and $\lambda$ sum to one, the score of arm $k$ under the
shifted intervals equals $(1-\lambda)(U_{k,t+1}+d)+\lambda(L_{k,t+1}+d) =
(1-\lambda)U_{k,t+1}+\lambda L_{k,t+1}+d$. The constant $d$ is common to all arms and
therefore cancels in any pairwise comparison.
\end{proof}
 
\begin{property}[Dominance-consistency]
\label{prop:dominance}
If the CP interval of arm $j$ stochastically dominates that of arm $k$ in the
sense that $U_{j,t+1} \ge U_{k,t+1}$ and $L_{j,t+1} \ge L_{k,t+1}$, with at least one
strict inequality, then the CP-Bandit score of arm $j$ strictly exceeds that of arm $k$
for every $\lambda \in [0,1]$.
\end{property}
\begin{proof}
The score difference is $(1-\lambda)(U_{j,t+1}-U_{k,t+1})+\lambda(L_{j,t+1}-L_{k,t+1})$.
Both terms are non-negative and at least one is strictly positive, so the sum is strictly
positive for all $\lambda\in[0,1]$.
\end{proof}
 
Property~\ref{prop:shift} ensures that the selection rule is not distorted by arbitrary location changes in the reward distribution, while Property~\ref{prop:dominance} guarantees that the policy never selects an arm whose entire predictive interval lies below that of another available arm, so that the exploration mechanism cannot be misled by a superficially attractive but genuinely dominated alternative. Further, special cases of $\lambda$ have natural interpretations. When $\lambda = 0$, this corresponds to the fully optimistic {\it CP-UCB} policy, which selects arms solely based on their conformal upper bound. In contrast, $\lambda = 1$ yields the fully conservative policy, driven exclusively by the lower bound. This corresponds to a \emph{maximin} or robust decision rule~\citep{wald1950statistical}:
the arm whose worst predicted outcome is the least severe is preferred. By focusing on the lower tail of the conformal interval, it provides a coverage-guaranteed distribution-free analogue of the popular Value-at-Risk (VaR) financial measure~\citep{mcneil2015quantitative}. This opens up promising directions in risk management, while enriching existing literature on multi-armed bandits under risk criteria~\citep{cassel2018general, pmlr-v29-Galichet13, chang2022unifying}. Intermediate choices $0 < \lambda < 1$ smoothly interpolate between the two extremes balancing between a reward-seeking and downside-protective behaviour and generating a spectrum of risk-sensitive policies. In particular, $\lambda = 0.5$ assigns equal weight to the upper and lower bounds, producing a selection rule that is naturally sensitive to the asymmetry of the predictive interval without privileging either bound. Given its extended generality, we term this family of policies {\it Conformal Bandit} or {\it CP-Bandit}.

\begin{comment}
    
% \paragraph{ (ESI) Policy}
To better exploit potential distributional asymmetries through the conformal bounds, we further introduce a dedicated index, called {\it Exploratory Skewness Index} (ESI), which measures the relative magnitude of the potential upside versus downside risk. This is defined as
\begin{equation} \label{eq:esi}
\mathrm{ESI}_{k,t}=\frac{U_{k,t}}{|L_{k,t}|},\quad k=1,\dots K,\quad t=1,\dots,T.
\end{equation}
Intuitively, an index value $\mathrm{ESI}_{k,t} > 1$ indicates a positively skewed distributional uncertainty (heavier upper tails suggesting potential high-reward opportunities), while $\mathrm{ESI}_{k,t} < 1$ reflects negative skewness (predictive mass concentrated toward losses, signalling higher downside risk); with $\mathrm{ESI}_{k,t} \approx 1$ we expect nearly symmetric uncertainty. 
The integration of such an index with bandits allows the identification of directional asymmetries into the exploration-exploitation trade-off. The corresponding {\it Conformal ESI} or {\it CP-ESI} policy is given by:
\begin{equation} \tag{\it CP-ESI}
a^*_t = \arg\max_{k \in \mathcal{A}} \mathrm{ESI}_{k,t}.
\label{eq:cp_esi}
\end{equation}
As we will illustrate in Section~\ref{sec: Simulation Studies}, this policy exhibits a similar behaviour to the {\it CP-Bandit} policy when $\lambda = 0.5$, that is, when both lower and upper conformal bounds are accounted for in the policy with the same weight.
\end{comment}

This variant illustrates the {\bf versatility of the {\it \bf conformal bandit framework}, enabling generalisations of the exploration-exploitation paradigm while preserving statistically valid uncertainty quantification in dynamic and uncertain environments}. 

\subsubsection{Randomised Conformal Bandits}

UCB policies are often criticised for their lack of randomisation when balancing between exploration and exploitation: given the filtration $\mathcal{F}_{t}$, UCB will always choose the same arm at round $t+1$. This may compromise both statistical and causal inference, as well as best arm identification, due to a potential fast convergence to a local optimum. Therefore, there has been a substantial surge of recent interest in the design of randomised policies or the extension of existing policies within randomised regimes. To this end, in addition to the pseudo-deterministic conformal bandits, a randomised version is now introduced. 

First, we note that our conformal policies (\ref{eq:cp_ucb} and~\ref{eq:cp_blend}) can be represented in a compact form as: 
\begin{equation}
a^{*}_{t} = \arg\max_{k \in \mathcal{A}} I_{k,t},
\label{eq:gen_policy} 
\end{equation}
where $I$ denotes the deterministic \emph{conformal index} given by $I_{k,t} = (1-\lambda)U_{k,t}+\lambda L_{k,t}$. A randomised conformal bandit can be framed to select an arm $A_t$ at round $t$, according to, e.g., the probability law:
\begin{align} \label{eq: rand_CP}
    \pi_{k, t}(a^{*}_{t}, \epsilon) = \mathbb{P}(A_t = k | \mathcal{F}_{t-1}) = (1-\epsilon) \mathbb{I}(k = a_t^*) + \epsilon \frac{\mathbb{I}(k \neq a_t^*)}{K-1},
\end{align}
which injects an additional exploration factor $\epsilon$ limited to the sub-optimal arms $k \neq a_t^*$. For $\epsilon = 0$, the randomised policy in Eq.~\eqref{eq: rand_CP} corresponds to the pseudo-deterministic {\it CP-Bandits}.

The exploration parameter $\epsilon$ may be kept constant over time, resembling a regularisation trick similar to the fixed probability clipping illustrated in Section~\ref{sec: notation}. In this work, we adopt a refined approach which allows $\epsilon$ to decay as information is accumulated, i.e., $t$ grows. Following~\cite{hadad_PNAS}, we formalise this relationship through an exponentially decaying exploration parameter, that is,
\begin{align} \label{eq: eps_t}
\epsilon_t \;=\; \frac{1}{K-1}\, t^{-\gamma},
\end{align}
where $K$ denotes the number of arms and $\gamma > 0$ controls the rate at which exploration vanishes. This ensures sufficiently rich exploration at the beginning while guaranteeing that the policy becomes increasingly exploitative as $t$ grows.
\begin{remark} \label{rk: exploration}
    In stationary environments, a decaying schedule is generally preferable: randomisation is most beneficial during the early learning phase (e.g., warm-up), while continued exploration becomes unnecessary, and potentially negative in view of regret minimisation, once the algorithm has effectively identified the optimal arm. In contrast, non-stationary reward settings might benefit from a constant or renewed exploratory phases, allowing dynamic adaptation to time-varying optimal arms. This concept will become more evident in the application study in Section~\ref{sec: application}.
\end{remark}

The pseudocode for implementing both deterministic and randomised conformal bandit policies is reported in \nameref{alg:conformal_bandits}. We set the warm-up phase to the minimum amount of information required to implement CP. This corresponds to two observations per arm, for a total of $t = 2K$ rounds, before a prediction interval can be constructed: one sample to fit the underlying quantile model (training set) and one sample for the conformal quantile (calibration set).

\begin{figure}[ht]
\centering
\fbox{
\parbox{0.95\linewidth}{
\textbf{Algorithm 1: (Randomised) Conformal Bandits}

\medskip
\textbf{Input:} Number of arms $K$, horizon $T$, conformal prediction strategy (e.g., CQR), miscoverage $\alpha \in (0,1)$,\\
\phantom{\textbf{Input:}} exploration parameter $\epsilon_t \in [0,1]$ (e.g., as a decreasing function of $t$; see Eq.~\eqref{eq: eps_t}), $\lambda \in [0,1].$\\
%\phantom{\textbf{Input:}} {\it conformal index} $I \in$ \{\eqref{eq:cp_ucb}, \eqref{eq:cp_blend}\} \\

\textbf{Output:} Selected arms $\{a^*_t\}_{t=1}^T$.
\smallskip

\begin{tabbing}
\hspace*{0.5cm}\=\hspace*{0.7cm}\=\hspace*{0.7cm}\=\kill

\ 1. \textbf{for} $t = 1$ \textbf{to} $2K$ \textbf{do} \` (Pure exploration – warm-up)\\
\ 2.\> Select arm $a^*_t = ((t-1) \bmod K) + 1$.\\
\ 3.\> Observe associated state-reward pair $(X_{t}, Y_{a^*_t,t})$.\\% and update train and calibration sets of arm $I_t$.\\
\ 4. \textbf{end for}\\[4pt]

\ 5. \textbf{for} $t = 2K+1$ \textbf{to} $T$ \textbf{do} \` (Conformal bandit selection)\\[2pt]
\ 6.\>\> \textbf{for} $k = 1,\dots,K$ \textbf{do}\\
\>\> Given observed state-reward pairs $(X_i, Y_{k,i})_{i = 1}^{N_{k,t-1}}$:\\
\ 7.\>\>\> \ \ \ Compute conformal intervals $\mathcal{C}_{k,t}^{1-\alpha} = \Big[L_{k,t},\, U_{k, t}\Big]$;\\
\ 8.\>\>\> \ \ \ Compute pseudo-deterministic conformal index $I_{k,t} = (1-\lambda)U_{k,t}+\lambda L_{k,t}$.\\
\ 9.\>\> \textbf{end for}\\[2pt]
10.\>\> Get pseudo-deterministic optimal arm $\tilde{a}_t^{*} = \arg\max_{k \in \mathcal{A}} I_{k,t}.$\\[4pt]

11.\>\> Draw $Z_t \sim \mathrm{Bernoulli}(1-\epsilon_t)$. \` (Bandit randomisation)\\
12.\>\> \textbf{if} $Z_t = 1$ \textbf{then}\\
13.\>\>\>$a_{t}^{*} = \tilde{a}_t^{*}$\\
14.\>\> \textbf{else}\\
15.\>\>\> $a_{t}^{*} \sim \text{Unif}_{\mathcal{A} \setminus \tilde{a}_t^{*}}$.\\
16.\>\> \textbf{end if}\\
17.\>\> Observe associated reward $Y_{a^*_t,t}$ and update state-reward pairs $(X_i, Y_{k,i})_{i = 1}^{N_{k,t}}$ for $k = a^*_t$.\\
18.\> \textbf{end for}\\[2pt]
19. \textbf{return} selected arms $\{a^*_t\}_{t=1}^T$.
\end{tabbing}
}
}
\captionsetup{labelformat=empty}
\caption{Algorithm 1: Conformal bandits}
\label{alg:conformal_bandits}
\end{figure}

\section{Simulation Studies}\label{sec: Simulation Studies}

Empirical performances of the proposed {\it Conformal Bandit} frameworks are assessed in specific small-gap and low-SNR regimes of interest, with $\Delta_k = \mu^*-\mu_k \in \{0.001, 0.005, 0.01, 0.05\}$ for all $k$'s and assuming an optimal arm $k^*$ exists and is unique. 
These values are motivated by the financial application considered in Section~\ref{sec: application}, where arm gaps vary over time around those ranges, as shown in Table~\ref{tab:arm_logreturn_statistics} and Panel~(b) of Figure~\ref{fig:cum_returns_and_absolute_gaps}. 
We consider a non-contextual stochastic $K$-armed bandit, with $K=3$, and independent and identically distributed (IID) arm rewards. The set of simulation studies are conducted under three different distributional regimes, with parameters reflecting real-world financial asset properties. All simulations (and empirical analysis presented in Section~\ref{sec: application}) were conducted in Python (with codes available for reproducibility on Github: \url{https://github.com/simonecuonzo/Conformal-bandits}). 

\begin{description}
\item[\textit{Gaussian rewards:}]
$Y_{k^*,t} \sim \mathcal{N}(\mu^*,\sigma^2)$, with $\mu^* = \mu_{k^*} \in \{0.001, 0.005, 0.01, 0.05\}$ and $\sigma = 0.01$, whereas $Y_{k,t}\sim \mathcal{N}(\mu_k,\sigma^2)$, with $\mu_k = 0$, for all $k \neq k^*$, for all $t$'s. %Mean and variance values reflect those of the real data analysed in Section~\ref{sec: application}.\\

\item[\textit{Student-$t$ rewards:}]
$Y_{k^*,t} = \mu^* + \sigma T_{t}$ and $Y_{k,t} = \sigma T_{t}$ for $k \neq k^*$, for all $t$'s; here, $T_{t} \sim t_{\nu}$ denotes a standard Student-$t$ random variable with $\nu = 3$ degrees of freedom. Mean and standard deviation parameters are as above. This distribution allows to investigate the impact of heavy-tailed behaviours, typical of financial data.

\item[\textit{Skewed Student-$t$ rewards:}] $Y_{k^*,t} = \mu^* + \sigma S_{\lambda_{k^*},t}$ and $Y_{k,t} = \sigma S_{\lambda_k,t}$ for $k \neq k^*$, with $\lambda_{k^*} = 0.3$, and $\lambda_k \in \{ -0.5, 0.6\}$ to account for both positive and negative skewness; here, $S_{\lambda,t} \sim \text{Skew-}t_{\nu, \lambda}$ denotes a Skew-$t$ distribution with skewness parameter $\lambda$ and $\nu = 3$ degrees of freedom. Mean and standard deviation parameters are as above. This scenario allows to simultaneously evaluate heavy tails and distributional asymmetry in the rewards, both present in financial environments due to the presence of extreme events and the leverage effect~\citep{engle1993measuring}. This represents a phenomenon where negative shocks (bad news) tend to increase volatility more than positive shocks (good news).
\end{description}
Each scenario is assessed in Monte Carlo (MC) studies of size $M = 1,000$, % generate multiple Monte Carlo (MC) replications and evaluate 
in terms of both bandit performance metrics, which quantify exploration–exploitation efficiency, and CP metrics, which assess the properties of the derived uncertainty intervals. Specifically, the following measures are considered.
\begin{description}
\item[\textit{Cumulative Regret:}] A central measure of performance in sequential decision-making, which quantifies the loss incurred by not selecting the optimal arm. We evaluate it through its empirical mean across the $M$ MC replications as:
\begin{equation}
\hat{R}_t = \frac{1}{M} \sum_{m=1}^{M}
\frac{1}{t}\sum_{i=1}^{t}
\left( \mu^{*} - \mu_{a_t^{(m)}}\right),
\qquad t = 1,\ldots,T,
\label{eq:pseudoregret}
\end{equation}
where $\mu^* = 0.05$ is the mean reward of the optimal arm, and 
$\mu_{a_t^{(m)}}$ denotes the mean reward associated with the arm selected at time $t$ in the $m$-th replication. Regret trajectories $\{R_t\}_{t=1}^{T}$ provide insight into the learning ability and speed of a bandit policy.

\item[\textit{Best-arm Selection:}] To quantify the convergence toward the optimal arm, we compute the average cumulative proportion of times the optimal arm is selected over $t$ rounds, that is,
\begin{equation}
\hat{\pi}_{k^*, t} = %\frac{\widehat{\mathbb{E}}(N_{k^*, t})}{t} = 
\frac{1}{M} \sum_{m=1}^{M}
\frac{1}{t} \sum_{i=1}^{t} 
\mathbb{I}({a^{(m)}_i = k^*}),
\qquad t = 1,\ldots,t.
\label{eq:bestarmshare}
\end{equation}
An increasing $\hat{\pi}_{k^*, t} \in [0,1]$ over time  indicates convergence toward the optimal arm $k^{*}$.

\item[\textit{Coverage:}] The main statistical validity property is the empirical coverage rate, given by the frequency with which the observed reward fall within its $1-\alpha$ conformal interval. For a horizon $T$, this is given by
\[
\widehat{\mathrm{COV}}^{1-\alpha}_{k}
=
\frac{1}{M} \sum_{m=1}^{M}
\frac{1}{T}\sum_{t=1}^{T}
\mathbb{I}({
Y^{(m)}_{k,t} \in \mathcal{C}_{k,t}^{1-\alpha,(m)}
}),\quad k=1,\dots,K,
\]
where $Y^{(m)}_{k,t}$ and $\mathcal{C}_{k,t}^{1-\alpha,(m)}$ are respectively the observed reward and the CP interval for arm $k$ at round $t$ in MC replication $m$.
%Coverage is first averaged across time within each Monte Carlo replication and then across replications.
A properly calibrated CP approach should achieve mean coverage close to the nominal target $(1-\alpha)$, even in finite-samples.

\item[\textit{Interval width:}] Letting the interval width or length at time $t$ for arm $k$ to be $\mathcal{L}_{k,t} = U_{k,t} - L_{k,t}$, we compute the mean interval width of arm $k$ as:
\[
\mathcal{L}_k = \frac{1}{M} \sum_{m=1}^{M} \frac{1}{T}\sum_{t=1}^{T} \mathcal{L}^{(m)}_{k,t}.
\]
Smaller widths correspond to more informative intervals, but must be evaluated jointly with coverage: sharp intervals with inadequate coverage indicate underestimation of uncertainty, while wide intervals with excessive coverage signal over-conservativeness.
Combined analysis of $\widehat{\mathrm{COV}}_k$ and $\mathcal{L}_k$ provides a comprehensive view of the coverage-width trade-off.
\end{description}

The bandit policies evaluated in our simulation studies consist of the novel policies characterising the {\it Conformal Bandit} framework, that is,
\begin{itemize}
    \item[(1)] \ref{eq:cp_ucb}, a fully reward-maximisation policy, equivalent to \ref{eq:cp_blend} with $\lambda = 0$;
    \item[(2)] \ref{eq:cp_blend} with $\lambda = 1$, a fully risk-protection policy;
    \item[(3)] \ref{eq:cp_blend} with $\lambda = 0.7$, a risk-aware policy, with greater emphasis on potential losses than rewards;
    \item[(4)] \ref{eq:cp_blend} with $\lambda = 0.5$, a risk-aware policy, with equal treatment of potential losses and rewards;
    %\item[(5)] \ref{eq:cp_esi}, using the novel {\it Exploratory Skewness Index} in Eq.~\eqref{eq:esi};
\end{itemize}
as well as: 
\begin{itemize}
    \item[(5)] the classical \ref{eq:ucb1}, illustrated in Section~\ref{sec: UCB1}; this represents our benchmark. For the UCB1 benchmark, we set \(\beta=4\), which under our parametrization recovers the canonical exploration term \(\sqrt{2\log(t)/N_i(t)}\). This form is widely used in the bandit literature and is known to provide a good trade-off between exploration and exploitation; see, e.g., \cite{auer2002finite,villar2015multiarmed}. The hyperparameter $c$ is informed by the application data in Table~\ref{tab:arm_logreturn_statistics} and is set to $c = 0.1$, reflecting the expected bounds for the reward data, with sample minimum and maximum of $a = -0.08$ and $b = 0.06$, respectively. %\textcolor{red}{[FORSE QUI SPIEGARE LEGGERMENTE MEGLIO perche abbiamo usato c=0.1?]}
\end{itemize}
All {\it Conformal Bandit} policies are based on CQR (see Section~\ref{sec:CQR}) and the {\it Adaptive Conformal Inference} correction~\citep{gibbs2021adaptive}. 

\begin{figure}[t]
\centering
\setlength{\tabcolsep}{6pt}

\begin{tabular}{cc}

% =====================================
% =========== Gaussian row ===========
% =====================================

% Gaussian Regret
\begin{subfigure}[t]{0.45\textwidth}
\centering
\caption*{Gaussian}
\includegraphics[width=\linewidth]{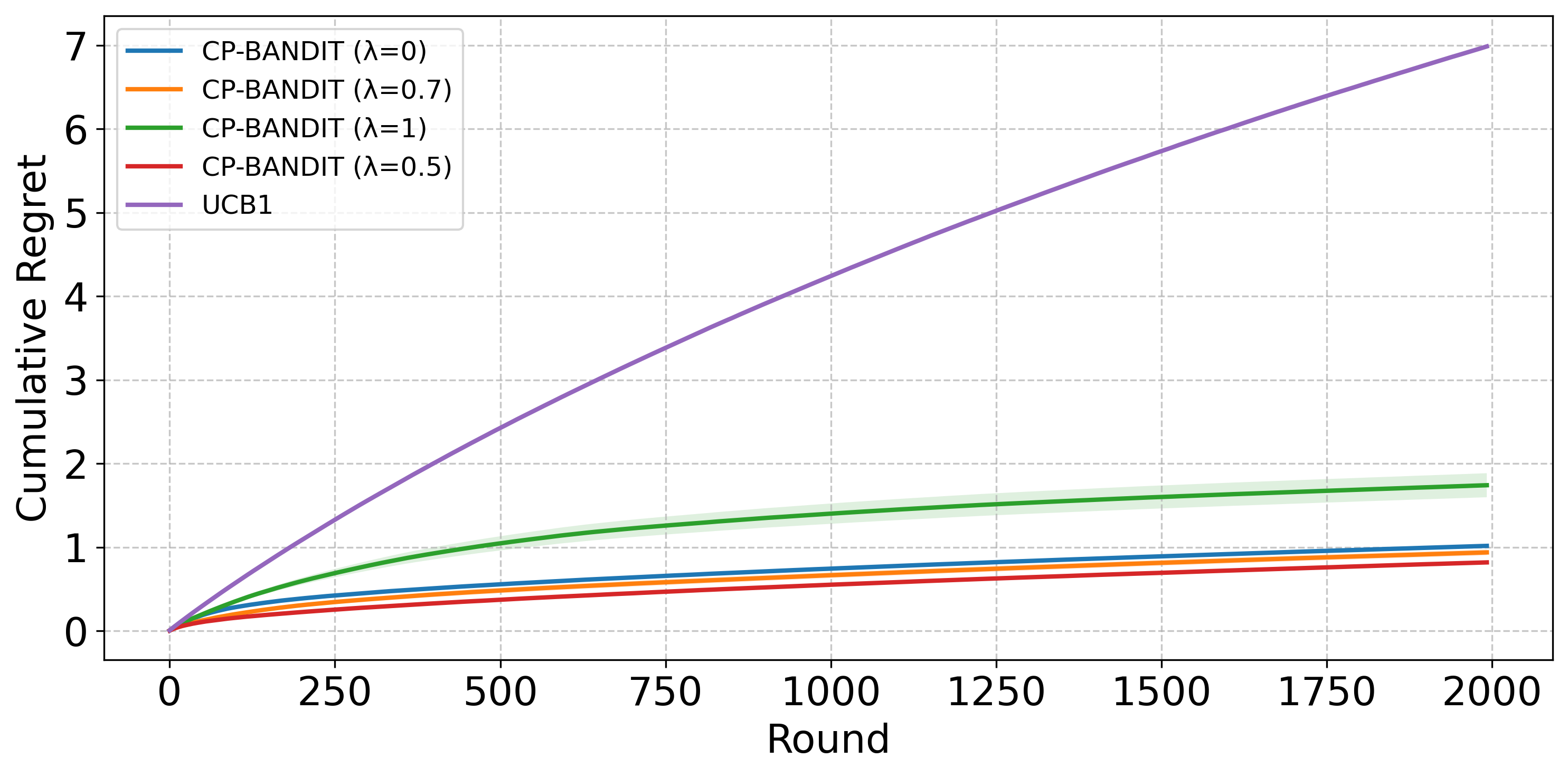}
\end{subfigure}
&
% Gaussian Best Arm
\begin{subfigure}[t]{0.45\textwidth}
\centering
\caption*{Gaussian}
\includegraphics[width=\linewidth]{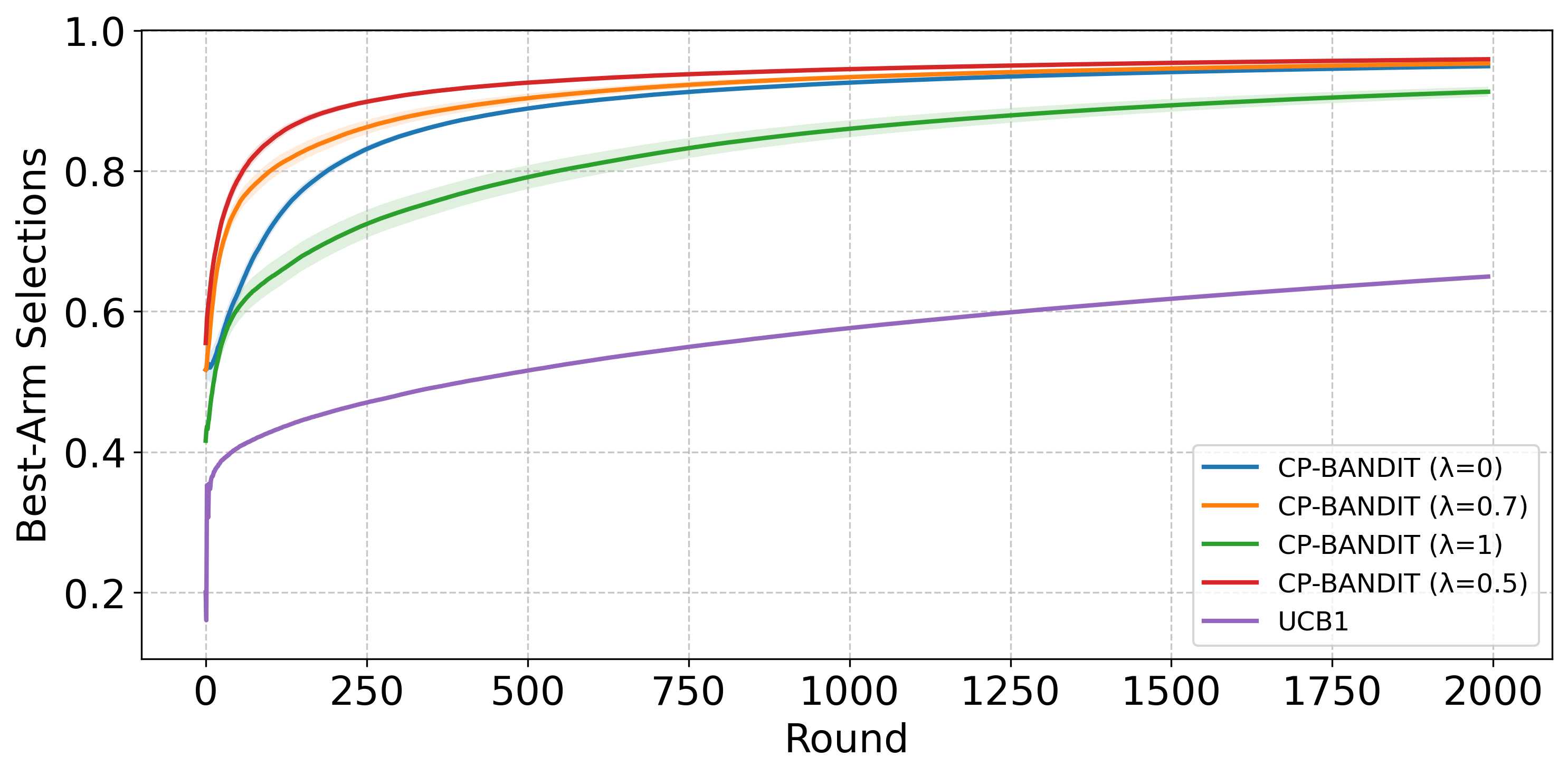}
\end{subfigure}
\\[1.1em]

% =====================================
% ========= Student-t row =============
% =====================================

% Student-t Regret
\begin{subfigure}[t]{0.45\textwidth}
\centering
\caption*{Student-$t$}
\includegraphics[width=\linewidth]{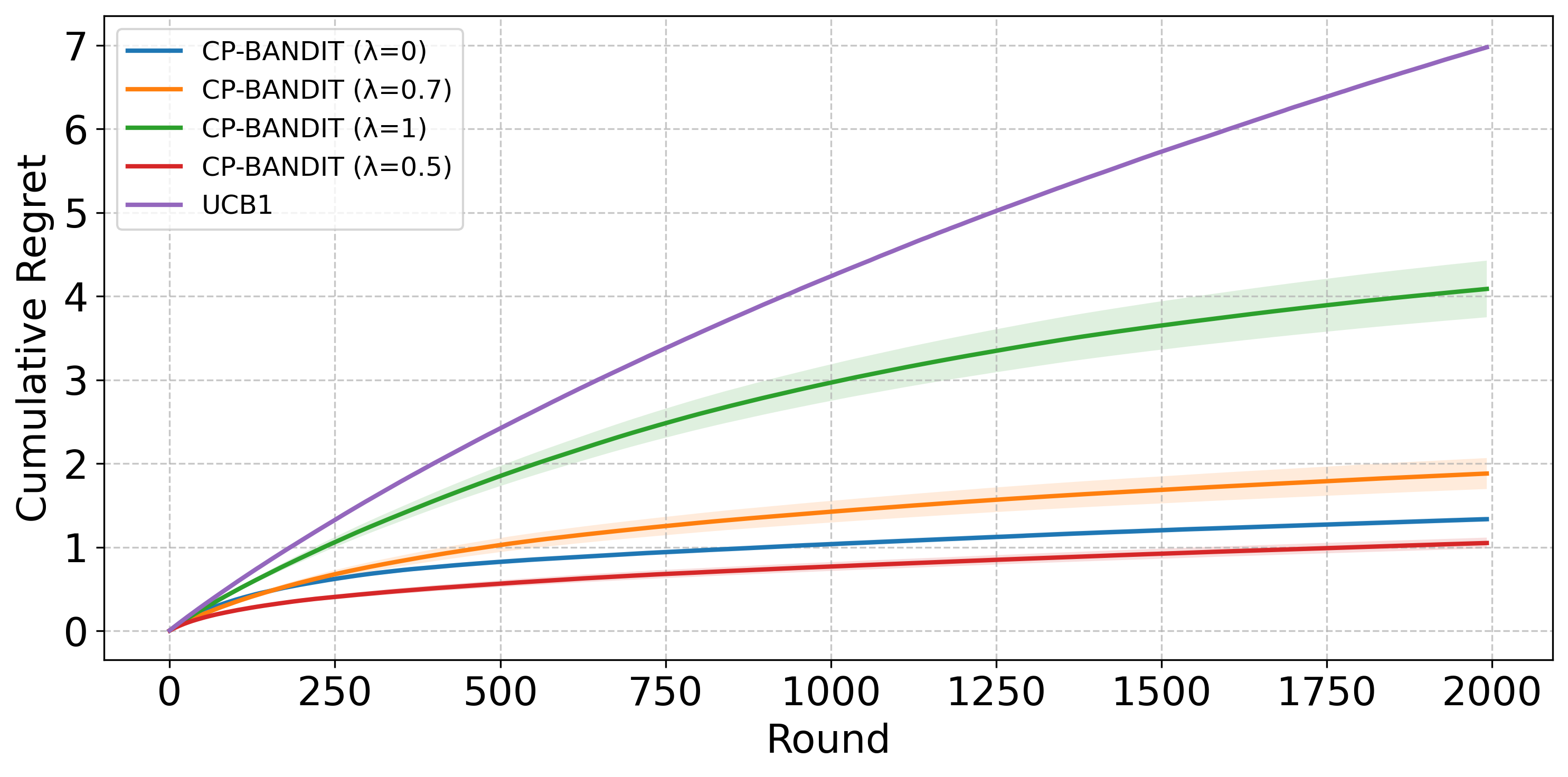}
\end{subfigure}
&
% Student-t Best Arm
\begin{subfigure}[t]{0.45\textwidth}
\centering
\caption*{Student-$t$}
\includegraphics[width=\linewidth]{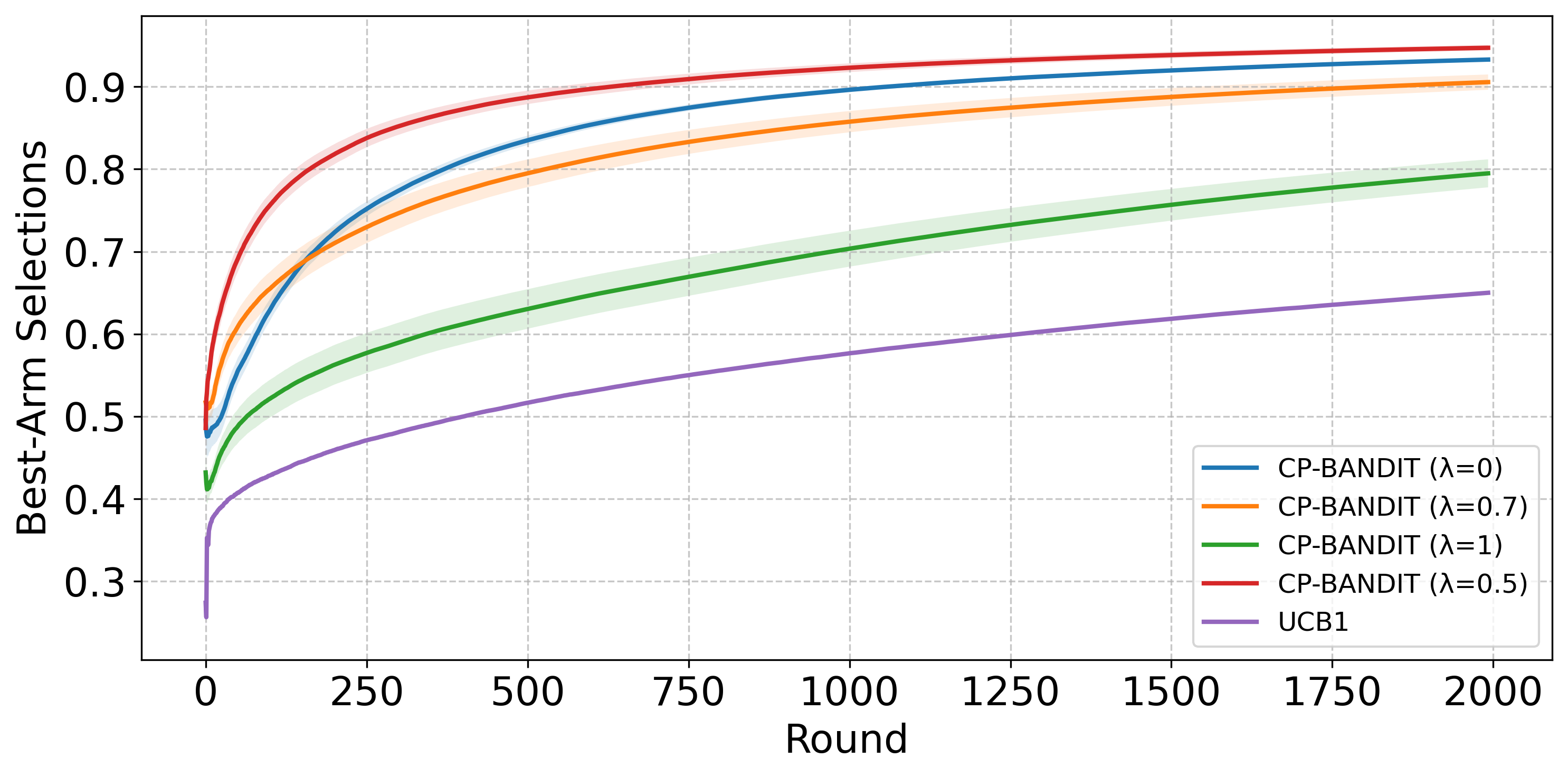}
\end{subfigure}
\\[1.1em]

% =====================================
% ========== Skew-t row ===============
% =====================================

% Skew-t Regret
\begin{subfigure}[t]{0.45\textwidth}
\centering
\caption*{Skew-$t$}
\includegraphics[width=\linewidth]{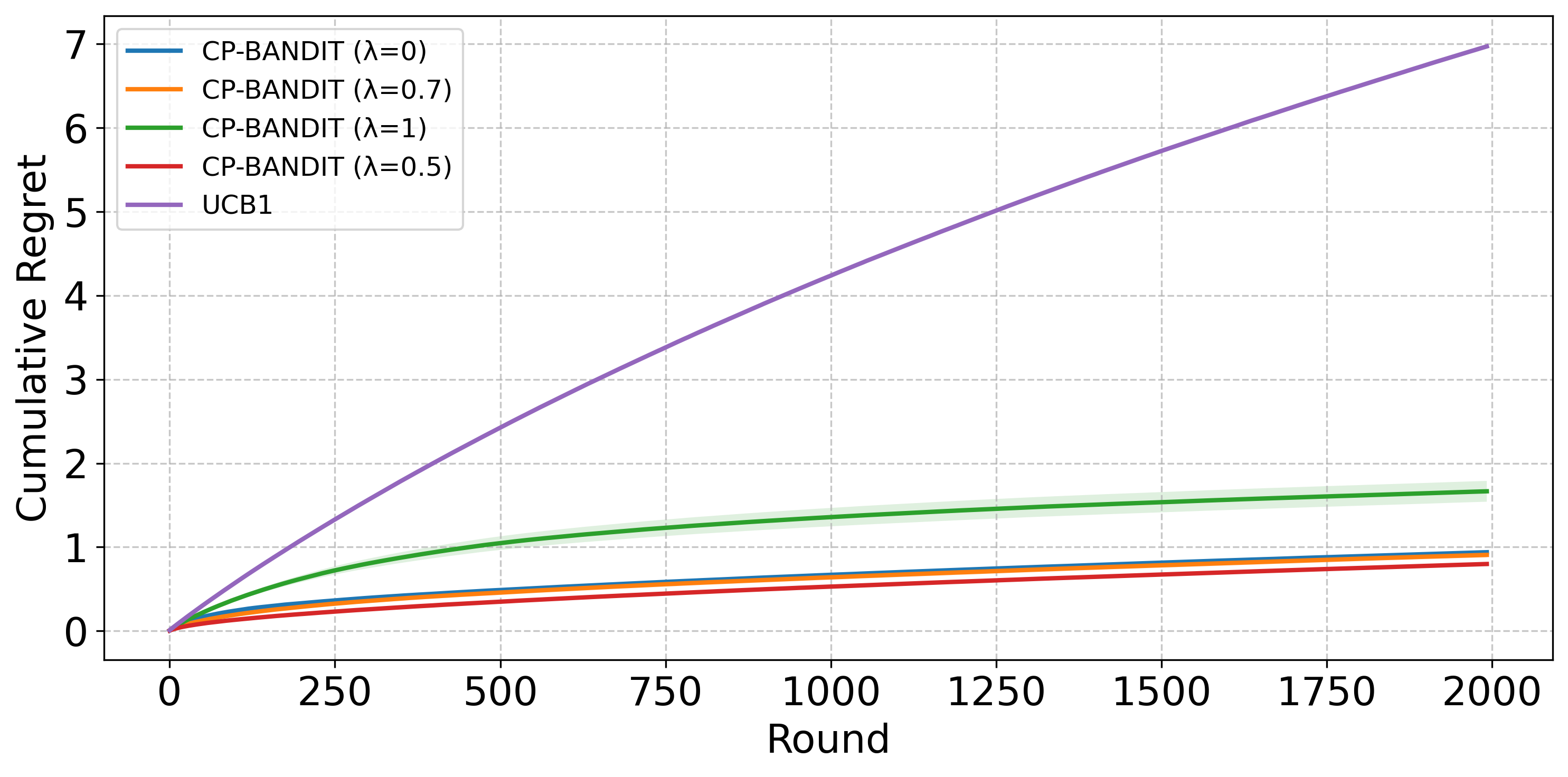}
\end{subfigure}
&
% Skew-t Best Arm
\begin{subfigure}[t]{0.45\textwidth}
\centering
\caption*{Skew-$t$}
\includegraphics[width=\linewidth]{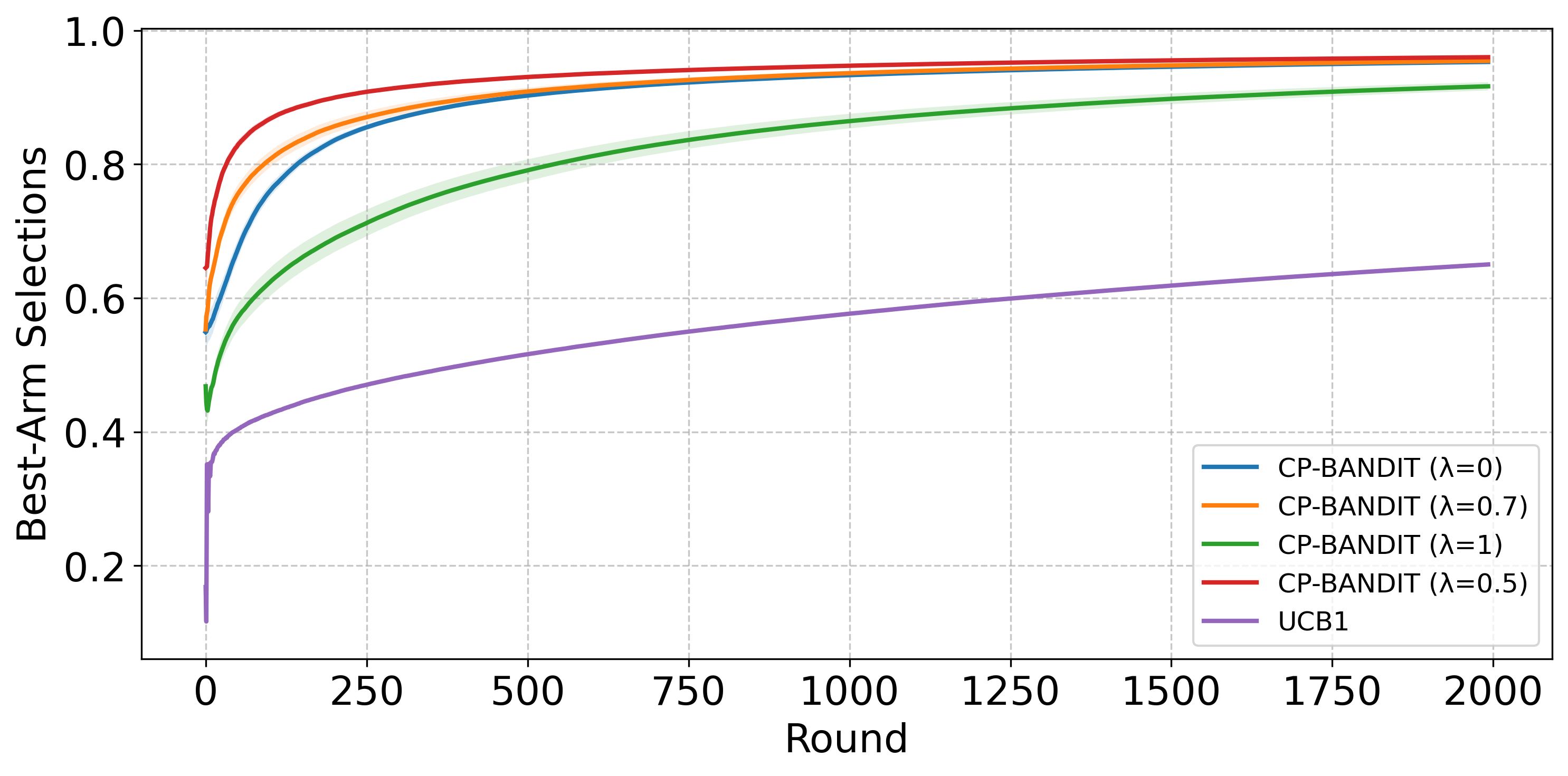}
\end{subfigure}

\end{tabular}

\caption{
Comparison between the Conformal Bandit variants and classical UCB1 in terms of cumulative regret and cumulative best-arm selection rates for the gap setting \(\Delta_k=0.01\), corresponding to \(\mu_1=\mu^*=0.01\) and \(\mu_2=\mu_3=0\), based on \(1000\) Monte Carlo simulations.
Rows correspond to different reward-generating environments: Gaussian, Student-$t$, and skew-$t$ with asymmetric tails.
The left column reports cumulative regret, while the right column reports cumulative best-arm selection rates.
Shaded regions represent $95\%$ Monte Carlo confidence intervals.
}
\label{fig:six_panel_0.01}
\end{figure}

The simulation results in terms of bandit and uncertainty performances are shown in Figure~\ref{fig:six_panel_0.01} and Table~\ref{tab:0.01}, respectively. The main paper only shows the results for the gap value $\Delta_k = 0.01$; results of the other gap values are deferred in the Appendix \ref{app:add_sim_stud}. First, Figure~\ref{fig:six_panel_0.01} shows that all {\it CP-Bandit} variants exhibit uniformly superior learning efficiency compared to \ref{eq:ucb1}, achieving both lower regret and higher best-arm selection rates across all distributional scenarios. Among the novel bandit policies, \ref{eq:cp_blend} ($\lambda = 0.5$) provides the best overall learning performances, demonstrating fast convergence toward selecting the optimal arm, and logarithmic regret even under heavy-tailed and skewed reward distributions. In contrast, an almost linear regret can be depicted for \ref{eq:ucb1}, %especially in Student-$t$ scenarios, 
confirming the challenge in small-gap regime. This challenge becomes even more evident when it comes to evaluating the statistical properties of the UCB bounds. In fact, Table~\ref{tab:0.01} highlights an overly conservative behaviour in the bounds constructions, with systematically wider intervals and over-coverage of future rewards. Across all scenarios, \ref{eq:cp_blend} variants achieve more informative (i.e., tighter) intervals, while preserving the nominal coverage $1-\alpha$, set to $0.8$ as a reasonable balancing value between statistical guarantees and reward gains. As expected, the under-explored arms are characterised by higher uncertainty (wider intervals) and slight overcoverage. Notably, \ref{eq:cp_blend} ($\lambda = 0.5$) again provides the best overall coverage performances in terms of interval coverage and width trade-off. We emphasise that, in contrast to the UCB1 policy (which fails to achieve nominal coverage in the Student-$t$ scenario for the optimal arm), the \ref{eq:cp_blend} framework leverages stronger statistical guarantees on this side. For \(\Delta=0.05\) (see Table~\ref{tab:0.05} in Appendix~\ref{app:add_sim_stud}), \ref{eq:ucb1} exhibits empirical coverage substantially below the nominal target in the Student-\(t\) setting, with coverage equal to \(71.42\%\).
%For $\Delta = 0.05$ (see Table \ref{tab:0.05} in the Appendix~\ref{app:add_sim_stud}), \ref{eq:ucb1} is characterised by reduced coverage with values $\approx 70\%$ for the optimal arm in the Student-$t$ scenario.

\begin{table*}[t]
\centering
\scriptsize
\renewcommand{\arraystretch}{1.25}
\setlength{\tabcolsep}{4pt}
\caption{
Comparison between the Conformal Bandit variants and classical UCB1 in terms of coverage (\%) and mean interval width for the gap setting \(\Delta_k=0.01\), corresponding to \(\mu_1=\mu^*=0.01\) and \(\mu_2=\mu_3=0\), across all reward-generating scenarios: Gaussian, Student-\(t\), and Skew-\(t\).
All results are averaged across $1,000$ MC replicates. Standard deviations across Monte Carlo runs are reported in parentheses.}
\label{tab:0.01}
\begin{tabularx}{\textwidth}{l | *{3}{X} | *{3}{X} | *{3}{X}}
\toprule
& \multicolumn{3}{c|}{\textbf{Gaussian}}
& \multicolumn{3}{c|}{\textbf{Student-$t$}}
& \multicolumn{3}{c}{\textbf{Skew-$t$}} \\
\textbf{Algorithm}
& Arm1 & Arm2 & Arm3
& Arm1 & Arm2 & Arm3
& Arm1 & Arm2 & Arm3 \\
\midrule
\multicolumn{10}{c}{\textbf{Coverage (\%)}} \\
\midrule

CP-Bandit ($\lambda=0$) 
& 80.13 (0.21) & 88.55 (3.57) & 88.69 (3.82)
& 80.11 (0.21) & 86.79 (3.33) & 86.67 (3.36)
& 80.10 (0.20) & 87.73 (4.13) & 87.85 (3.73) \\

CP-Bandit ($\lambda=0.7$) 
& 80.12 (0.22) & 89.43 (4.27) & 89.25 (4.18)
& 80.22 (1.11) & 87.14 (4.57) & 86.99 (4.59)
& 80.10 (0.20) & 88.23 (4.22) & 88.07 (4.33) \\

CP-Bandit ($\lambda=1$) 
& 80.17 (0.89) & 88.62 (4.72) & 88.72 (4.73)
& 80.60 (2.01) & 86.51 (4.78) & 86.40 (4.74)
& 80.12 (0.43) & 87.61 (4.32) & 87.42 (4.54) \\

CP-Bandit ($\lambda=0.5$) 
& 80.12 (0.21) & 89.45 (4.29) & 89.41 (4.34)
& 80.12 (0.48) & 87.45 (4.38) & 87.35 (4.28)
& 80.10 (0.20) & 88.12 (4.29) & 88.48 (4.11) \\

UCB1
& 86.51 (0.95) & 98.91 (0.57) & 98.94 (0.56)
& 78.12 (1.22) & 93.16 (1.46) & 93.10 (1.47)
& 92.65 (0.75) & 98.33 (0.65) & 98.36 (0.70) \\

\midrule
\multicolumn{10}{c}{\textbf{Mean Interval Width}} \\
\midrule

CP-Bandit ($\lambda=0$) 
& 0.0260 (0.0006) & 0.0365 (0.0046) & 0.0365 (0.0047)
& 0.0336 (0.0010) & 0.0491 (0.0097) & 0.0488 (0.0107)
& 0.0188 (0.0006) & 0.0276 (0.0072) & 0.0263 (0.0065) \\

CP-Bandit ($\lambda=0.7$) 
& 0.0260 (0.0006) & 0.0385 (0.0061) & 0.0385 (0.0065)
& 0.0340 (0.0047) & 0.0538 (0.0160) & 0.0527 (0.0147)
& 0.0188 (0.0006) & 0.0288 (0.0084) & 0.0282 (0.0158) \\

CP-Bandit ($\lambda=1$) 
& 0.0261 (0.0017) & 0.0378 (0.0066) & 0.0379 (0.0070)
& 0.0357 (0.0087) & 0.0532 (0.0172) & 0.0522 (0.0152)
& 0.0189 (0.0014) & 0.0284 (0.0084) & 0.0276 (0.0152) \\

CP-Bandit ($\lambda=0.5$) 
& 0.0260 (0.0006) & 0.0385 (0.0062) & 0.0383 (0.0062)
& 0.0336 (0.0016) & 0.0544 (0.0153) & 0.0543 (0.0149)
& 0.0188 (0.0006) & 0.0291 (0.0088) & 0.0281 (0.0105) \\

UCB1
& 0.0368 (0.0004) & 0.0657 (0.0015) & 0.0658 (0.0015)
& 0.0368 (0.0006) & 0.0659 (0.0024) & 0.0659 (0.0026)
& 0.0368 (0.0003) & 0.0658 (0.0015) & 0.0658 (0.0014) \\

\bottomrule
\end{tabularx}
\end{table*}

\section{Small-Gap Regime Case Study: {\it \bf Conformal Bandits} for Portfolio Allocation} \label{sec: application}

For the application of interest, we consider a discrete-time self-financing investment environment over a finite horizon $T$.
Trading occurs at discrete times 
$t_j = j \, \Delta t$, for $j = 0, \ldots, m$, 
where $\Delta t$ denotes the rebalancing interval (here, one day), and $m$ is the total number of trading periods.
At each date $t_j$, the investor allocates capital across $n$ risky assets whose prices are denoted by $P_{j,i}$, $i = 1, \ldots, n$.
Let $R_j = (R_{j,1}, \ldots, R_{j,n})^\top$ denote the vector of log returns from $t_{j-1}$ to $t_j$, defined as
$R_{j,i} = \log{(P_{j,i}/P_{j-1,i})}$ so that $R_{j,i}$ represents the logarithmic change in the price of asset $i$ over period $\Delta_j= t_j-t_{j-1}$. Transaction costs and dividends are ignored in this baseline formulation to isolate the effect of the allocation dynamics, although in practice they can be incorporated as extensions. At each rebalancing time $t_j$, the investor selects a weight vector $W_j = (w_{j,1}, \ldots, w_{j,n})^\top$, where $w_{j,i}$ denotes the proportion of wealth allocated to asset $i$ at time $t_j$, satisfying $W_j^\top \mathbf{1} = 1$, where $\mathbf{1}$ denotes the $n$-dimensional vector of ones. Positive weights ($w_{j,i} > 0$) correspond to long positions, while negative weights ($w_{j,i} < 0$) represent short positions.

Rather than treating each individual asset as a separate arm in the MAB formulation, we define the arms as different portfolio strategies. This aggregation substantially reduces the dimensionality of the decisions (i.e., the number of arms), lowers the trading turnover, and enhances stability. In this work, we consider three well-established portfolio strategies, represented by:
\begin{description}
    \item[\textit{Sell-All} (SA):] a full liquidation strategy that holds no risky assets, i.e., a pure cash or neutral position, with $W_j^{\mathrm{SA}} = 0$;
    
    \item[\textit{Equally-Weighted} (EW):] an allocation strategy where all assets receive identical portfolio weights at each rebalancing period, with $W_j^{\mathrm{EW}} = \frac{1}{n}\mathbf{1}$;
    
    \item[\textit{Mean–Variance} (MV):] an active allocation strategy based on the Markowitz mean-variance framework~\citep{markowitz1952modern}, which balances expected return and risk as
    \begin{equation}
    W_j^{\mathrm{MV}} = 
    \arg\min_{W_j^\top \mathbf{1} = 1}
    \big( W_j^\top \Sigma_j W_j - R_j^\top W_j \big),
    \label{eq:mv}
    \end{equation}
    with $\Sigma_j$ being the covariance matrix of asset returns at time $t_j$, and $R_j^\top W_j$ the expected portfolio return.
\end{description}
Therefore, at each rebalancing period $t_j$, the agent selects an arm $A_j \in \mathcal{A} = \{W_j^{\mathrm{SA}}, W_j^{\mathrm{EW}},  W_j^{\mathrm{MV}}\}$ and generates a stochastic reward $Y_{a_j,j} = a_j^\top R_j$, representing the realised return of the portfolio selected at period $t_j$. 

All portfolios are constructed using a set of diversified exchange-traded funds (ETFs) that span multiple asset classes and economic sectors, including commodities, health care, technology, real estate, and both equity and fixed-income markets. 
Table~\ref{tab:etfs} summarises the ETFs used in the portfolio constructions and highlights the corresponding asset classes. We consider daily data from January 2018 to February 2025, publicly available from Yahoo Finance (Github repository: \url{https://github.com/simonecuonzo/Conformal-bandits}). Descriptive statistics of the data are provided in Table~\ref{tab:arm_logreturn_statistics}. %, reporting the sample mean and sample standard deviation (SD) of the reward associated with each of the three arms, represented by the daily log-return of each portfolio. %The MV portfolio exhibits a lower standard deviation than the EW portfolio, indicating lower risk, at the cost of a lower average return.%\todo{A summary of the data, reflecting the time-varying gaps of the considered arms along with their volatility is reported in Figure XX.}
\begin{table}[ht]
\centering
\begin{tabularx}{\textwidth}{X|X}
\hline
\textbf{Ticker} & \textbf{Asset class / sector} \\ \hline
TLT  & US Government Bonds             \\
GLD  & Commodity (Gold)                \\
VNQ  & US Real Estate                  \\
EFA  & International Equity (Developed Markets ex-US) \\
VOO  & US Equity (Broad Market)        \\
EMB  & Emerging Market Bonds           \\
DBC  & Broad Commodities               \\
XLF  & US Equity -- Financials         \\
XLK  & US Equity -- Technology         \\
XLV  & US Equity -- Health Care        \\ \hline
\end{tabularx}
\caption{
ETFs (identified by their tickers) used in the portfolio strategies. The label ``ex-US'' indicates that US-listed equities are excluded.
}
\label{tab:etfs}
\end{table}

\begin{table}[htbp]
    \centering
    \caption{
Portfolio arms and summary statistics–sample mean, standard deviation (SD), minimum, and maximum over the evaluation period–of the associated rewards, represented by the daily log-return.}
    \label{tab:arm_logreturn_statistics}
    \small
    \begin{tabular}{llcccc}
        \toprule
        Arm name & Portfolio allocation strategy & Sample mean & Sample SD & Min & Max \\
        \midrule

        MV 
        & \(W_j^{\mathrm{MV}} =
        \displaystyle \arg\min_{W_j^\top \mathbf{1}=1}
        \left( W_j^\top \Sigma_j W_j - R_j^\top W_j \right)\) 
        & 0.0001 
        & 0.0056 
        & -0.0587 
        & 0.0326 \\

        SA 
        & \(W_j^{\mathrm{SA}} = 0\) 
        & 0.0000 
        & 0.0000 
        & 0.0000 
        & 0.0000 \\

        EW 
        & \(W_j^{\mathrm{EW}} = \frac{1}{n}\mathbf{1}\) 
        & 0.0003 
        & 0.0083 
        & -0.0852 
        & 0.0631 \\

        \bottomrule
    \end{tabular}
\end{table}

\begin{remark}
In this work, aligned with existing literature~\citep{shen_wang_jiang_zha2015, shen_wang2016_portfolio_blending, huo_fu2017_riskaware, fujishima_nakagawa2022, hu2025multi}, we focus on bandit policies operating under a partial-information framework where only the selected arm's reward is observed. However, in practice, for the problem at hand, all conformalized policies can also be deployed in a full-information setting, where at each round $t$ the agent observes all arm rewards, not only that of the selected arm. In this scenario, exploration becomes less critical since its fundamental role is learning about under-explored arms, rendering the randomised CP-bandit version in Eq.~\eqref{eq: rand_CP} of limited practical benefit. In principle, access to complete reward information should lead to important improvements over the standard partial-information setting. A dedicated investigation for the CP extensions, reported in the Appendix \ref{app:full_information}, shows no substantial gains, aligning with existing work~\citep{botosan2024optimal}. %Nonetheless, when more sophisticated regime-aware CP policies are considered (see Section~\ref{sec: app_HMM}), the gain is more relevant. 
We also note that operating under full-information may come at the cost of increased computational burden, as incorporating all available rewards requires additional processing at each round. A natural question, therefore, is whether the performance gains justify the higher computational requirements. %\textcolor{red}{[NINA: IN realta con i nuovi risultati le performances di CP-UCB full info aumentano considerevolmente rispetto a quello partial info (giarda tabelle di performance finanziarie + grafici]}

\end{remark}

\subsection{An Extension to Regime-Aware Bandit Policies} \label{sec: app_HMM}

Financial markets exhibit pronounced regime-dependent behaviour, characterised by alternating periods of expansion, contraction, and structural instability. These phases differ not only in terms of expected returns, but also in volatility, tail risk, cross-asset dependence, and reward asymmetry. A policy that ignores regime information may conflate fundamentally heterogeneous market conditions, leading to suboptimal decisions. For example, in bullish or neutral environments, rewards can be more stable and positively skewed, making optimistic selection rules appropriate. In contrast, bearish regimes tend to exhibit higher volatility, heavier tails, and negative asymmetry, calling for more conservative, downside-protective actions.

This observation naturally suggests the need for \textit{regime-aware bandit policies}, in which the allocation rule adjusts dynamically to the prevailing market conditions. Rather than applying a single exploration–exploitation mechanism uniformly across all periods, the policy explicitly conditions its behaviour on the underlying market phase, as inferred through a regime-switching model. In our application, we classify regimes into three economically interpretable states: \texttt{Bull} (calm), \texttt{Neutral} (relatively calm), and \texttt{Bear} (turbulent). Their identification is performed using HMMs~\citep{zucchini_macdonald_langrock2016} fitted to past S\&P500 returns. The characterisation of the HMM structure is provided in the Appendix \ref{app:HMM}, together with the procedural descriptions of the proposed regime-aware policies, illustrated below (Appendix \ref{app:Algorithms' description}). 

% Such data cover multiple macro–financial regimes including bull markets, periods of heightened volatility, and episodes of severe drawdowns (e.g., the COVID-19 crash).

\subsubsection{Regime-Aware Conformal Bandits}
Using the CP bounds introduced in Section~\ref{sec: CP-bandits}, we define a regime-dependent action rule that favours upside exploitation in calm regimes and downside protection during distressed periods; specifically,
\begin{equation}
a_j =
\begin{cases}
\arg\max_{k \in \mathcal{A}} U_{k,t_j}, & \text{if } t_j \in \{\texttt{Bull}, \texttt{Neutral}\},\\[4pt]
\arg\max_{k \in \mathcal{A}} L_{k,t_j}, & \text{if } t_j \in \{\texttt{Bear}\},
\end{cases}
\label{eq:cp_regimeaware}
\end{equation}
where $U_{k,t_j}$ and $L_{k,t_j}$ denote the upper and lower CP bounds for arm $k$ at time $t_j$.
In expansionary regimes, the algorithm behaves \textit{opportunistically} or \textit{optimistically}, selecting the arm with the highest upper bound. During downturns, the policy becomes \textit{defensive}, choosing the arm with the least severe lower predictive bound (largest $L_k(t_j)$), thereby, minimising potential losses. This reflects the asymmetric risk attitudes typically seen in financial decision-making.
This is equivalent to alternating policies~\ref{eq:cp_ucb} and \ref{eq:cp_blend} according to the inferred regime. 

\subsubsection{Regime-Aware MV-UCB1}
Given its centrality in finance, we also extend the mean-variance extension of UCB1 in Eq.~\eqref{eq:mvucb1} to regime-aware setting, and use it as a comparator. The idea is to use the classical empirical mean-driven rule during stable regimes, and to switch to a volatility-adjusted one when the market enters a distressed state. Formally,
\begin{equation}\label{eq:Regime-Aware_MV-UCB1}
a_j =
\begin{cases}
\displaystyle 
\arg\max_{k \in \mathcal{A}}
\Big\{\widehat{\mu}_{k,t_j}
+ c\sqrt{\frac{\beta\log t_j}{2N_{k,t_j}}}
\Big\}, 
& \text{if } t_j \in \{\texttt{Bull}, \texttt{Neutral}\},\\[10pt]
\displaystyle \arg\max_{k \in \mathcal{A}}
\Big\{\widehat{MV}^{\rho}_{k,t_j} + c\sqrt{\frac{\beta\log t_j}{2N_{k,t_j}}}
\Big\}, 
& \text{if } t_j \in \{\texttt{Bear}\},
\end{cases}
\end{equation}
where $\widehat{MV}^{\rho}_{k,t_j}
= \rho \, \widehat{\mu}_{k,t_j} - (1 - \rho)\, \widehat{\sigma}_{k,t_j}$ is the empirical mean-variance score of arm $k$. In bullish or neutral conditions, the policy relies on pure mean-based optimism. In bearish regimes, the algorithm shifts toward a \textit{risk-adjusted} score that penalises volatility, in line with financial portfolio theory and with the higher downside risk observed during market contractions. Again, we set \(\beta=4\) and \(c=0.1\), consistently with the values adopted in the simulation study.

To accommodate potential non-stationarities in the real data, all {\it Conformal Bandits}, including the regime-aware ones, employ a fixed exploration threshold $\epsilon = 0.03$, ensuring that the bandit maintains sufficient adaptability to distributional shifts over time (see Remark~\ref{rk: exploration}). Further, they are all complemented by the ACI scheme of~\cite{gibbs2021adaptive} to address non-exchangeability; baseline miscoverage level is set to $\alpha=0.2$. For benchmarking purposes, we additionally report the cumulative returns of the three evaluated portfolios: SA, EW, MV.

\subsection{Results and Discussion} 
\begin{figure}[p]
    \centering

    \textbf{(a)}\par
    \includegraphics[width=0.95\textwidth]{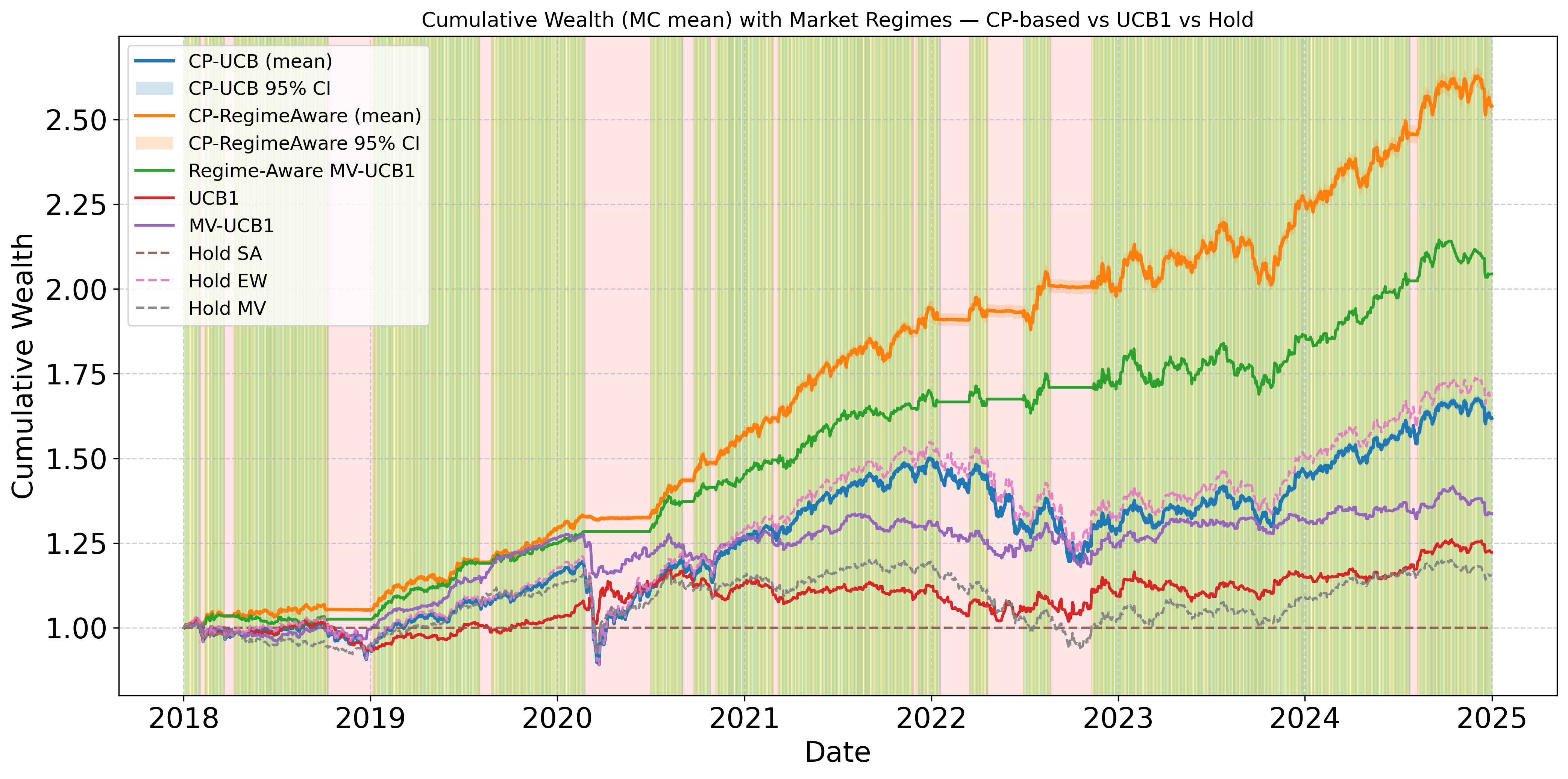}

    \vspace{0.4cm}

    \textbf{(b)}\par
    \includegraphics[width=0.95\textwidth]{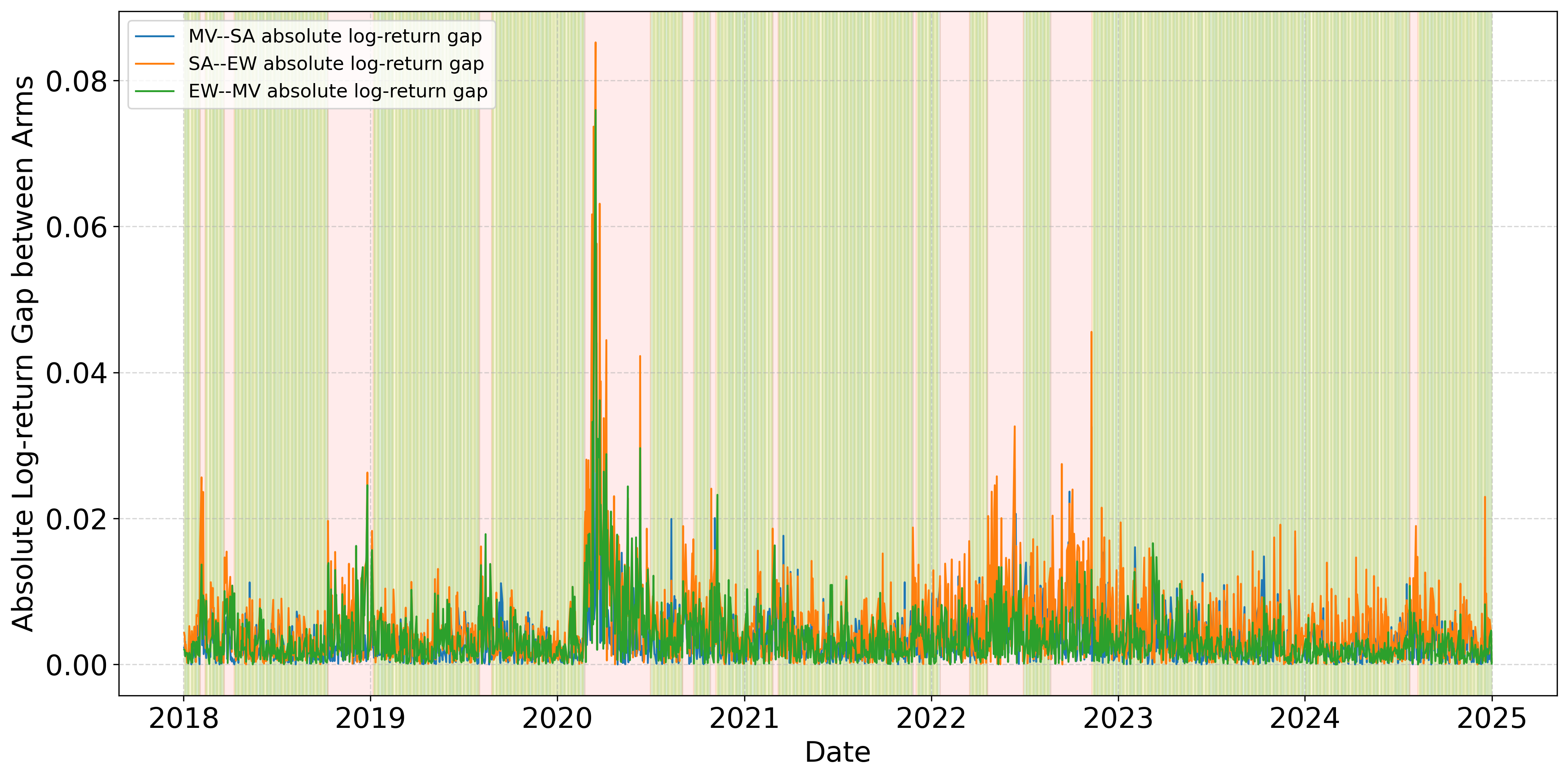}
    \caption{
    Panel (a) reports the cumulative wealth of CP-based and UCB-based bandit policies under a partial-information setting, compared with EW, MV and SA portfolio benchmarks. Shaded bands around randomised CP policies indicate \(95\%\) confidence intervals computed over \(500\) MC runs. Panel (b) shows the absolute log-return gaps between portfolio arms over time. The sample means of the MV--SA, SA--EW, and EW--MV absolute gaps are 
\(0.003793\), \(0.005476\), and \(0.003652\), respectively. In both panels, background shading highlights market regimes inferred via HMM: \textit{green} denotes \texttt{Bull} phases, \textit{gray} denotes \texttt{Neutral} markets, and \textit{pink} denotes \texttt{Bear} episodes.}
    \label{fig:cum_returns_and_absolute_gaps}
\end{figure}

Panel (a) of Figure~\ref{fig:cum_returns_and_absolute_gaps} illustrates the cumulative return or wealth trajectories associated with each strategy over the sample period. In the following, we refer to the policy defined in \eqref{eq:cp_regimeaware} as \emph{Regime-Aware CP}. Across both classical UCB and conformal variants, the regime-aware extensions consistently outperform their non–regime-aware counterparts. During extended \texttt{Bear} or high-volatility episodes (highlighted in pink), both the {\it Regime-Aware CP} and {\it Regime-Aware MV-UCB1} policies frequently switch into the more conservative SA portfolio. This adaptive behaviour substantially mitigates losses precisely in periods where \ref{eq:ucb1}, \ref{eq:cp_ucb}, and \ref{eq:mvucb1} remain overexposed to risky assets, leading to markedly weaker performance for these policies. The superior behaviour of regime-aware methods stems from their intrinsic sensitivity to non-stationarity: they are explicitly designed to adjust to structural breaks and volatility shifts, phenomena widely documented in empirical finance. %As a result, they are naturally better suited for environments in which reward distributions evolve over time. 
Among all stationary (non-regime-aware) bandit strategies, the \ref{eq:cp_ucb} policy achieves the highest cumulative wealth, highlighting the benefit of CP uncertainty quantification even when no explicit regime information is incorporated.

Table~\ref{tab:partial_info_portfolio_problem} provids a detailed comparison in terms of classical financial performance metrics: total return, Sharpe ratio, Max Drawdown, and Calmar ratio. These metrics jointly summarise both profitability and risk: total return captures overall growth, the Sharpe ratio adjusts returns for volatility, maximum drawdown reflects vulnerability to large losses, and the Calmar ratio evaluates return relative to worst-case drawdowns. A more detailed description can be found in \cite{bacon2022_risk_adjusted}. These results reinforce the graphical evidence provided in Panel (a) of Figure~\ref{fig:cum_returns_and_absolute_gaps}. Specifically, the {\it Regime-Aware CP} policy delivers by far the highest total return and Sharpe ratio, while maintaining the lowest drawdown and the highest Calmar ratio. Similarly, {\it Regime-Aware MV-UCB1} substantially outperforms its stationary counterpart. By contrast, stationary strategies exhibit markedly lower risk-adjusted performance, with \ref{eq:cp_ucb} being the strongest among them, still significantly behind the regime-aware methods.
\begin{comment}
    
\begin{figure}[t]
    \centering
    \includegraphics[width=\textwidth]{Figures/Cumulative_Wealth_with_Market_Regimes.png}
    \caption{
    Cumulative wealth of CP-based and UCB-based bandit policies under a partial-information setting, compared with EW, MV and SA portfolio benchmarks. Background shading highlights market regimes inferred via HMM: \textit{green} denotes \texttt{Bull} phases, \textit{gray} \texttt{Neutral} markets, and \textit{pink} \texttt{Bear} episodes. Shaded bands around randomised CP policies indicate $95\%$ confidence intervals computed over $1,000$ MC runs.
    }
    \label{fig:cum_returns}
\end{figure}

\end{comment}

\begin{table}[ht]
\centering
\caption{Comparison among the different policies under partial information in terms of standard financial performance metrics. Values are reported as means, with standard deviations in parentheses for randomised policies.}
\label{tab:partial_info_portfolio_problem}
\begin{tabular}{lcccc}
\toprule
\textbf{Strategy}
& \textbf{Total Return}
& \textbf{Annualised Sharpe}
& \textbf{Max Drawdown}
& \textbf{Calmar} \\
\midrule

CP-UCB
& $0.6185 \pm 0.1338$
& $0.5960 \pm 0.0877$
& $0.2587 \pm 0.0206$
& $0.2749 \pm 0.0588$ \\

Regime-Aware CP
& $1.5397 \pm 0.3171$
& $1.7988 \pm 0.2541$
& $0.0858 \pm 0.0064$
& $1.6576 \pm 0.3082$ \\

UCB1
& 0.2229
& 0.3923
& 0.1291
& 0.2263 \\

MV-UCB1
& 0.3366
& 0.6124
& 0.1173
& 0.3613 \\

Regime-Aware MV-UCB1
& 1.0442
& 1.5907
& 0.0818
& 1.3169 \\

Hold MV
& 0.1548
& 0.2776
& 0.2219
& 0.0938 \\

Hold EW
& 0.6774
& 0.6268
& 0.2656
& 0.2893 \\

\bottomrule
\end{tabular}
\end{table}

\section{Discussion and Future Work} \label{sec: conclusion}

This work introduced the \textit{Conformal Bandits} framework, a novel class of policies that bridge the regret-minimising decision-making nature of bandits with the statistical validity of CP. Building on the classical UCB-type of policies, we extended the classical Hoeffding-based confidence bounds for the mean rewards of the arm $\mu_k, k=1,\dots,K$, with distribution- and model-free CP intervals for a future observable reward $Y_{t+1}$. This represents a shift toward the class of {\it predictive bandits}, which have shown regret promise in other works~\citep{liu2023nonstationary,duranmartingale, giovagnoli2025note}, under larger-gaps and mainly under a Bayesian setting. 

In this work, we were motivated by a portfolio allocation problem, a typical small-gap and low-SNR regime~\citep{demiguel2009optimal}. Through extensive simulation studies and real-world financial data, we demonstrated superior performance of our {\it Conformal Bandits} in this challenging setting, where the classical UCB1 exhibits linear regret. By being based on adaptive data-driven intervals constructed through conformalised quantile regression, our policy makes it possible to better distinguish upside potential from downside risk by exploiting distributional asymmetry. A further extension of the novel class of policies to regime-switching market dynamics (through HMM), demonstrated substantial increase in the cumulative portfolio wealth  when applied to the problem of interest.

The key advantages of our framework lie in its ability to enhance the {\it exploration-exploitation} efficiency of bandit algorithms by combining their regret efficiency with the statistical rigour of conformal prediction, providing valid uncertainty quantification without major distributional assumptions. Validity of CP extends beyond well-specified models, making it a robust solution to model mispecification or subjective priors under a Bayesian perspective~\citep[see e.g.,][]{deliu2026interplay}. This is particularly valuable in real-world applications characterised by small data, complex reward distributions, and the need for principled risk management.

\subsection{Limitations}
Some limitations warrant discussion. First, while our regime-aware extension addresses non-stationarity through explicit regime identification via HMMs, a fully non-stationary bandit policy, capable of adapting to gradual distributional drifts or structural breaks without explicit regime specification, may better suit the setting of interest. Second, CP relies fundamentally on exchangeability, which is frequently violated in time series. To address this limitation, we employed {\it Adaptive Conformal Inference}~\citep{gibbs2021adaptive}, guaranteed to ensure asymptotic coverage, and showing finite-sample coverage in our empirical studies. Alternative forms of weak exchangeability may exist and may be exploited in future work, including block-exchangeable structures, where data exhibit dependence within groups but independence across groups. Developing conformal bandit variants tailored to such structures represents a promising research direction. Third, our framework treats arm rewards as independent, neglecting potential correlations between arms. Incorporating correlation structures could enhance both statistical properties and decision making. 

\subsection{Future Directions}
Several promising avenues emerge from this work. First, extending the CP framework to other bandit policies, particularly \textit{Thompson Sampling}~\citep{thompson1933likelihood}, represents a natural next step. Since Thompson Sampling operates under a Bayesian paradigm, integrating CP could provide calibrated prediction intervals with guaranteed frequentist properties~\citep{deliu2026interplay}. 
Second, a formal theoretical regret analysis for {\it Conformal Bandits} remains an important avenue for future work. Although comprehensive empirical evidence shows superior regret performance in small-gaps (with logarithmic behaviour), formal regret bounds might validate this finding, but it poses significant challenges. The non-parametric, distribution-free nature of CP, precisely what makes it attractive, complicates traditional regret analysis, which typically relies on additional distributional assumptions (e.g., sub-Gaussian rewards).
Beyond these methodological extensions, our framework has broad applicability across domains exhibiting small-gap regimes. As discussed in Section~\ref{sec:intro}, both medical treatments~\citep{villar2015multiarmed} and digital health interventions yield negligible average effects~\citep{szaszi2022no}, all settings where distinguishing between competing options with minimal expected reward differences is critical, yet classical bandit algorithms struggle. 

In conclusion, this work opens the door to diverse research directions spanning applied and theoretical machine learning, statistics, and finance. By demonstrating that finite-sample statistical validity and regret efficiency can be jointly achieved, we hope to stimulate further integration of these traditionally separate research streams, ultimately enabling more robust and reliable sequential decision-making in complex, uncertain environments.

\paragraph{Acknowledgements} This work was supported by Sapienza Research Grants, Ateneo Piccoli 2024–RP1241905F06F855.

\bibliographystyle{abbrv}
\nocite{*}
\bibliography{main}

\appendix

\section*{Appendix}
%\addcontentsline{toc}{section}{Appendix} 

\section{Additional Simulation Studies}\label{app:add_sim_stud}

To extend the analysis reported in the main text for the small-gap configuration, we consider additional mean-gap regimes while keeping the reward-generating mechanisms unchanged. In the baseline experiment, the bandit instance is characterized by a unique optimal arm \(k^*\) and by a constant gap $\Delta_k = \mu^* - \mu_k = 0.01$, $k \neq k^*$, between the mean reward of the optimal arm and those of all suboptimal arms. 

\begin{figure}[t]
\centering
\setlength{\tabcolsep}{6pt}

\begin{tabular}{cc}

% =====================================
% =========== Gaussian row ===========
% =====================================

% Gaussian Regret
\begin{subfigure}[t]{0.45\textwidth}
\centering
\includegraphics[width=\linewidth]{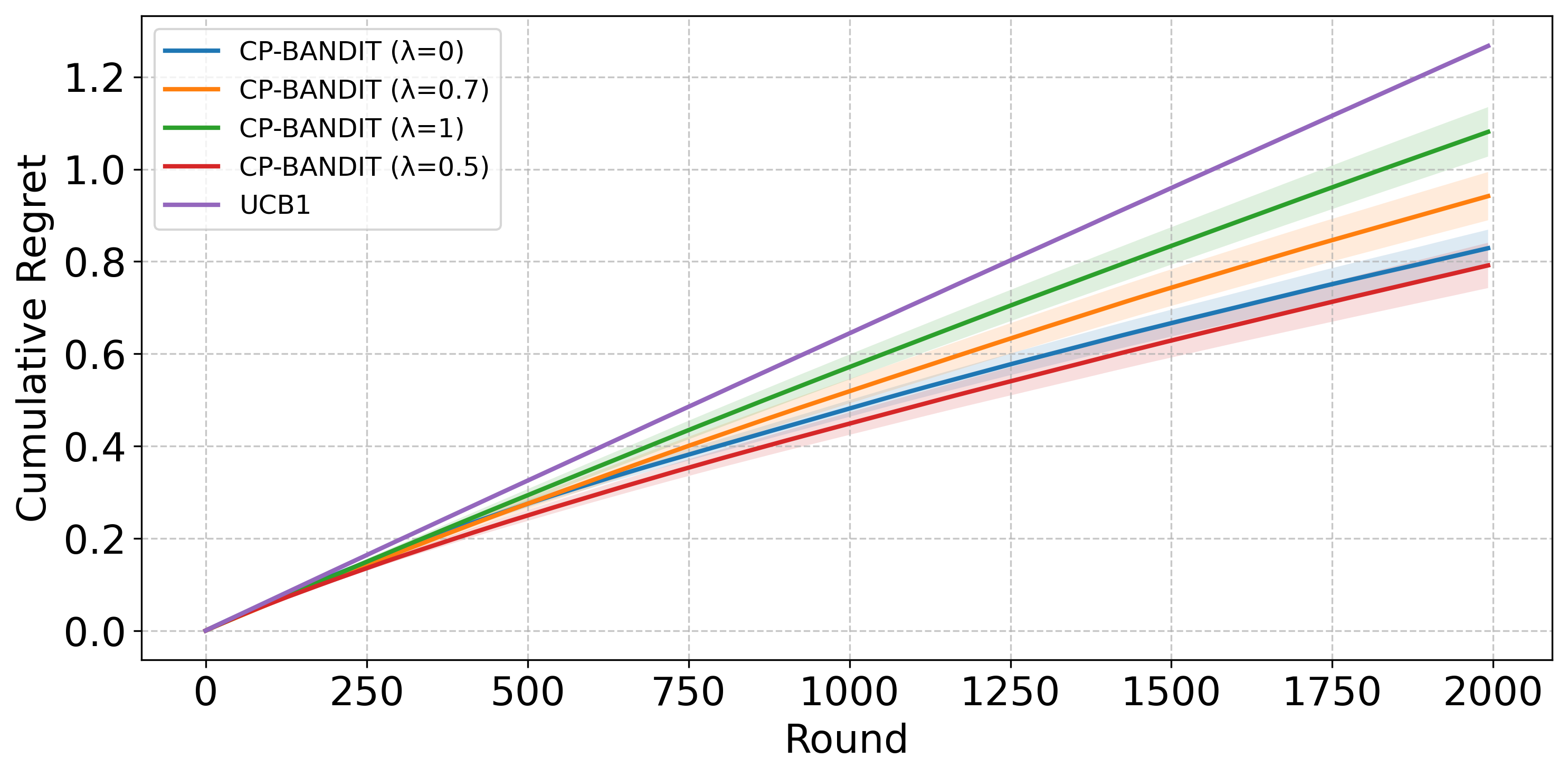}
\caption*{Gaussian: Regret}
\end{subfigure}
&
% Gaussian Best Arm
\begin{subfigure}[t]{0.45\textwidth}
\centering
\includegraphics[width=\linewidth]{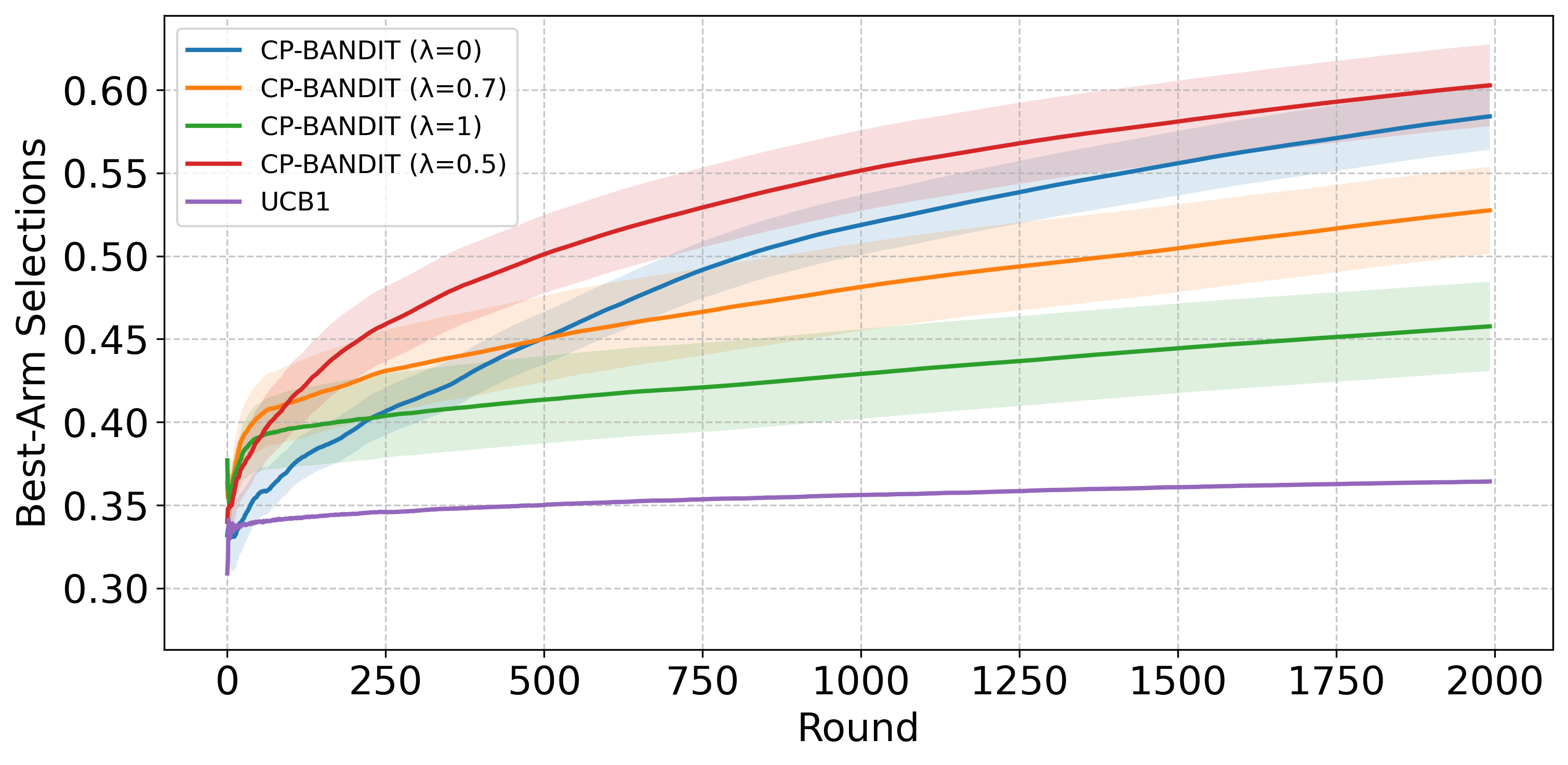}
\caption*{Gaussian: Best-Arm}
\end{subfigure}
\\[1.1em]

% =====================================
% ========= Student-t row =============
% =====================================

% Student-t Regret
\begin{subfigure}[t]{0.45\textwidth}
\centering
\includegraphics[width=\linewidth]{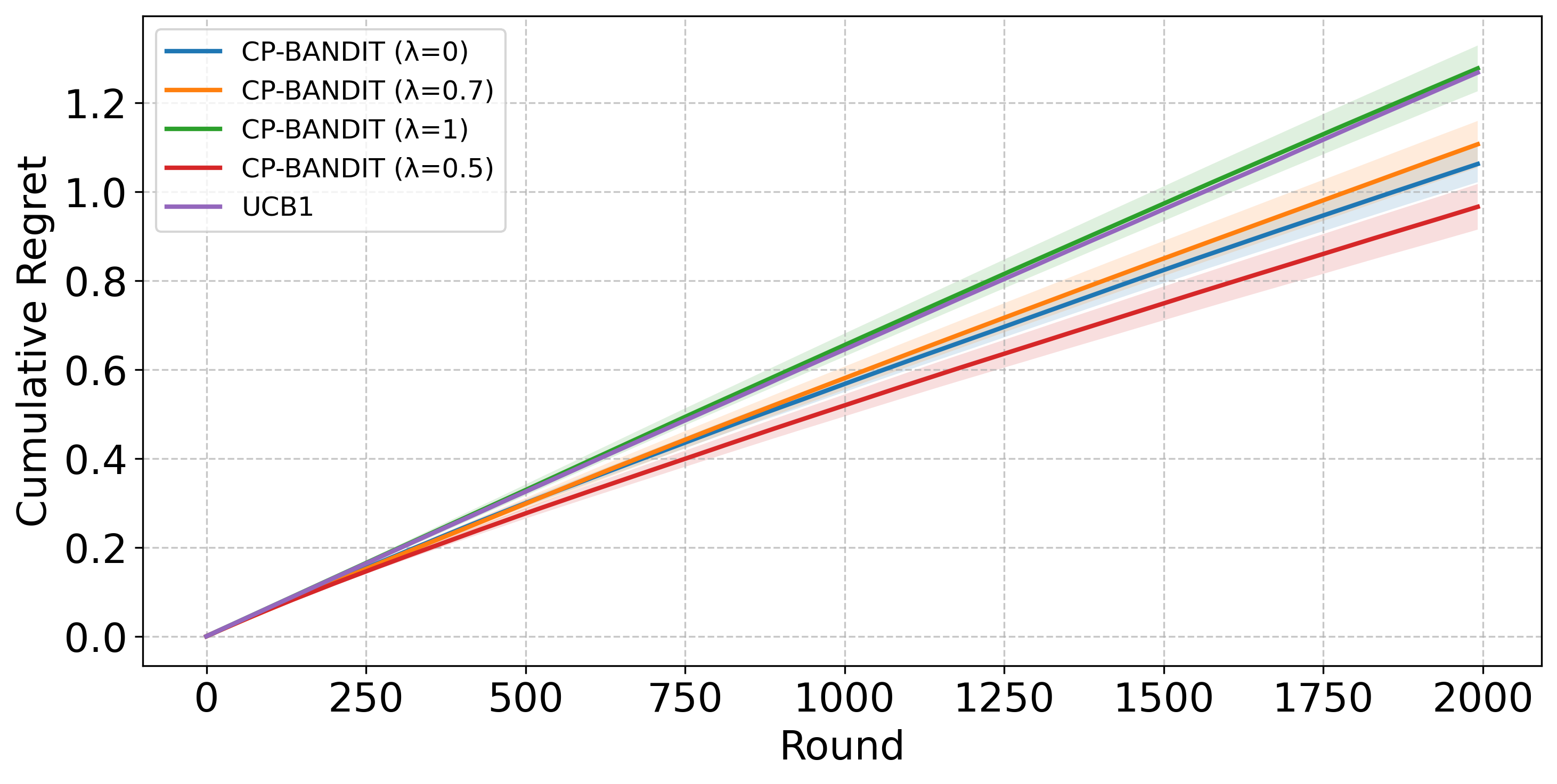}
\caption*{Student-$t$: Regret}
\end{subfigure}
&
% Student-t Best Arm
\begin{subfigure}[t]{0.45\textwidth}
\centering
\includegraphics[width=\linewidth]{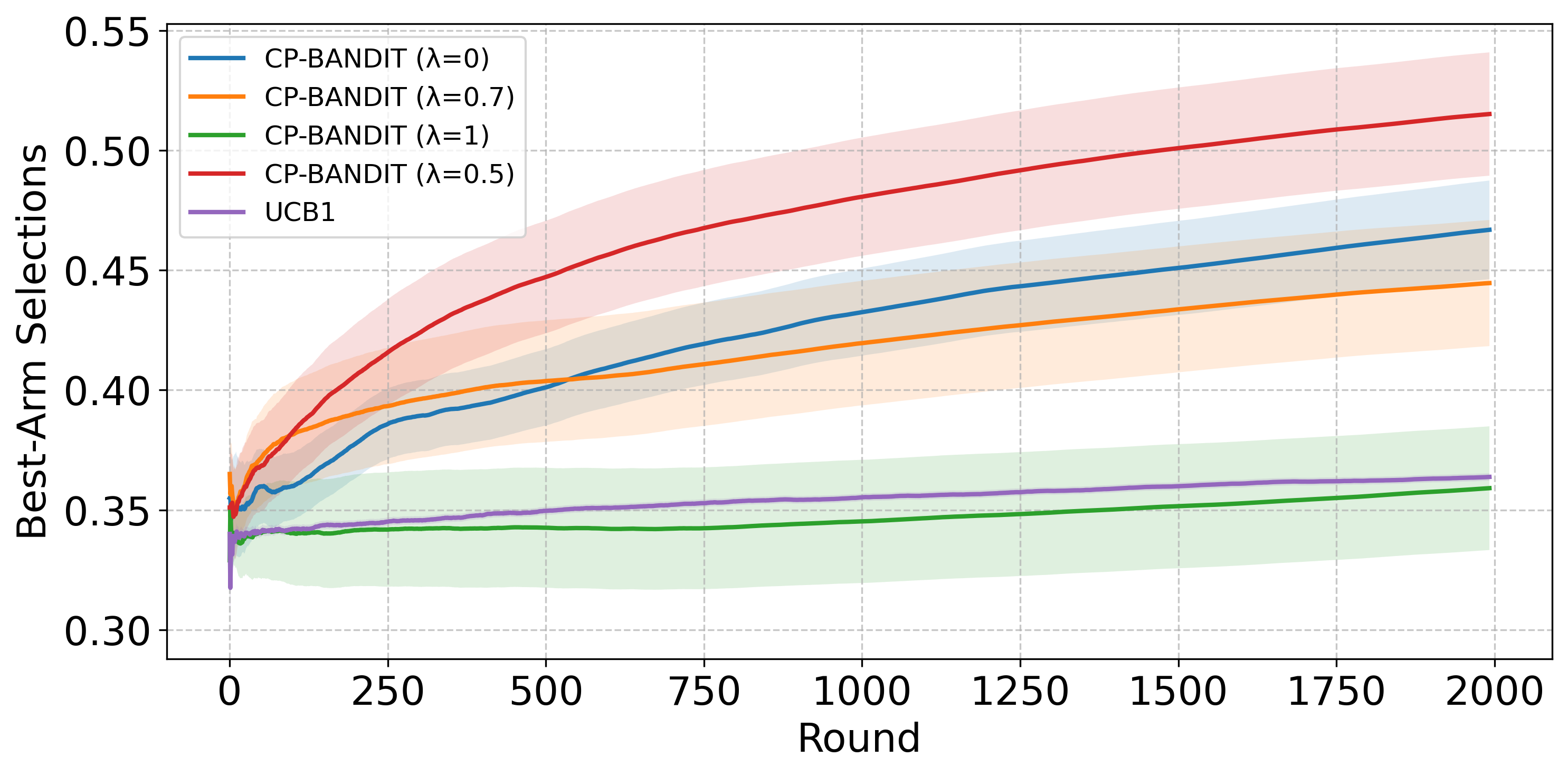}
\caption*{Student-$t$: Best-Arm}
\end{subfigure}
\\[1.1em]

% =====================================
% ========== Skew-t row ===============
% =====================================

% Skew-t Regret
\begin{subfigure}[t]{0.45\textwidth}
\centering
\includegraphics[width=\linewidth]{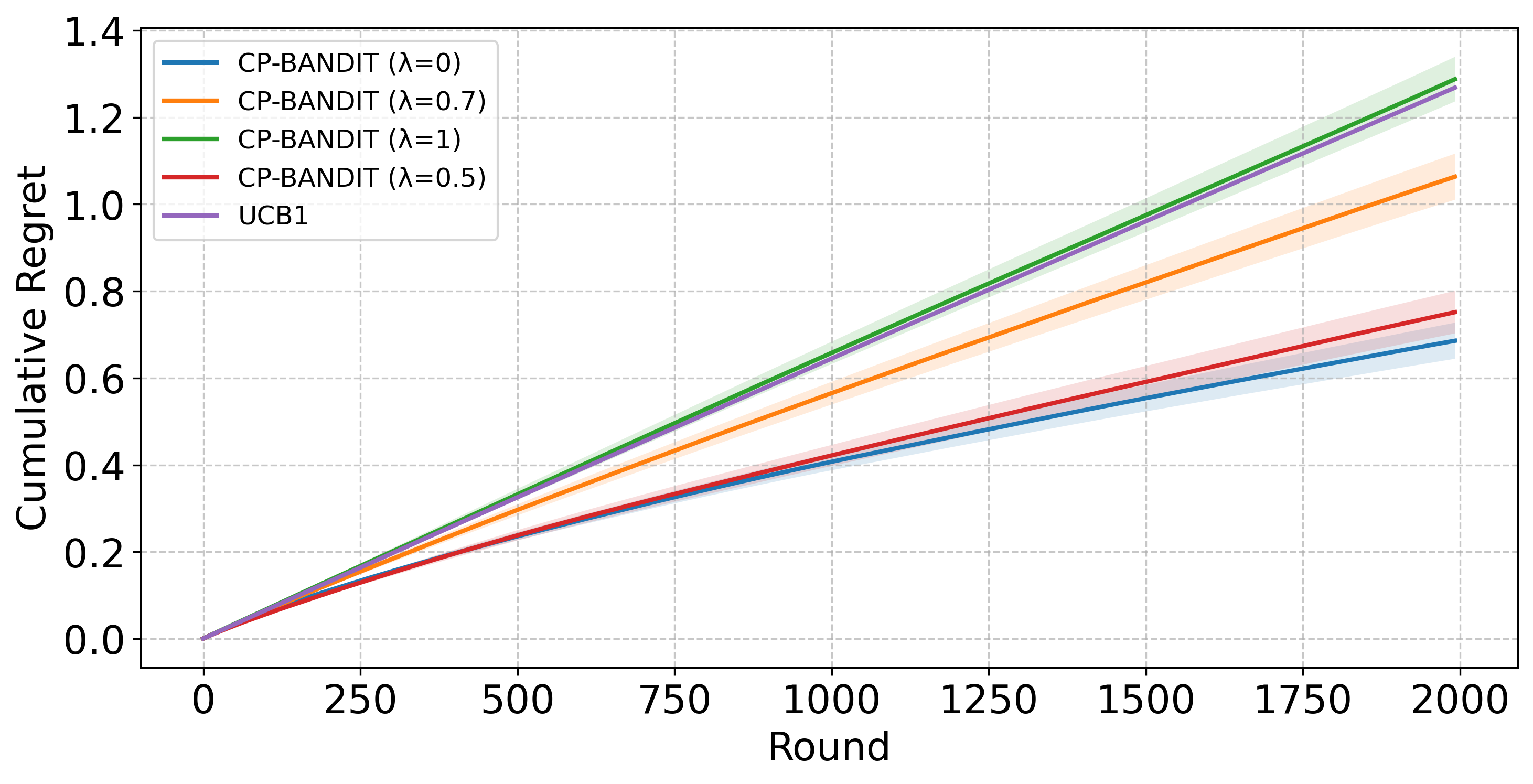}
\caption*{Skew-$t$: Regret}
\end{subfigure}
&
% Skew-t Best Arm
\begin{subfigure}[t]{0.45\textwidth}
\centering
\includegraphics[width=\linewidth]{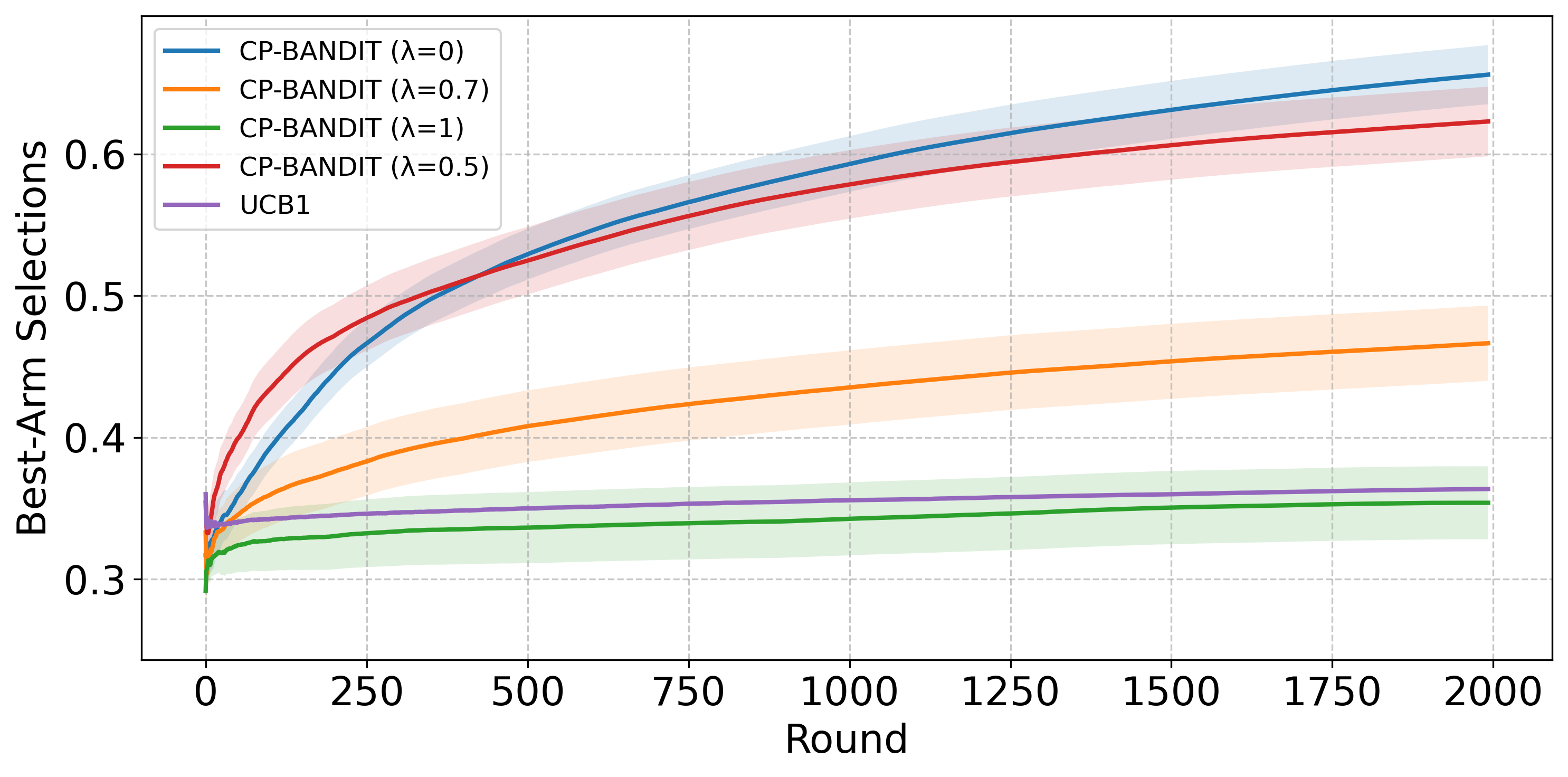}
\caption*{Skew-$t$: Best-Arm}
\end{subfigure}

\end{tabular}

\caption{
Comparison between the Conformal Bandit variants and classical UCB1 in terms of cumulative regret and cumulative best-arm selection rates for the gap setting \(\Delta_k=0.001\), corresponding to \(\mu_1=\mu^*=0.001\) and \(\mu_2=\mu_3=0\), based on \(1000\) Monte Carlo simulations.
Rows correspond to different reward-generating environments: Gaussian, Student-$t$, and skew-$t$ with asymmetric tails.
The left column reports cumulative regret, while the right column reports cumulative best-arm selection rates.
Shaded regions represent $95\%$ Monte Carlo confidence intervals.}
\label{fig:six_panel_0.001}
\end{figure}

\begin{figure}[t]
\centering
\setlength{\tabcolsep}{6pt}

\begin{tabular}{cc}

% =====================================
% =========== Gaussian row ===========
% =====================================

% Gaussian Regret
\begin{subfigure}[t]{0.45\textwidth}
\centering
\includegraphics[width=\linewidth]{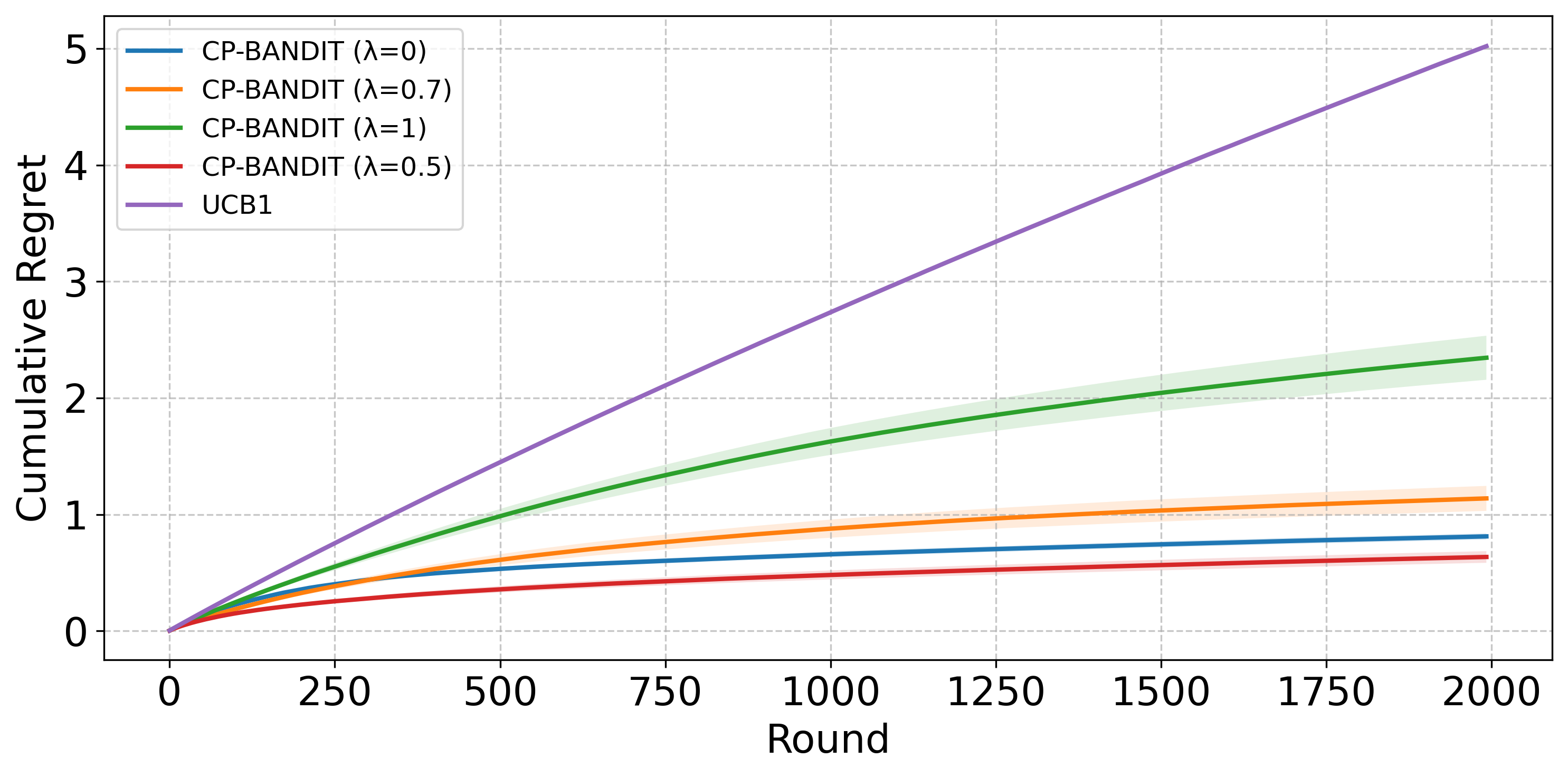}
\caption*{Gaussian: Regret}
\end{subfigure}
&
% Gaussian Best Arm
\begin{subfigure}[t]{0.45\textwidth}
\centering
\includegraphics[width=\linewidth]{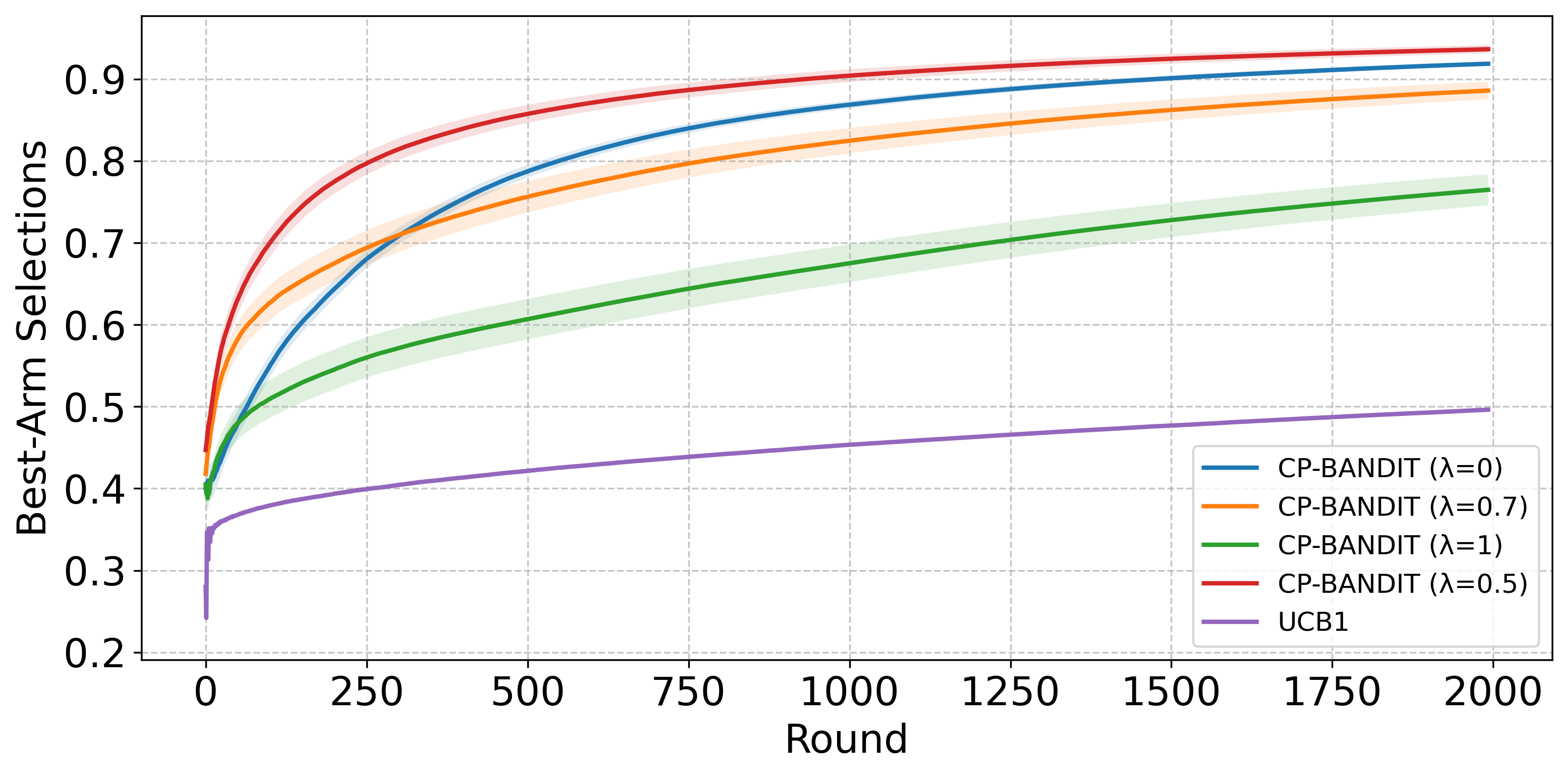}
\caption*{Gaussian: Best-Arm}
\end{subfigure}
\\[1.1em]

% =====================================
% ========= Student-t row =============
% =====================================

% Student-t Regret
\begin{subfigure}[t]{0.45\textwidth}
\centering
\includegraphics[width=\linewidth]{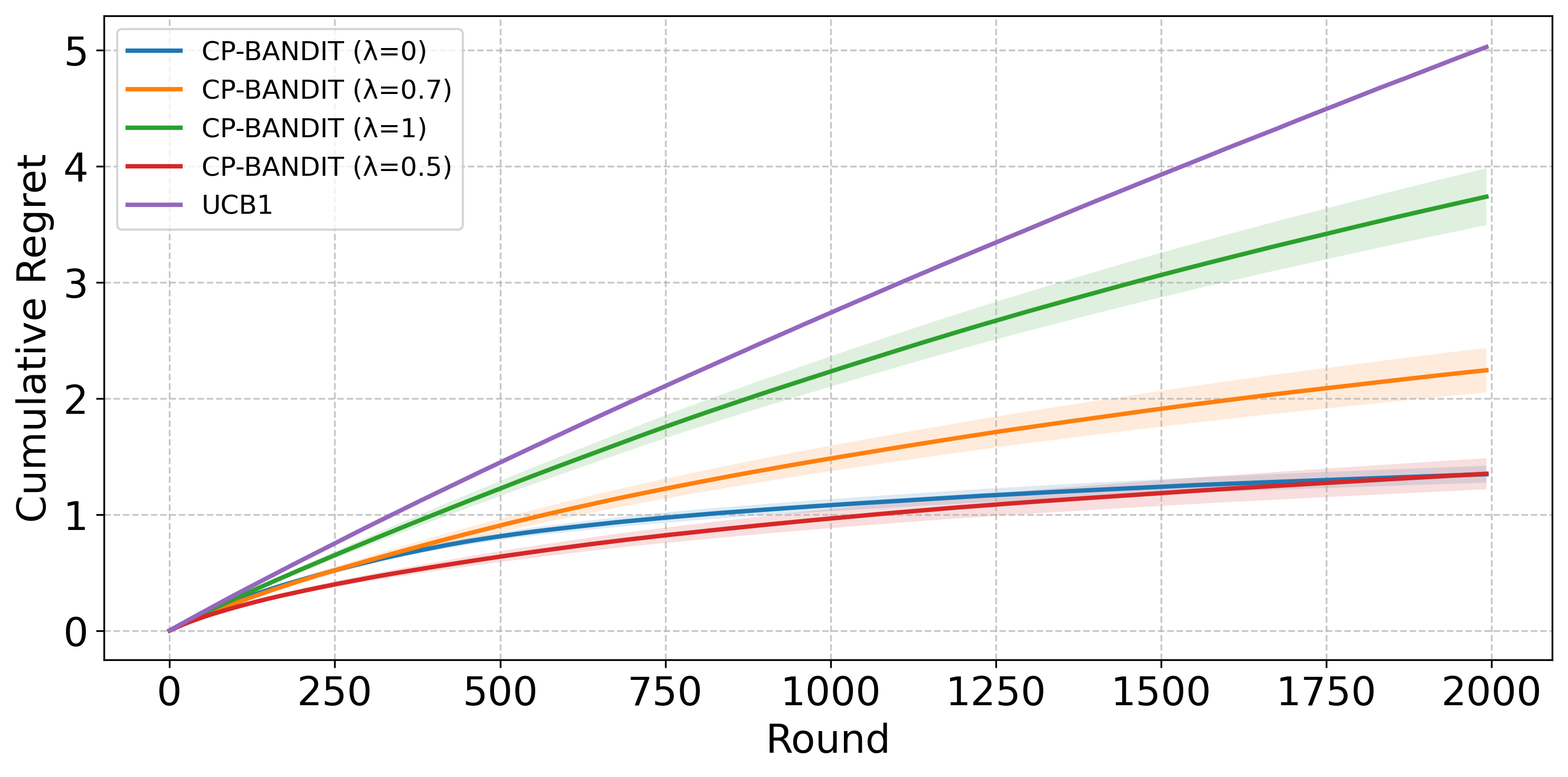}
\caption*{Student-$t$: Regret}
\end{subfigure}
&
% Student-t Best Arm
\begin{subfigure}[t]{0.45\textwidth}
\centering
\includegraphics[width=\linewidth]{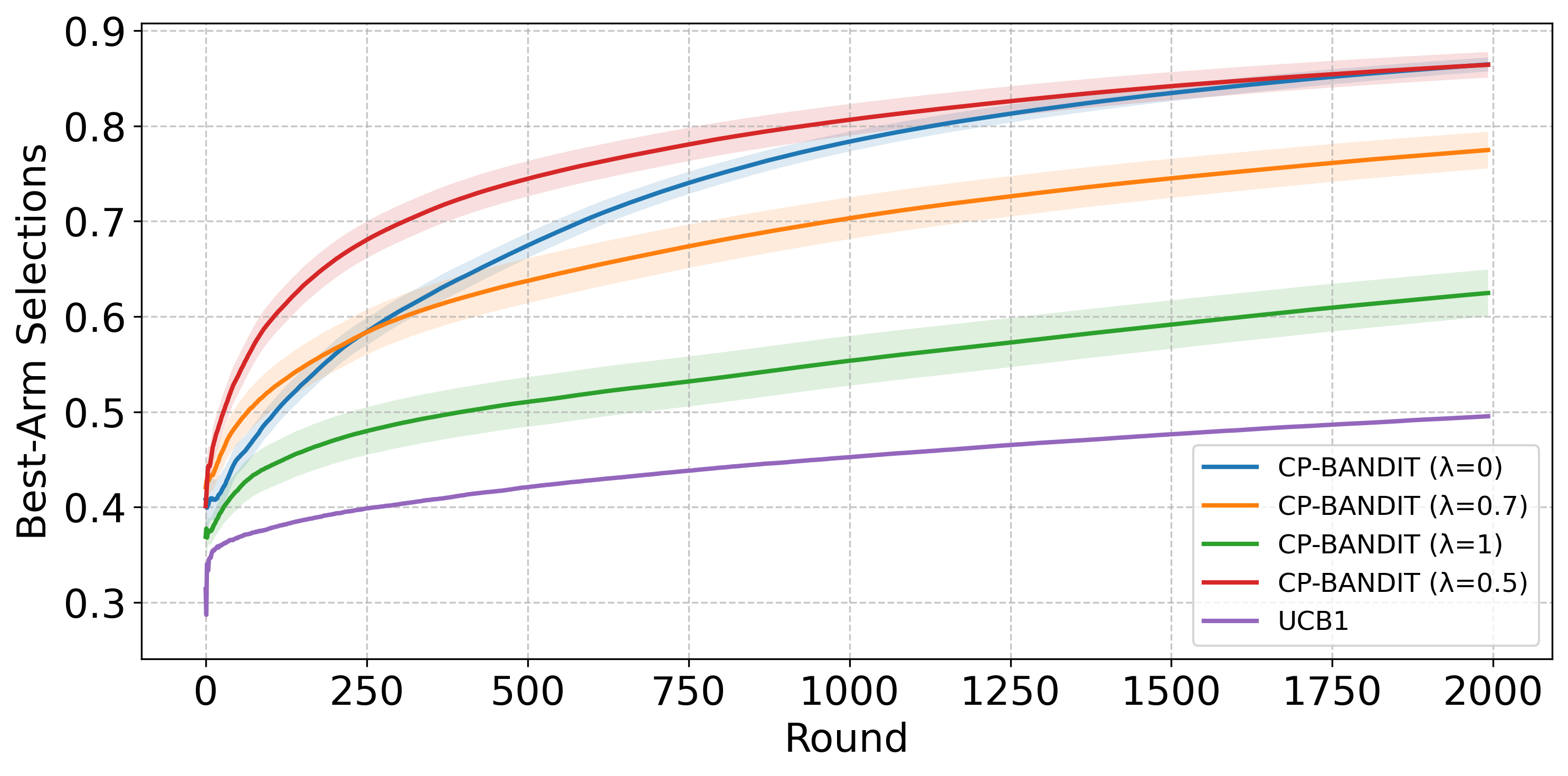}
\caption*{Student-$t$: Best-Arm}
\end{subfigure}
\\[1.1em]

% =====================================
% ========== Skew-t row ===============
% =====================================

% Skew-t Regret
\begin{subfigure}[t]{0.45\textwidth}
\centering
\includegraphics[width=\linewidth]{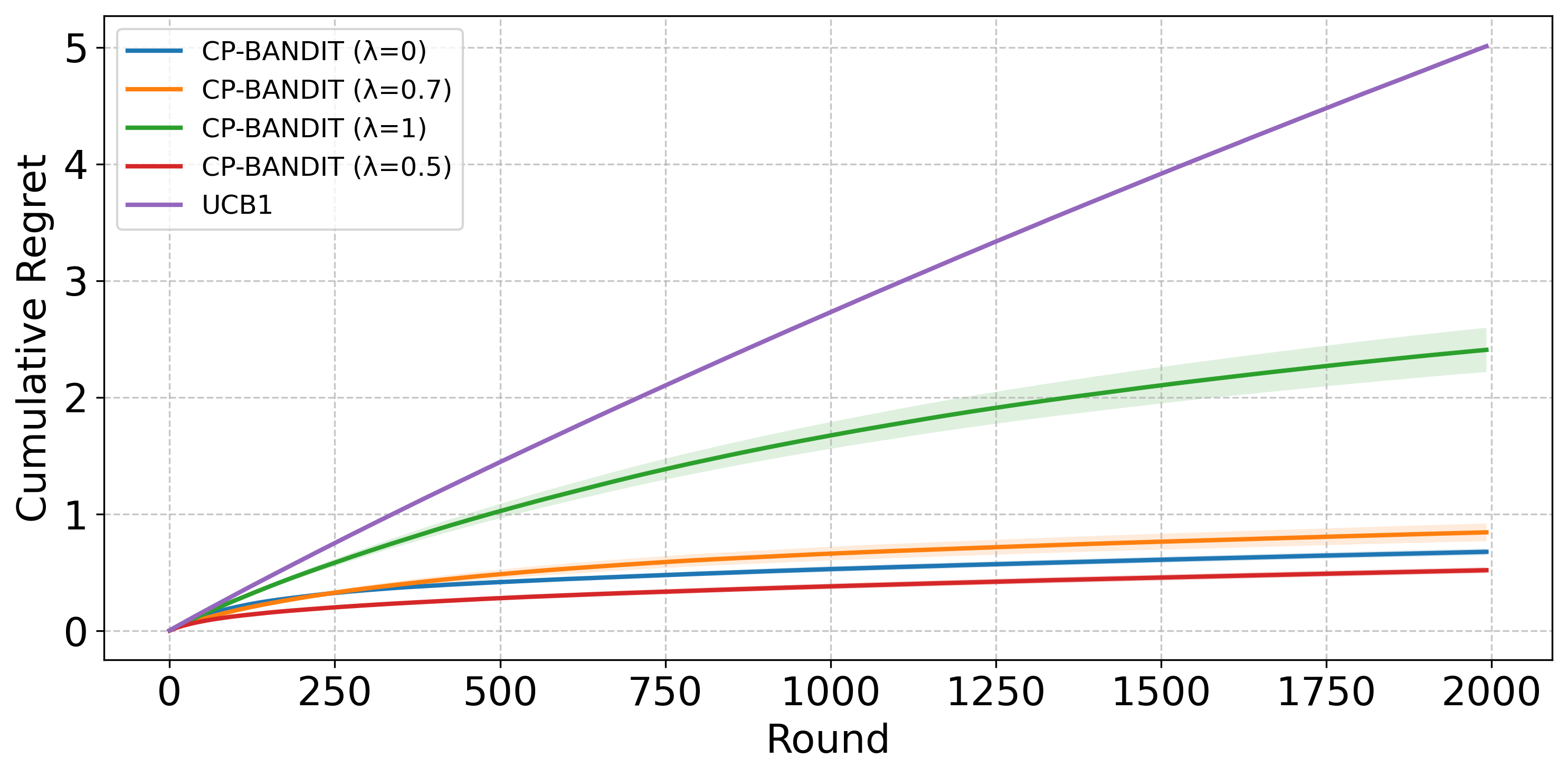}
\caption*{Skew-$t$: Regret}
\end{subfigure}
&
% Skew-t Best Arm
\begin{subfigure}[t]{0.45\textwidth}
\centering
\includegraphics[width=\linewidth]{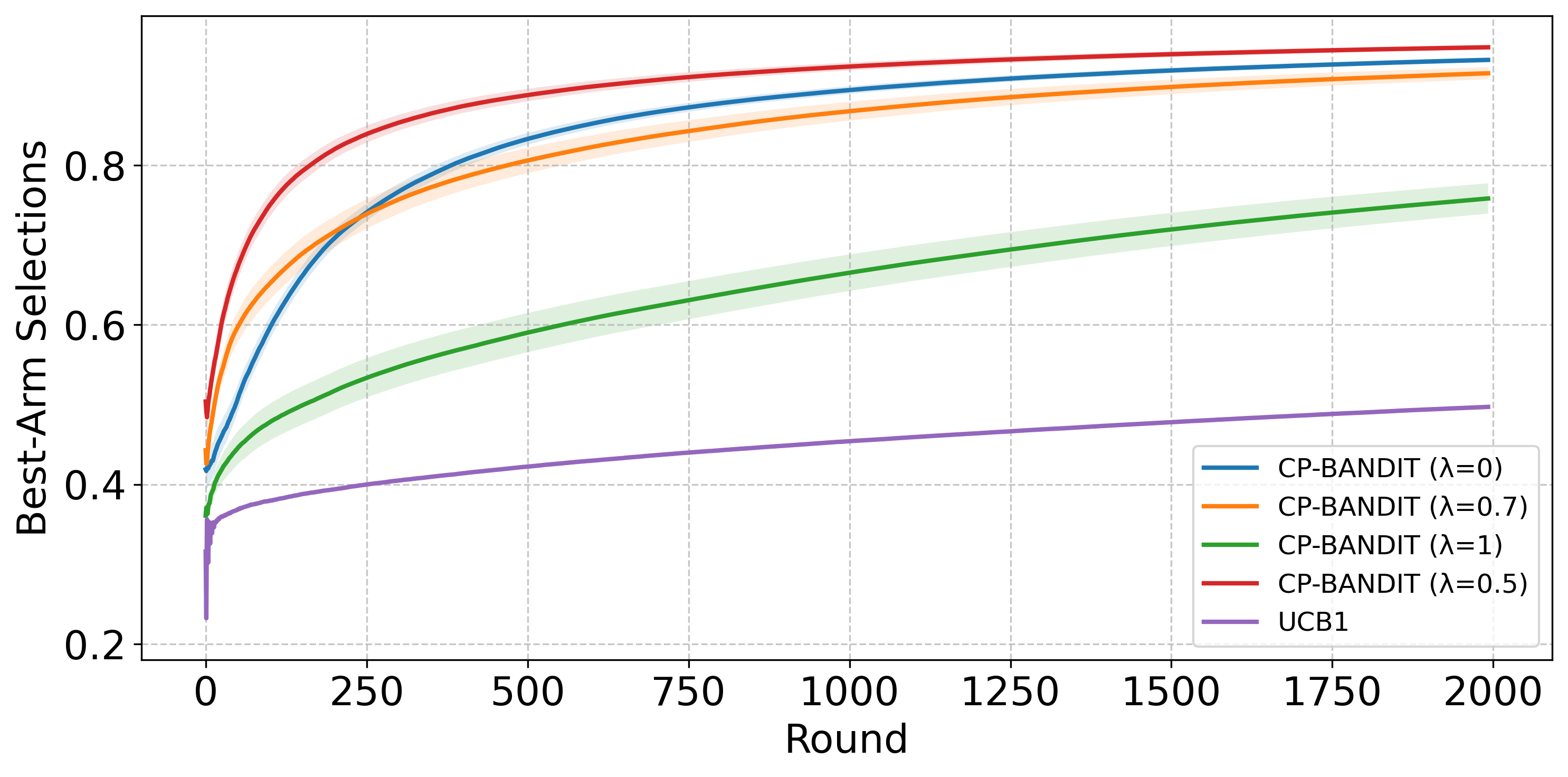}
\caption*{Skew-$t$: Best-Arm}
\end{subfigure}

\end{tabular}

\caption{
Comparison between the Conformal Bandit variants and classical UCB1 in terms of cumulative regret and cumulative best-arm selection rates for the gap setting \(\Delta_k=0.005\), corresponding to \(\mu_1=\mu^*=0.005\) and \(\mu_2=\mu_3=0\), based on \(1000\) Monte Carlo simulations.
Rows correspond to different reward-generating environments: Gaussian, Student-$t$, and skew-$t$ with asymmetric tails.
The left column reports cumulative regret, while the right column reports cumulative best-arm selection rates.
Shaded regions represent $95\%$ Monte Carlo confidence intervals.
}
\label{fig:six_panel_0.005}
\end{figure}

\begin{figure}[t]
\centering
\setlength{\tabcolsep}{6pt}

\begin{tabular}{cc}

% =====================================
% =========== Gaussian row ===========
% =====================================

% Gaussian Regret
\begin{subfigure}[t]{0.45\textwidth}
\centering
\includegraphics[width=\linewidth]{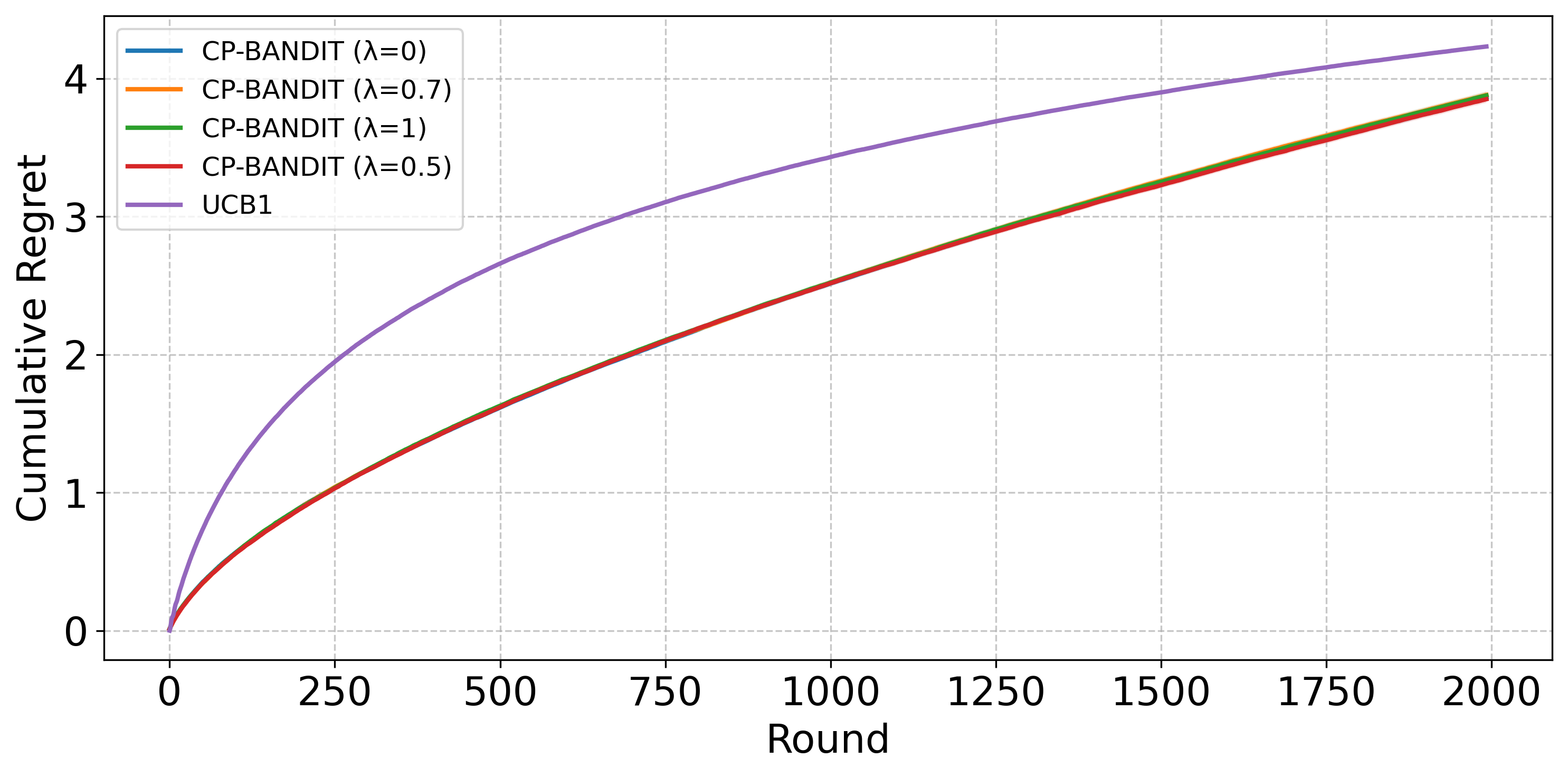}
\caption*{Gaussian: Regret}
\end{subfigure}
&
% Gaussian Best Arm
\begin{subfigure}[t]{0.45\textwidth}
\centering
\includegraphics[width=\linewidth]{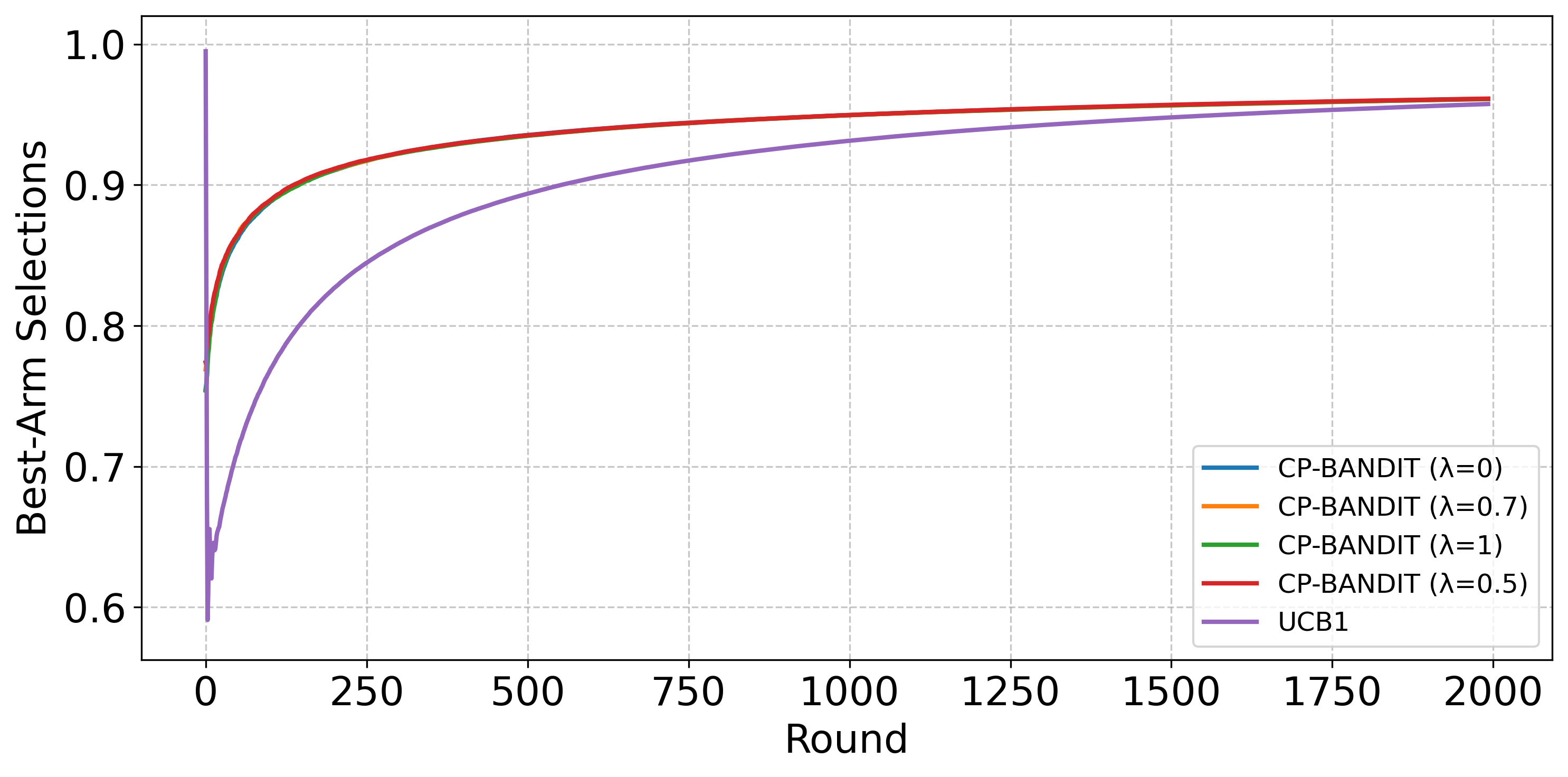}
\caption*{Gaussian: Best-Arm}
\end{subfigure}
\\[1.1em]

% =====================================
% ========= Student-t row =============
% =====================================

% Student-t Regret
\begin{subfigure}[t]{0.45\textwidth}
\centering
\includegraphics[width=\linewidth]{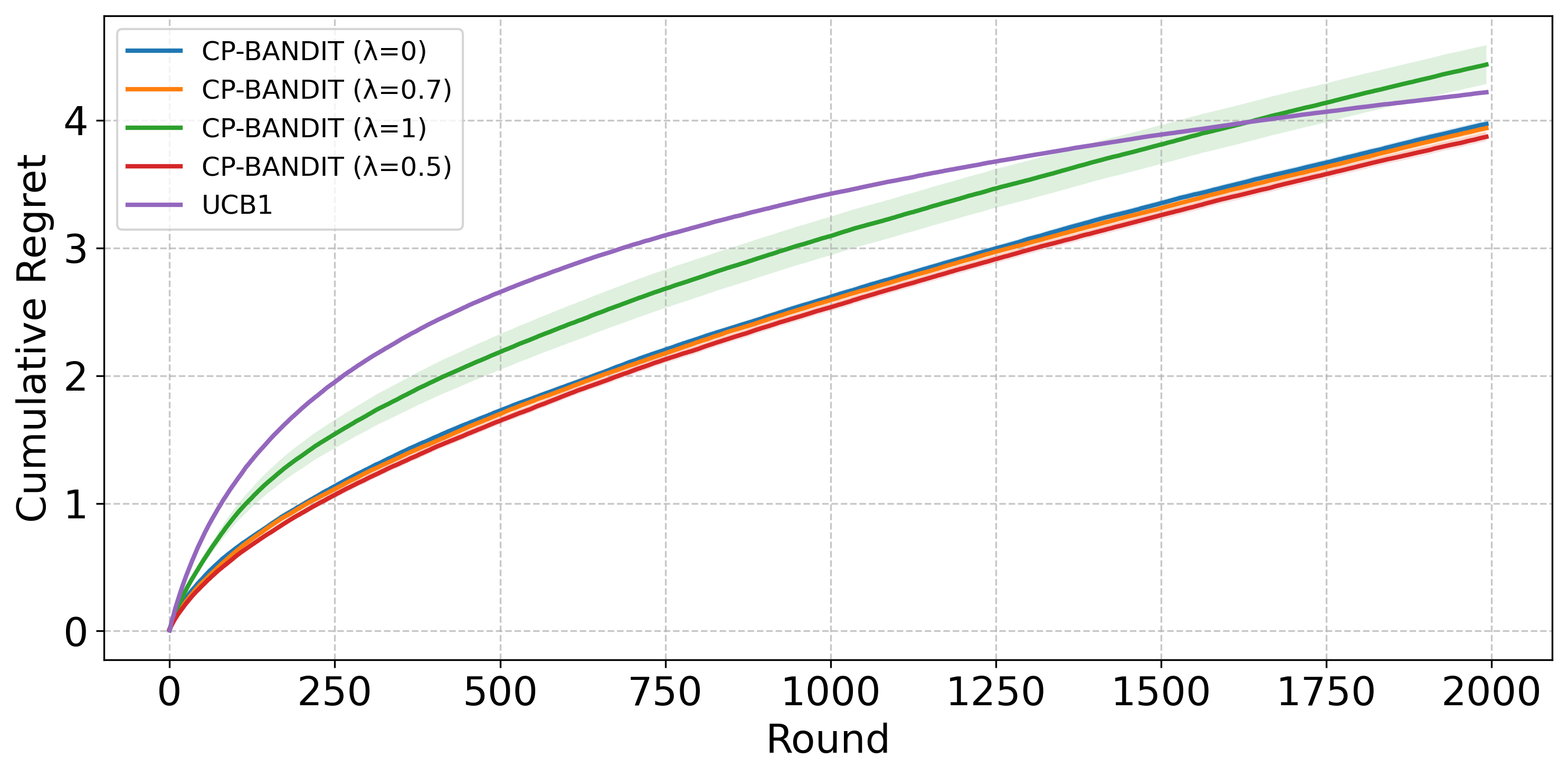}
\caption*{Student-$t$: Regret}
\end{subfigure}
&
% Student-t Best Arm
\begin{subfigure}[t]{0.45\textwidth}
\centering
\includegraphics[width=\linewidth]{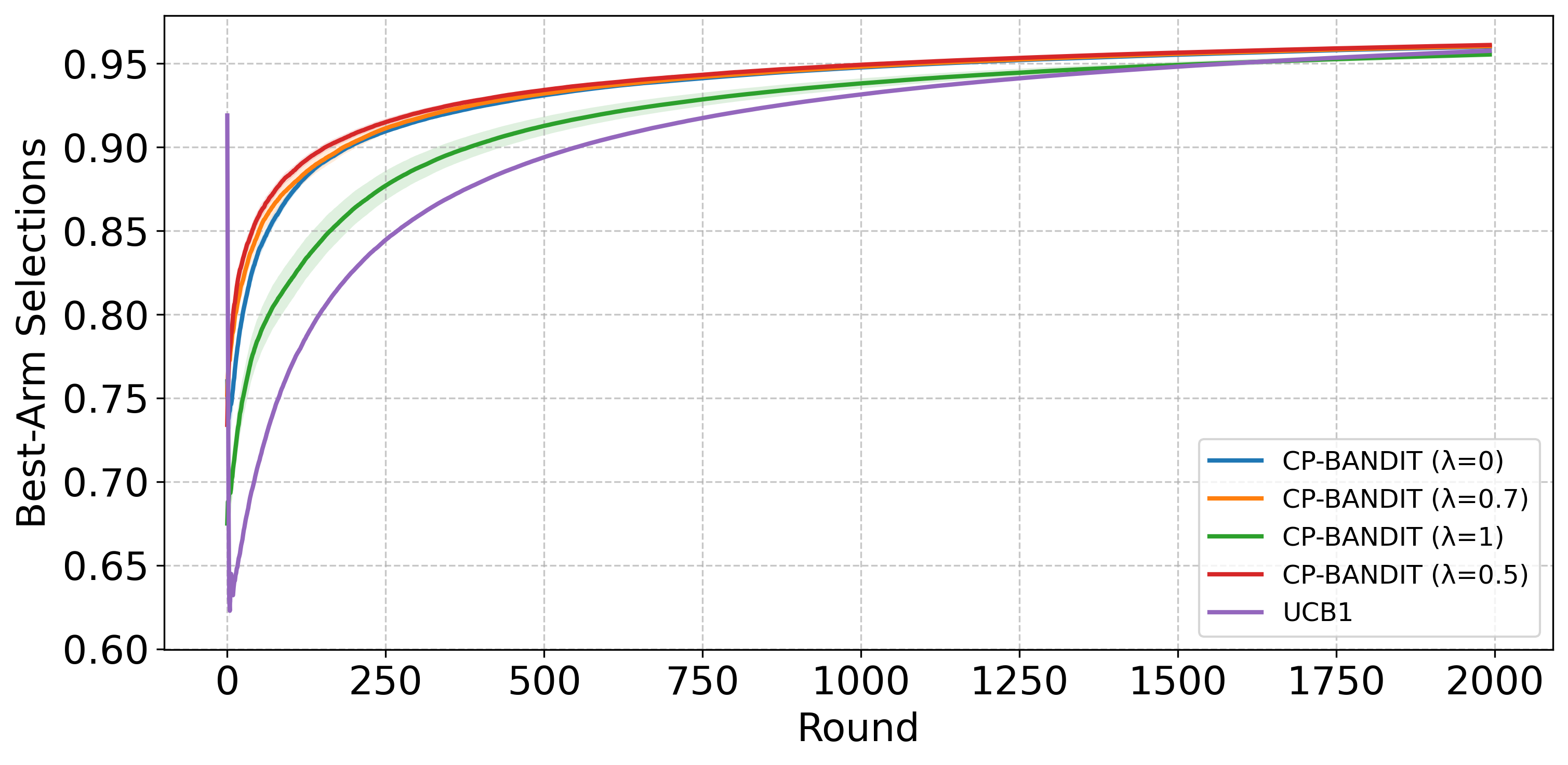}
\caption*{Student-$t$: Best-Arm}
\end{subfigure}
\\[1.1em]

% =====================================
% ========== Skew-t row ===============
% =====================================

% Skew-t Regret
\begin{subfigure}[t]{0.45\textwidth}
\centering
\includegraphics[width=\linewidth]{Figures/Regret_plot_skew-t_0.005.png}
\caption*{Skew-$t$: Regret}
\end{subfigure}
&
% Skew-t Best Arm
\begin{subfigure}[t]{0.45\textwidth}
\centering
\includegraphics[width=\linewidth]{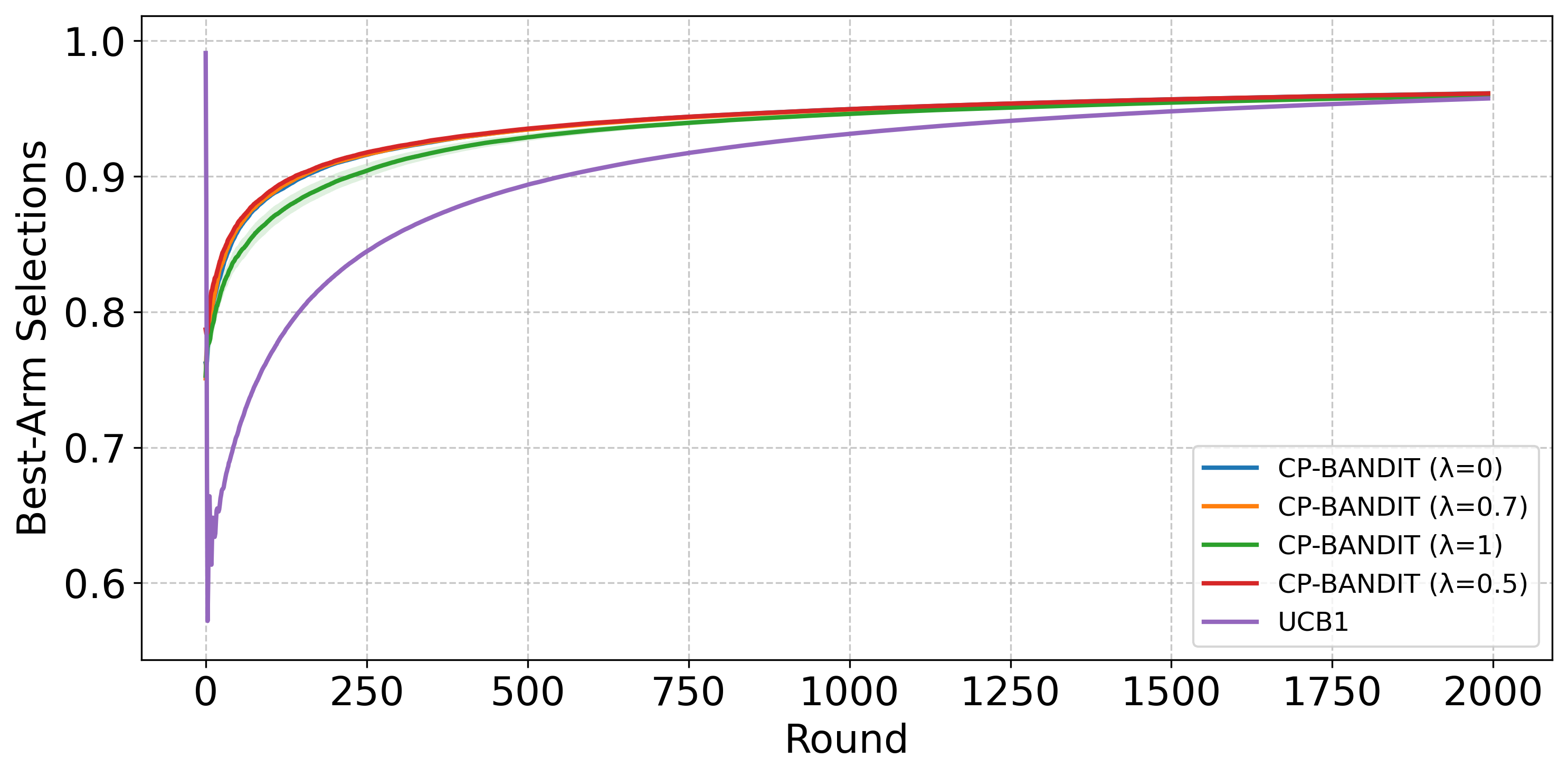}
\caption*{Skew-$t$: Best-Arm}
\end{subfigure}

\end{tabular}

\caption{
Comparison between the Conformal Bandit variants and classical UCB1 in terms of cumulative regret and cumulative best-arm selection rates for the gap setting \(\Delta_k=0.05\), corresponding to \(\mu_1=\mu^*=0.05\) and \(\mu_2=\mu_3=0\), based on \(1000\) Monte Carlo simulations.
Rows correspond to different reward-generating environments: Gaussian, Student-$t$, and skew-$t$ with asymmetric tails.
The left column reports cumulative regret, while the right column reports cumulative best-arm selection rates.
Shaded regions represent $95\%$ Monte Carlo confidence intervals.
}
\label{fig:six_panel_0.05}
\end{figure}

In this section, we preserve the same distributional settings considered in the main experiments, namely Gaussian, Student-\(t\), and skewed Student-\(t\) reward environments, and vary only the separation between the optimal and suboptimal mean rewards. Specifically, we consider the additional gap values $\Delta_k = \mu^* - \mu_k \in \{0.001, 0.005, 0.05\}$, $k \neq k^*$. The corresponding simulation results, covering both bandit performance and uncertainty quantification, are reported in Figures~\ref{fig:six_panel_0.001}, \ref{fig:six_panel_0.005}, and \ref{fig:six_panel_0.05}, and in Tables~\ref{tab:0.001}, \ref{tab:0.005}, and \ref{tab:0.05}, respectively. Relative to the baseline configuration \(\Delta_k=0.01\), these cases correspond respectively to a tenfold compression, a twofold compression, and a fivefold enlargement of the expected reward gap. This allows us to assess the robustness of the proposed methods as the statistical difficulty of identifying the optimal arm varies. All other components of the simulation pipeline are kept unchanged: the bandit horizon, the conformal prediction procedure, the number of Monte Carlo simulations and the evaluation metrics remain identical to those described in Section \ref{sec: Simulation Studies} of the main text.

The results show a clear dependence of the learning dynamics on the signal-to-noise ratio induced by the reward gap. When the gap is very small, \(\Delta_k=0.001\), the optimal and suboptimal arms are difficult to distinguish within the finite horizon considered. This is reflected in the best-arm selection curves, which remain far from one for most methods, and in cumulative regret trajectories that are close to linear over a substantial portion of the horizon. Nevertheless, the conformal bandit variants consistently outperform \ref{eq:ucb1} across the three reward-generating environments. 

As the gap increases to \(\Delta_k=0.005\), the separation between the policies becomes more pronounced. The conformal bandit variants improve their rate of optimal-arm identification and display lower cumulative regret than classical \ref{eq:ucb1} across the three reward-generating environments. For the largest gap considered, \(\Delta_k=0.05\), the optimal arm becomes substantially easier to identify, and all methods exhibit improved best-arm selection performance. Nevertheless, the conformal bandit variants retain a clear advantage across all reward-generating scenarios. This advantage is particularly pronounced in the skewed Student-\(t\) environment, where \ref{eq:ucb1} still exhibits an approximately linear cumulative regret profile over the considered horizon. This suggests slower adaptation to the optimal arm, despite the larger mean separation, when the reward distribution is asymmetric. However, the apparent improvement of \ref{eq:ucb1} in the Gaussian and Student-\(t\) scenarios should be interpreted together with its uncertainty quantification performance. As shown in Table~\ref{tab:0.05}, \ref{eq:ucb1} does not uniformly achieve the nominal coverage level \(1-\alpha=0.8\). In particular, for the optimal arm, its empirical coverage is slightly below the target in the Gaussian setting and substantially below the target in the Student-\(t\) setting, with values equal to \(79.39\%\) and \(71.42\%\), respectively. By contrast, the coverage of the suboptimal arms is close to \(100\%\), indicating that \ref{eq:ucb1} produces a strongly unbalanced uncertainty profile across arms in this large-gap regime.

Among the conformal policies, \ref{eq:cp_blend} \((\lambda=0.5)\) exhibits the most stable overall behaviour across gap sizes and reward distributions. This policy typically combines fast convergence in the best-arm selection rate with low cumulative regret, suggesting that balancing upper-tail optimism with lower-tail information can improve the exploration--exploitation trade-off. In contrast, more extreme choices of \(\lambda\), and in particular \(\lambda=1\), may become overly conservative in some scenarios, leading to slower identification of the optimal arm, especially under heavy-tailed or positively skewed reward distributions. The learning advantage of \ref{eq:cp_blend} \((\lambda=0.5)\) is accompanied by a favourable uncertainty quantification profile. Across the considered scenarios, the conformal bandit variants generally produce more informative, i.e. tighter, predictive intervals while maintaining empirical coverage close to the nominal level \(1-\alpha=0.8\). This value of the nominal coverage is chosen as a reasonable compromise between statistical reliability and reward maximisation. As expected, arms that are selected less frequently tend to exhibit larger uncertainty, resulting in wider intervals and mild overcoverage.

\begin{table*}[t]
\centering
\scriptsize
\renewcommand{\arraystretch}{1.25}
\setlength{\tabcolsep}{4pt}
\caption{Comparison between the Conformal Bandit variants and classical UCB1 in terms of coverage (\%) and mean interval width for the gap setting \(\Delta_k=0.001\), corresponding to \(\mu_1=\mu^*=0.001\) and \(\mu_2=\mu_3=0\), across all reward-generating scenarios: Gaussian, Student-\(t\), and Skew-\(t\).
All results are averaged across $1,000$ MC replicates. Standard deviations across Monte Carlo runs are reported in parentheses.}
\label{tab:0.001}
\begin{tabularx}{\textwidth}{l | *{3}{X} | *{3}{X} | *{3}{X}}
\toprule
& \multicolumn{3}{c|}{\textbf{Gaussian}} 
& \multicolumn{3}{c|}{\textbf{Student-$t$}} 
& \multicolumn{3}{c}{\textbf{Skew-$t$}} \\
\textbf{Algorithm} 
& Arm1 & Arm2 & Arm3
& Arm1 & Arm2 & Arm3
& Arm1 & Arm2 & Arm3 \\
\midrule
\multicolumn{10}{c}{\textbf{Coverage (\%)}} \\
\midrule

CP-Bandit ($\lambda=0$)        
& 81.62 (2.63) & 84.36 (3.21) & 84.55 (3.25)
& 82.12 (2.67) & 83.57 (2.93) & 83.48 (2.97)
& 81.23 (2.36) & 85.74 (2.95) & 84.25 (3.53) \\

CP-Bandit ($\lambda=0.7$)  
& 83.35 (4.89) & 86.01 (5.36) & 86.07 (5.30)
& 83.41 (4.61) & 84.58 (4.78) & 84.39 (4.84)
& 83.27 (4.54) & 86.78 (4.48) & 83.51 (4.58) \\

CP-Bandit ($\lambda=1$)  
& 84.04 (5.12) & 85.84 (5.63) & 86.00 (5.52)
& 84.17 (4.65) & 84.21 (4.71) & 84.43 (4.84)
& 84.07 (4.67) & 85.87 (4.91) & 83.19 (4.59) \\

CP-Bandit ($\lambda=0.5$)  
& 82.48 (4.20) & 85.91 (4.76) & 85.98 (4.83)
& 82.75 (4.14) & 84.60 (4.28) & 84.58 (4.29)
& 81.84 (3.27) & 87.00 (4.23) & 84.29 (4.29) \\

UCB1
& 94.21 (0.94) & 95.49 (0.87) & 95.46 (0.89)
& 86.23 (1.52) & 87.74 (1.47) & 87.75 (1.62)
& 96.11 (0.77) & 96.92 (0.64) & 97.11 (0.69) \\

\midrule
\multicolumn{10}{c}{\textbf{Mean Interval Width}} \\
\midrule

CP-Bandit ($\lambda=0$)  
& 0.0272 (0.0023) & 0.0299 (0.0033) & 0.0301 (0.0033)
& 0.0370 (0.0058) & 0.0397 (0.0066) & 0.0397 (0.0082)
& 0.0197 (0.0022) & 0.0236 (0.0049) & 0.0207 (0.0036) \\

CP-Bandit ($\lambda=0.7$)  
& 0.0306 (0.0070) & 0.0342 (0.0074) & 0.0347 (0.0076)
& 0.0444 (0.0157) & 0.0476 (0.0156) & 0.0470 (0.0162)
& 0.0238 (0.0076) & 0.0278 (0.0089) & 0.0222 (0.0105) \\

CP-Bandit ($\lambda=1$)  
& 0.0322 (0.0078) & 0.0346 (0.0079) & 0.0348 (0.0080)
& 0.0477 (0.0199) & 0.0491 (0.0183) & 0.0479 (0.0155)
& 0.0266 (0.0100) & 0.0271 (0.0092) & 0.0229 (0.0128) \\

CP-Bandit ($\lambda=0.5$)  
& 0.0290 (0.0055) & 0.0332 (0.0061) & 0.0334 (0.0065)
& 0.0410 (0.0129) & 0.0460 (0.0140) & 0.0458 (0.0140)
& 0.0210 (0.0045) & 0.0272 (0.0086) & 0.0212 (0.0043) \\

UCB1
& 0.0479 (0.0009) & 0.0509 (0.0010) & 0.0510 (0.0010)
& 0.0480 (0.0015) & 0.0509 (0.0017) & 0.0510 (0.0018)
& 0.0480 (0.0009) & 0.0508 (0.0010) & 0.0510 (0.0010) \\

\bottomrule
\end{tabularx}
\end{table*}

\begin{table*}[t]
\centering
\scriptsize
\renewcommand{\arraystretch}{1.25}
\setlength{\tabcolsep}{4pt}
\caption{
Comparison between the Conformal Bandit variants and classical UCB1 in terms of coverage (\%) and mean interval width for the gap setting \(\Delta_k=0.005\), corresponding to \(\mu_1=\mu^*=0.005\) and \(\mu_2=\mu_3=0\), across all reward-generating scenarios: Gaussian, Student-\(t\), and Skew-\(t\).
All results are averaged across $1,000$ MC replicates. Standard deviations across Monte Carlo runs are reported in parentheses.}
\label{tab:0.005}
\begin{tabularx}{\textwidth}{l | *{3}{X} | *{3}{X} | *{3}{X}}
\toprule
& \multicolumn{3}{c|}{\textbf{Gaussian}}
& \multicolumn{3}{c|}{\textbf{Student-$t$}}
& \multicolumn{3}{c}{\textbf{Skew-$t$}} \\
\textbf{Algorithm}
& Arm1 & Arm2 & Arm3
& Arm1 & Arm2 & Arm3
& Arm1 & Arm2 & Arm3 \\
\midrule
\multicolumn{10}{c}{\textbf{Coverage (\%)}} \\
\midrule

CP-Bandit ($\lambda=0$) 
& 80.14 (0.23) & 87.57 (3.02) & 87.74 (3.20)
& 80.17 (0.55) & 85.84 (2.81) & 85.72 (2.92)
& 80.11 (0.25) & 87.31 (3.56) & 87.16 (3.31) \\

CP-Bandit ($\lambda=0.7$) 
& 80.31 (1.44) & 88.38 (4.82) & 88.39 (4.59)
& 80.98 (2.81) & 86.40 (4.77) & 86.16 (4.78)
& 80.16 (0.74) & 87.95 (4.21) & 87.33 (4.45) \\

CP-Bandit ($\lambda=1$) 
& 81.08 (3.14) & 87.60 (5.33) & 87.80 (5.20)
& 81.88 (3.55) & 85.73 (4.82) & 85.76 (4.92)
& 80.86 (2.45) & 86.93 (4.58) & 86.13 (5.15) \\

CP-Bandit ($\lambda=0.5$) 
& 80.16 (0.75) & 88.78 (4.28) & 88.79 (4.29)
& 80.46 (1.84) & 86.50 (4.43) & 86.52 (4.22)
& 80.10 (0.21) & 87.91 (4.22) & 87.81 (3.98) \\

UCB1
& 90.51 (0.99) & 97.23 (0.79) & 97.26 (0.79)
& 82.18 (1.40) & 90.12 (1.56) & 90.08 (1.55)
& 94.50 (0.78) & 97.58 (0.63) & 97.65 (0.67) \\

\midrule
\multicolumn{10}{c}{\textbf{Mean Interval Width}} \\
\midrule

CP-Bandit ($\lambda=0$) 
& 0.0260 (0.0006) & 0.0344 (0.0036) & 0.0343 (0.0039)
& 0.0337 (0.0014) & 0.0451 (0.0082) & 0.0450 (0.0096)
& 0.0188 (0.0006) & 0.0264 (0.0064) & 0.0246 (0.0053) \\

CP-Bandit ($\lambda=0.7$) 
& 0.0263 (0.0023) & 0.0371 (0.0067) & 0.0374 (0.0069)
& 0.0368 (0.0106) & 0.0518 (0.0156) & 0.0508 (0.0152)
& 0.0190 (0.0016) & 0.0285 (0.0085) & 0.0274 (0.0160) \\

CP-Bandit ($\lambda=1$) 
& 0.0277 (0.0053) & 0.0366 (0.0074) & 0.0368 (0.0075)
& 0.0404 (0.0135) & 0.0514 (0.0176) & 0.0504 (0.0153)
& 0.0206 (0.0060) & 0.0279 (0.0088) & 0.0262 (0.0153) \\

CP-Bandit ($\lambda=0.5$) 
& 0.0261 (0.0010) & 0.0373 (0.0062) & 0.0373 (0.0063)
& 0.0346 (0.0056) & 0.0513 (0.0151) & 0.0517 (0.0157)
& 0.0188 (0.0006) & 0.0285 (0.0085) & 0.0264 (0.0068) \\

UCB1
& 0.0417 (0.0006) & 0.0563 (0.0012) & 0.0564 (0.0012)
& 0.0418 (0.0010) & 0.0564 (0.0021) & 0.0564 (0.0021)
& 0.0416 (0.0006) & 0.0564 (0.0012) & 0.0565 (0.0012) \\

\bottomrule
\end{tabularx}
\end{table*}

\begin{table*}[t]
\centering
\scriptsize
\renewcommand{\arraystretch}{1.25}
\setlength{\tabcolsep}{4pt}
\caption{
Comparison between the Conformal Bandit variants and classical UCB1 in terms of coverage (\%) and mean interval width for the gap setting \(\Delta_k=0.05\), corresponding to \(\mu_1=\mu^*=0.05\) and \(\mu_2=\mu_3=0\), across all reward-generating scenarios: Gaussian, Student-\(t\), and Skew-\(t\).
All results are averaged across $1,000$ MC replicates. Standard deviations across Monte Carlo runs are reported in parentheses.}
\label{tab:0.05}
\begin{tabularx}{\textwidth}{l | *{3}{X} | *{3}{X} | *{3}{X}}
\toprule
& \multicolumn{3}{c|}{\textbf{Gaussian}}
& \multicolumn{3}{c|}{\textbf{Student-$t$}}
& \multicolumn{3}{c}{\textbf{Skew-$t$}} \\
\textbf{Algorithm}
& Arm1 & Arm2 & Arm3
& Arm1 & Arm2 & Arm3
& Arm1 & Arm2 & Arm3 \\
\midrule
\multicolumn{10}{c}{\textbf{Coverage (\%)}} \\
\midrule

CP-Bandit ($\lambda=0$) 
& 80.12 (0.21) & 89.52 (4.30) & 89.68 (4.25)
& 80.11 (0.21) & 87.55 (4.43) & 87.57 (4.42)
& 80.10 (0.20) & 88.05 (4.51) & 88.29 (4.28) \\

CP-Bandit ($\lambda=0.7$) 
& 80.12 (0.21) & 89.58 (4.29) & 89.50 (4.22)
& 80.10 (0.21) & 87.78 (4.50) & 87.76 (4.59)
& 80.10 (0.20) & 88.27 (4.29) & 88.29 (4.30) \\

CP-Bandit ($\lambda=1$) 
& 80.12 (0.21) & 89.50 (4.41) & 89.73 (4.32)
& 80.11 (0.21) & 87.73 (4.24) & 87.65 (4.34)
& 80.10 (0.20) & 88.12 (4.33) & 88.41 (4.35) \\

CP-Bandit ($\lambda=0.5$) 
& 80.13 (0.21) & 89.37 (4.35) & 89.52 (4.46)
& 80.10 (0.21) & 87.89 (4.59) & 87.68 (4.48)
& 80.10 (0.20) & 87.95 (4.44) & 88.57 (4.25) \\

UCB1
& 79.39 (0.87) & 100.00 (0.00) & 100.00 (0.00)
& 71.42 (0.99) & 99.52 (1.08) & 99.48 (1.12)
& 89.00 (0.73) & 99.87 (0.55) & 99.87 (0.54) \\

\midrule
\multicolumn{10}{c}{\textbf{Mean Interval Width}} \\
\midrule

CP-Bandit ($\lambda=0$) 
& 0.0260 (0.0006) & 0.0389 (0.0063) & 0.0388 (0.0063)
& 0.0335 (0.0010) & 0.0550 (0.0145) & 0.0550 (0.0142)
& 0.0188 (0.0006) & 0.0290 (0.0086) & 0.0283 (0.0089) \\

CP-Bandit ($\lambda=0.7$) 
& 0.0260 (0.0006) & 0.0386 (0.0060) & 0.0387 (0.0064)
& 0.0335 (0.0010) & 0.0562 (0.0164) & 0.0549 (0.0148)
& 0.0188 (0.0006) & 0.0294 (0.0089) & 0.0285 (0.0146) \\

CP-Bandit ($\lambda=1$) 
& 0.0260 (0.0006) & 0.0388 (0.0061) & 0.0391 (0.0064)
& 0.0335 (0.0010) & 0.0558 (0.0165) & 0.0548 (0.0146)
& 0.0188 (0.0006) & 0.0291 (0.0083) & 0.0284 (0.0092) \\

CP-Bandit ($\lambda=0.5$) 
& 0.0260 (0.0006) & 0.0387 (0.0063) & 0.0387 (0.0062)
& 0.0335 (0.0010) & 0.0560 (0.0164) & 0.0564 (0.0163)
& 0.0188 (0.0006) & 0.0293 (0.0099) & 0.0283 (0.0094) \\

UCB1
& 0.0302 (0.0000) & 0.1506 (0.0033) & 0.1506 (0.0031)
& 0.0302 (0.0001) & 0.1508 (0.0053) & 0.1510 (0.0053)
& 0.0302 (0.0000) & 0.1505 (0.0031) & 0.1507 (0.0029) \\

\bottomrule
\end{tabularx}
\end{table*}

\section{Hidden Markov Models for Market Regime Detection}
\label{app:HMM}

To estimate latent market regimes, we employ a Hidden Markov Model (HMM), a probabilistic framework in which the observed return series is generated by an unobserved (hidden) state process. A detailed description of HMM can be found in \cite{zucchini_macdonald_langrock2016}. In our setting, each hidden state corresponds to a distinct market regime (Bull, Neutral, Bear), and transitions between regimes follow a first-order Markov chain.

Let $\{R_t\}_{t=1}^T$ denote the observed returns and $\{Z_t\}_{t=1}^T$ the latent regime sequence, where $Z_t \in \{1,\ldots,S\}$ for $S$ regimes. An HMM is specified by:
\begin{itemize}
    \item the initial distribution $\delta_i = \mathbb{P}(Z_1 = i)$;
    \item the transition matrix $A = (a_{ij})$, where $a_{ij} = \mathbb{P}(Z_{t+1}=j \mid Z_t=i)$;
    \item the emission distribution $p(R_t \mid Z_t)$ describing observations conditional on the regime.
\end{itemize}

In this work we adopt a \textit{Gaussian-emission HMM} with $S=3$, implemented via the \texttt{GaussianHMM} class of the \texttt{hmmlearn} library. Conditional on regime $s$, returns follow:
\[
R_t \mid (Z_t = s) \sim \mathcal{N}(\mu_s, \Sigma_s),
\]
where $\mu_s$ and $\Sigma_s$ are the regime-specific mean vector and covariance matrix. The parameters $(\{\delta_s\}_{s=1}^S, A, \{\mu_s, \Sigma_s\}_{s=1}^S)$ are estimated through maximum likelihood via the Expectation--Maximization (EM) algorithm~\cite{dempster_laird_rubin1977}.

Once the model is fitted, the most likely regime at time $t$ is obtained using the \textit{filtering distribution}:
\[
\widehat{Z}_t = \arg\max_{s} \mathbb{P}(Z_t = s \mid R_{1:t}),
\]
computed through the forward--backward algorithm \cite{zucchini_macdonald_langrock2016}. The resulting inferred regimes provide a data-driven segmentation of market conditions and are used to activate the regime-aware variants of the conformal bandit policies.

\subsection{Pseudo-algorithms of the Regime-Aware Policies}\label{app:Algorithms' description}
\begin{figure}[ht]
\centering
\fbox{
\parbox{0.95\linewidth}{
\textbf{Algorithm 2: (Randomised) Regime-Aware Conformal Bandits}

\medskip
\textbf{Input:} Number of arms $K$, horizon $T$, conformal prediction strategy (e.g., CQR), miscoverage $\alpha \in (0,1)$,\\
\phantom{\textbf{Input:}} exploration parameter $\epsilon_t \in [0,1]$ (e.g., as a decreasing function of $t$; see Eq. (14)),\\
\phantom{\textbf{Input:}} {\it conformal index} $I \in$ {(CP-UCB), (CP-Bandit), (CP-ESI)}, regime-detection model (e.g., HMM). \\

\textbf{Output:} Selected arms $\{a^*_t\}_{t=1}^T$.
\smallskip

\begin{tabbing}
\hspace*{0.5cm}\=\hspace*{0.7cm}\=\hspace*{0.7cm}\=\kill

1. \textbf{for} $t = 1$ \textbf{to} $2K$ \textbf{do} \` (Pure exploration – warm-up)\\
2.\> Select arm $a^*_t = ((t-1) \bmod K) + 1$.\\
3.\> Observe associated state-reward pair $(X_{t}, Y_{a^*_t,t})$.\\
4. \textbf{end for}\\[4pt]

5. \textbf{for} $t = 2K+1$ \textbf{to} $T$ \textbf{do} \` ((Regime-Aware) Conformal bandit selection)\\[2pt]
6.\>\> Infer market regime $\mathcal{R}_t \in \{\texttt{Bull},\texttt{Neutral},\texttt{Bear}\}$ from regime-detection model.\\
7.\>\> \textbf{for} $k = 1,\dots,K$ \textbf{do}\\
8.\>\> Given observed state-reward pairs $(X_i, Y_{k,i})_{i = 1}^{N_{k,t-1}}$:\\
9.\>\>\> \ \ \ Compute conformal intervals $\mathcal{C}_{k,t}^{1-\alpha} = \Big[L_{k,t},\, U_{k, t}\Big]$;\\
10.\>\>\> \ \ \ Compute regime-aware pseudo-deterministic conformal index $I_{k,\mathcal{R}_t,t}$:\\
11.\>\>\> \ \ \ \textbf{if} $\mathcal{R}_t \in \{\texttt{Bull},\texttt{Neutral}\}$, set $I_{k,\mathcal{R}_t,t}=U_{k,t}$ \textit{(optimistic)}.\\
12.\>\>\> \ \ \ \textbf{if} $\mathcal{R}_t=\texttt{Bear}$, set $I_{k,\mathcal{R}_t,t}=-|L_{k,t}|$ \textit{(downside-protective)}.\\
13.\>\> \textbf{end for}\\[2pt]
14.\>\> Get regime-aware pseudo-deterministic optimal arm $\tilde{a}_t^{*} = \arg\max_{k \in \mathcal{A}} I_{k,\mathcal{R}_t,t}.$\\[4pt]

15.\>\> Draw $Z_t \sim \mathrm{Bernoulli}(1-\epsilon_t)$. \` (Bandit randomisation)\\
16.\>\> \textbf{if} $Z_t = 1$ \textbf{then}\\
17.\>\>\>$a_{t}^{*} = \tilde{a}_t^{*}$\\
18.\>\> \textbf{else}\\
19.\>\>\> $a_{t}^{*} \sim \text{Unif}_{\mathcal{A} \setminus \tilde{a}_t^{*}}$.\\
20.\>\> \textbf{end if}\\
21.\>\> Observe associated reward $Y_{a^*_t,t}$ and update state-reward pairs $(X_i, Y_{k,i})_{i = 1}^{N_{k,t}}$ for $k = a^*_t$.\\
22.\> \textbf{end for}\\[2pt]
23. \textbf{return} selected arms $\{a^*_t\}_{t=1}^T$.
\end{tabbing}
}
}
\captionsetup{labelformat=empty}
\caption{Algorithm 2: Regime-Aware Conformal bandits}
\label{alg:regime_aware_Conformal_bandits}
\end{figure}

\begin{figure}[ht]
\centering
\fbox{
\parbox{0.95\linewidth}{
\textbf{Algorithm 3: Regime-Aware MV-UCB1}

\medskip
\textbf{Input:} Number of arms $K$; horizon $T$; UCB-exploration constant $\beta$; mean--variance weight $\rho$;\\
\phantom{\textbf{Input:}} regime-switching model (e.g. HMM).\\
\textbf{Output:} Selected arms $\{a^*_t\}_{t=1}^T$

\medskip
\begin{tabbing}
\hspace*{0.5cm}\=\hspace*{0.7cm}\=\hspace*{0.7cm}\=\kill

1.\> Initialise empirical means $\widehat{\mu}_{k,0}$, variances $\widehat{\sigma}^2_{k,0}$, and pull counts $N_{k,0}=0$.\\[2pt]

2.\> \textbf{for} $t = 1$ \textbf{to} $K$ \textbf{do}\\
3.\>\> Play arm $a^*_t = t$.\\
4.\>\> Observe reward $Y_{a^*_t,t}$ and update $\widehat{\mu}_{a^*_t,t}$, $\widehat{\sigma}^2_{a^*_t,t}$, $N_{a^*_t,t}$.\\
5.\> \textbf{end for}\\[4pt]
6.\> \textbf{for} $t = K+1$ \textbf{to} $T$ \textbf{do}\\
7.\>\> Infer regime $\mathcal{R}_t \in \{\texttt{Bull}, \texttt{Neutral}, \texttt{Bear}\}$ from regime model.\\
8.\>\> \textbf{for} $k = 1,\dots,K$ \textbf{do}\\
9.\>\>\> Given the observed reward $(Y_{k,i})_{i=1}^{N_{k,t-1}}$:\\
10.\>\>\>  $\quad $  Compute UCB exploration term for each arm: $B_{k,t} = \sqrt{\frac{\beta \log t}{N_{k,t}}}$.\\
11.\>\>\>   $\quad $      \textbf{if} $\mathcal{R}_t \in \{\texttt{Bull}, \texttt{Neutral}\}$, set $S_{k,t} = \widehat{\mu}_{k,t} + B_{k,t}$.\\
12.\>\>\> $\quad $ \textbf{if} $\mathcal{R}_t=\texttt{Bear}$, compute mean--variance score $\widehat{MV}^{\rho}_{k,t}
=\rho \, \widehat{\mu}_{k,t}-(1 - \rho)\,\widehat{\sigma}_{k,t}$\\
13.\>\>\> $\quad $ $\quad $ and set $S_{k,t} = \widehat{MV}^{\rho}_{k,t} + B_{k,t}$.\\
14.\>\>\> $\quad $ \textbf{end if}\\
15.\>\> \textbf{end for}\\[2pt]
16.\>\> Select arm $a^*_t = \arg\max_{k \in \mathcal{A}} S_{k,t}$.\\
17.\>\> Observe reward $Y_{a^*_t,t}$ and update $\widehat{\mu}_{a^*_t,t}$, $\widehat{\sigma}^2_{a^*_t,t}$, $N_{a^*_t,t}$.\\
18.\> \textbf{end for}\\[2pt]
19.\> \textbf{return} $\{a^*_t\}_{t=1}^T$.
\end{tabbing}
}}
\captionsetup{labelformat=empty}
\caption{Algorithm 3: Regime-Aware MV-UCB1}
\label{alg:regime_aware_mv_ucb1}
\end{figure}
Procedural descriptions of the proposed regime-aware policies in \eqref{eq:cp_regimeaware} and \eqref{eq:Regime-Aware_MV-UCB1} are reported in \nameref{alg:regime_aware_Conformal_bandits} and \nameref{alg:regime_aware_mv_ucb1}, respectively.

\section{Case Study: Comparison to Full-information Bandits}
\label{app:full_information}

\begin{figure}[t]
    \centering
    \includegraphics[width=\textwidth]{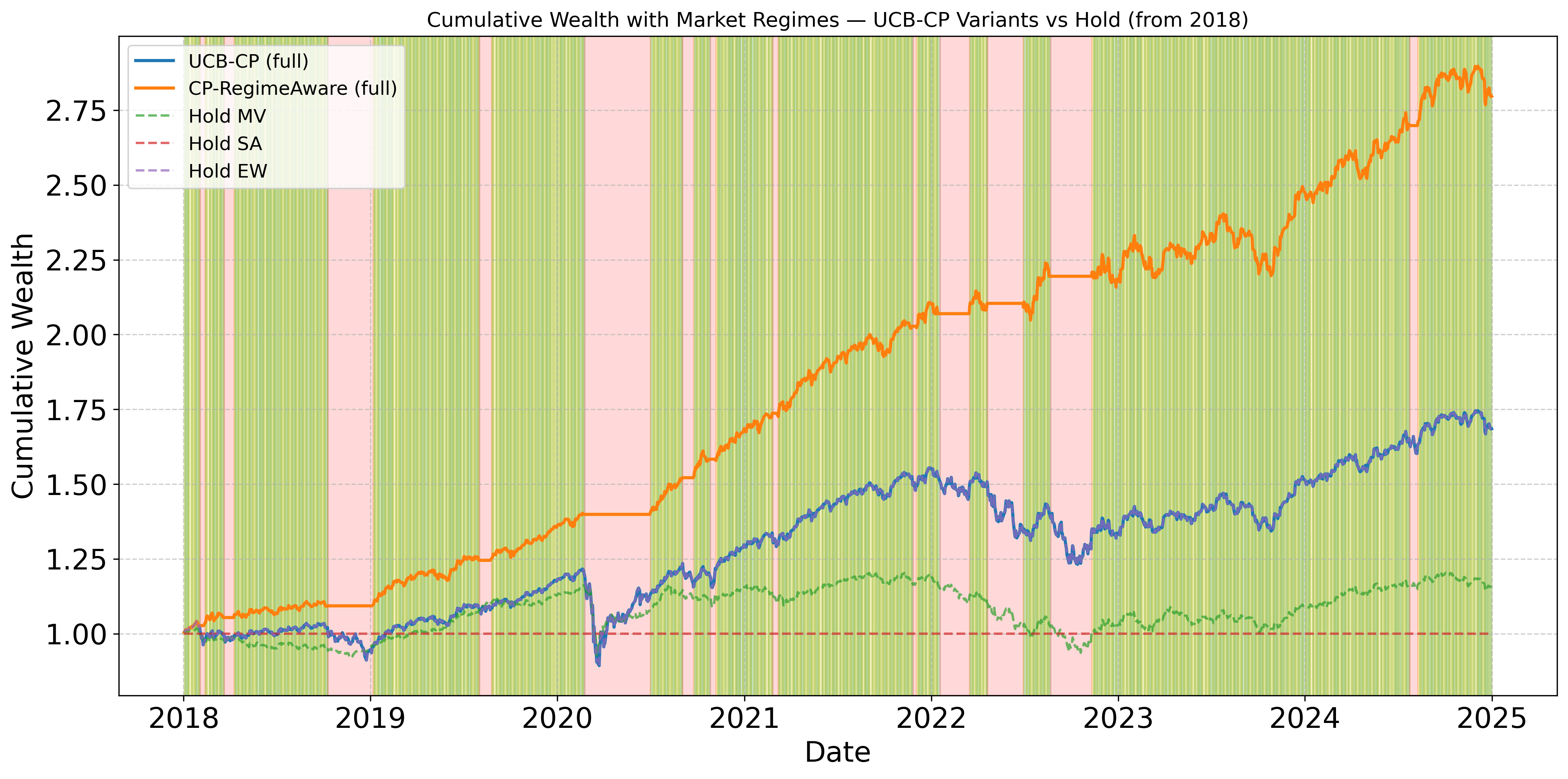}
    \caption{
    Cumulative wealth of CP-based bandit policies under a full-information setting, compared with EW, MV and SA portfolio benchmarks. Background shading highlights market regimes inferred via HMM: \textit{green} denotes \texttt{Bull} phases, \textit{gray} \texttt{Neutral} markets, and \textit{pink} \texttt{Bear} episodes. Shaded bands around randomised CP policies indicate $95\%$ confidence intervals computed over $1,000$ MC runs.
    }
    \label{fig:full_cum_returns}
\end{figure}

In the main text we focused on the classical bandit (partial-information) setting, where at each round $t$ only the reward of the selected arm is observed.
For comparison, in this appendix we consider a \emph{full-information} scenario, in which at each time $t$ the entire reward vector 
$\{Y_{k,t}\}_{k \in \mathcal{A}}$ is observed, regardless of which arm is chosen.\\
In this regime, exploration is unnecessary, as all rewards are revealed at each step.
Accordingly, we employ the \emph{deterministic} versions of the conformal bandit policies introduced in Section \ref{sec: CP-bandits}, together with their regime-aware extensions.
For a direct comparison with the partial-information results, we evaluate the same policies—namely, the CP-UCB rule~\ref{eq:cp_ucb} and the Conformal Regime-Aware policy~\ref{eq:cp_regimeaware} — using the same dataset and trading period considered in the main experiments.
At each rebalancing date, portfolio weights are allocated according to the selected CP-based policy, using full access to the realised rewards of all strategic arms.
Figure~\ref{fig:full_cum_returns} reports the resulting cumulative wealth trajectories and benchmarks them against the Equally-Weighted (EW) portfolio and the Mean--Variance (MV) portfolio.
As in the main analysis, the coloured background identifies market regimes---Bull (green), Neutral (gray), and Bear (pink)---inferred via the Hidden Markov Model (HMM) described in Appendix~\ref{app:HMM}.\\
As expected, Figure~\ref{fig:full_cum_returns} shows that full-information policies outperform their bandit counterparts, as they operate with maximal learning efficiency and face no uncertainty about unobserved rewards.
This leads to higher cumulative wealth—especially for the regime-aware policy—although at the expense of larger data requirements and increased computational complexity. The performance metrics in Table~\ref{tab:full_info_perf} reinforce the graphical evidence. 

\begin{table}[ht]
\centering
\caption{Performance metrics in the full-information setting (2018--2025).}
\label{tab:full_info_perf}
\begin{tabular}{lcccc}
\toprule
\textbf{Strategy} & \textbf{Total Return} & \textbf{Annualized Sharpe} &
\textbf{Max Drawdown} & \textbf{Calmar} \\
\midrule
UCB-CP (full)            & 0.6774 & 0.6268 & 0.2656 & 0.2893 \\
CP-RegimeAware (full)    & 1.7843 & 1.9966 & 0.0861 & 1.8333 \\
Hold MV                  & 0.1548 & 0.2776 & 0.2219 & 0.0938 \\
Hold EW                  & 0.6774 & 0.6268 & 0.2656 & 0.2893 \\
\bottomrule
\end{tabular}
\end{table}

\end{document}